\documentclass{article}

\usepackage[preprint,nonatbib]{neurips_2024}

\usepackage[utf8]{inputenc} 
\usepackage[T1]{fontenc}    
\usepackage{hyperref}       
\usepackage{url}            
\usepackage{booktabs}       
\usepackage{amsfonts}       
\usepackage{nicefrac}       
\usepackage{microtype}      
\usepackage{xcolor}         
\usepackage{graphicx}       
\usepackage{tabularx}
\usepackage{subcaption}

\newenvironment{customquote}[1]%
  {\list{}{\leftmargin=#1\rightmargin=#1}\item[]}%
  {\endlist}

\title{Alice in Wonderland: Simple Tasks Showing Complete Reasoning Breakdown in State-Of-the-Art Large Language Models}

\author{%
\textbf{Marianna Nezhurina}$^{1, 2, 4 *}$ \quad \textbf{Lucia Cipolina-Kun}$^{1,2,3}$ \quad 
\textbf{Mehdi Cherti}$^{1, 2, 4}$ \quad \textbf{Jenia Jitsev}$^{1, 2, 4 *}$ \\
$^{1}$LAION \quad $^{2}$Juelich Supercomputing Center (JSC), Research Center Juelich (FZJ) \\  $^{3}$ School of Electrical and Electronic Engineering, University of Bristol  \\ $^{4}$ Open-$\Psi$ (Open-Sci) Collective \\
\texttt{$^{*}$Corresponding authors}: \texttt{\{m.nezhurina,j.jitsev\}@fz-juelich.de,contact@laion.ai} \\
}

\begin{document}

\maketitle

\begin{abstract}
Large Language Models (LLMs) are often described as being instances of foundation models - that is, models that possess strong generalization and therefore transfer robustly across various tasks and conditions in few-shot or zero-shot manner, while exhibiting scaling laws that predict generalization improvement when increasing the pre-training scale. Such claims rely on measurements by various standardized benchmarks that suppose to reflect core functions like generalization and reasoning, where state-of-the-art (SOTA) models score high. We demonstrate here a dramatic breakdown of generalization and basic reasoning of all SOTA models which claim strong function, including advanced models like GPT-4 or Claude 3 Opus trained at the largest scales, using a simple, short common sense problem formulated in concise natural language, easily solvable by humans (AIW problem). The breakdown is dramatic as it manifests on a simple problem in both low average performance and strong performance fluctuations on natural variations in problem template that do not change either problem structure or its difficulty at all. By testing models on further control problems with similar form, we rule out that breakdown might be rooted in minor low-level issues like natural language or numbers parsing. We also observe strong overconfidence in the wrong solutions, expressed in form of plausible sounding explanation-like confabulations. Various standard interventions in an attempt to get the right solution, like chain-of-thought prompting, or urging the models to reconsider the wrong solutions again by multi step re-evaluation, fail. We use these observations to stimulate re-assessment of the capabilities of current generation of LLMs as claimed by standardized benchmarks. Such re-assessment also requires common action to create standardized benchmarks that would allow proper detection of such deficits in generalization and reasoning that obviously remain undiscovered by current state-of-the-art evaluation procedures, where SOTA LLMs manage to score high. \footnote{Code for reproducing experiments in the paper and raw experiments data can be found at \href{https://github.com/LAION-AI/AIW}{AIW repo}}.
\end{abstract}


\section{Introduction}


In the recent breakthroughs in transferable learning that were achieved in various classical domains of machine learning like visual recognition \cite{radford2021learning} or language understanding \cite{devlin2018bert,raffel2020exploring,brown2020language}, large language models (LLMs) have played a very prominent role. The generic form and scalability of autoregressive language modelling~\cite{brown2020language} allowed to push towards training scales not achievable before with conventional supervised label-based learning. Scaling laws derived from experiments on smaller scales hinted on strong function and generalization capability appearing at larger scales~\cite{Kaplan2020, hoffmann2022}, which was then confirmed by training models at the large scales, measuring their performance on set of standardized benchmarks (MMLU, HellaSwag, ARC, MATH, GSM8k, etc) where they scored high on few- and zero-shot transfer across various tasks~\cite{kojima2022large}, following accurately the predictions~\cite{brown2020language, Kaplan2020, achiam2023gpt,touvron2023llama,touvron2023llama2,jiang2023mistral}.

There were however observations made by various works that questioned the claimed strong generalization, transfer and reasoning capabilities attributed to LLMs \cite{mitchell2023we}. These works pointed out various function failures that were seemingly incompatible with postulated strong capabilities as measured by standardized benchmarks \cite{wu2023reasoning,golchin2023time, li2024task,frieder2024mathematical}. However, it has also been noted that observed failures can frequently be addressed through simple adjustments to the prompts or by repeated execution and evaluation using majority voting, or by requesting the model to perform self-verification \cite{kadavath2022language, wang2023selfconsistency,zhou2023leasttomost, zhang2024self,pan2024automatically}. For many further anecdotical reports of failures, there was lack of systematic evaluation, e.g. without controlling accurately the conditions under which failures occur, and often unclarity about potentially ambigious formulations that were leading to failures.  It remained thus unclear where those observations of failures were pointing to some fundamental deficits in core model function like generalization and reasoning, or whether those were just due to ill-posed problem formulation or due to minor issues easily resolvable by simple interventions, leaving claim of strong core function backed up by high scores on standardized benchmarks and also by observed strong performance on selected complex tasks unaffected.

\begin{figure}[tb!]
    \centering
    \includegraphics[width=1.0\textwidth]{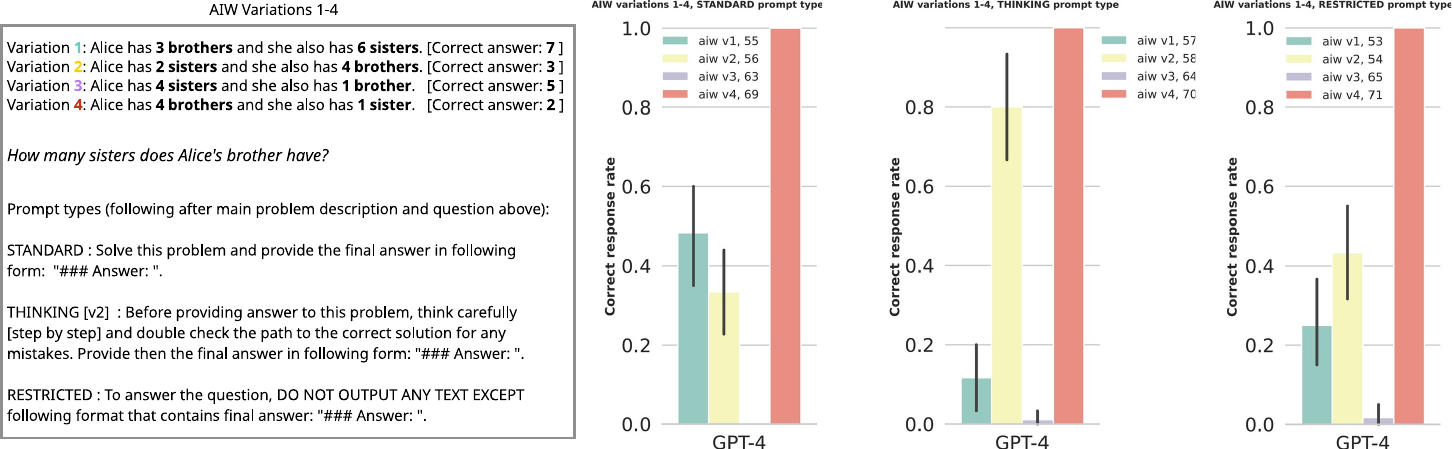}
    \caption{AIW problem is simple common sense math problem with short, concise formulation, serving as a minimalist setting to test model generalization,  Problem template is used to create AIW variations 1-4 (\textbf{left}) by instantiating different numbers $N,M$ of brothers and sisters, keeping problem structure and difficulty unchanged. We measure sensitivity of models to those variations, observing strong performance fluctuations. Here on example of GPT-4 (gpt-4-0613) tested with various prompt types STANDARD, THINKING, RESTRICTED, a color per each variation 1-4, executing 60 trials per each variation. Correct response rate varies strongly depending on the variation. E.g., it drops close to 0 for variation 3, while going up to 1 for variation 4. This observation is consistent for different prompt types - STANDARD, THINKING and RESTRICTED (from left to right), showing that observed fluctuations are not due to prompt variation. Strong performance fluctuations on natural, structure preserving variations of such a simple problem points to severe lack of robustness and generalization deficits. Numbers in the legend are prompt IDs (Suppl. Tab. \ref{tab:aiw_prompt_types})}
    \label{fig:aiw_original_fluctuations_expl}
\end{figure}

To shed light on this situation, we study whether the claim of SOTA LLMs possessing strong functions across various complex problems can be put to test by using problems that are very simple, in contrast to those employed by various standardized benchmarks. We introduce a short conventional common sense problem template that can be formulated without any ambiguities in concise natural language and can be easily solved by humans. The problem (in following Alice in Wonderland, AIW problem) has following template: \textit{\textbf{"Alice has $N$ brothers and she also has $M$ sisters. How many sisters does Alice's brother have?"}}. Crucially, instantiating natural numbers $N,M$ $\leq 7$ in the problem template allows us to naturally introduce controlled source of systematic variations that do not change problem structure and difficulty at all and thus should not affect ability to solve it (Fig. \ref{fig:aiw_original_fluctuations_expl}). 


We use then this technique of creating problem structure and difficulty preserving variations to measure  models' sensitivity to problem irrelevant perturbations across multiple repetitive trials, testing models' generalization - should it be intact, models should handle the simple problem equally well across variations (Fig. \ref{fig:aiw_original_fluctuations_expl}).

Surprisingly, when confronted with the AIW problem and its structure preserving variations, all SOTA models including advanced large-scale ones (eg GPT-4~\cite{GPT4}, Claude 3 Opus~\cite{anthropic2024claude}) suffer severe function breakdown. This breakdown manifests (i) in average correct response rates that are unexpectedly low for such a simple problem and in (ii) strong fluctuations in correct response rates across AIW problem variations despite those being entirely irrelevant for coping with the problem. Strong fluctuations remain despite using various standard interventions to improve model function like chain-of-thought prompting. We conduct control experiments using various AIW problem versions designed to test models' ability to handle operations required for solving original AIW problem (AIW Light problems). By observing that models are successfully coping with those control AIW problems, we are able to rule out that observed failures might be rooted in minor low level issues, e.g tokenization/natural language parsing, handling the basic family structure, binding attributes, or executing arithmetic operations necessary to solve the problem. Despite a specific form of the simple AIW problem, we can thus conclude that observed failure has generic character. The lack of robustness revealed in all SOTA models by problem irrelevant variations of a simple problem points to severe generic deficits in generalization and reasoning.

Confronted with evidence of zero shot generalization breakdown all tested SOTA LLMs exhibit on such a simple problem, we investigate further AIW problem versions to see whether the same phenomenon of model sensitivity to problem structure preserving variations appears consistently. We use same technique of varying numbers in problem templates and observe same strong fluctuations of model performance also on other problem settings, obtaining further evidence that the generalization deficit is generic and not unique to original AIW formulation. We also see most recent reasoning models like DeepSeek-R1 or o1-mini exhibiting strong fluctuations on AIW problem versions, revealing lack of robustness and generalization deficit also for this model class.

The observed breakdown of all tested SOTA LLMs can be considered surprising and dramatic, not only because LLMs show rather low average correct response rates on such a simple problem, but importantly also because advanced SOTA LLMs claiming strong function exhibit strong performance fluctuations on AIW variations that do not change problem structure or its difficulty at all, being just instantiations of different numbers natural to the problem. The lack of robustness in such a simple setting reveals generalization deficits and clearly contradicts claims of robust problem solving on graduate student or olympiad level, as put forward by advanced LLMs, eg GPT-4, Claude 3 Opus, or recent reasoning models like DeepSeek-R1 or o1-mini - as arguably, presented AIW problem versions are far below those difficulty levels. We do not conclude entire absence of reasoning from the observations. Advanced large-scale LLMs (GPT-4/4o, Claude 3 Opus/Claude 3.5 Sonnet, Qwen 2.5 72B, Llama 3.1 405B) have correct response rates substantially higher than zero (in contrast to smaller scale models which show consistently low correct response rates close to 0), and inspecting outputs leading to final correct answers reveals in most cases correct reasoning arriving there. The frequency of such correct reasoning varies though strongly across natural, problem structure preserving variations, making clear that reasoning is fragile and easily disturbed by problem irrelevant perturbations, which is hallmark of poor generalization.

We also observe further breakdown characteristics that makes it severe. We measure distribution of natural numbers responses on output, showing that for AIW variations with low correct response rate, peaks are on wrong answers, excluding majority voting methods as a fix. We observe that wrong responses are often accompanied by persuasive explanation-like confabulations and overconfident tone about correctness of the wrong solutions provided by the models, which can further mislead model users into trusting wrong responses. This is especially troubling for scenarios where users have no simple way to check for the correctness of the provided solutions. We also see models failing to properly detect mistakes and to revise wrong solutions when encouraged to do so in experiments with multi-turn AIW problem interaction and self-verification.

The observed breakdown of function and generalization is in strong contrast to scores on standardized benchmarks, which contain problems of higher difficulty. Many tested models that score high on such benchmarks show correct response rates close to zero across simple AIW problem variations. Claim put forward by standardized benchmarks to properly reflect model capabilities such as generalization and reasoning cannot be upheld in face of the evident failure to spot such severe function deficits as revealed in the simple AIW problem setting. Our study highlights necessity to re-assess capabilities of SOTA LLMs by creating novel benchmarks that properly reflect their true abilities to generalize and reason. Such benchmarks will be able to correctly spot deficits overlooked so far and thus show the path for improvement of current still unsatisfactory state.

\section{Methods \& Experiment Setup}
\label{sec:methods_experiment}

\subsection{Simple common sense reasoning problems and their variations}
\label{subsec:aiw_problem}

\textbf{AIW Problem.} To measure models' sensitivity to problem irrelevant variations and thus probe the zero-shot generalization, we use following problem template: \textit{\textbf{"Alice has $N$ brothers and she also has $M$ sisters. How many sisters does Alice's brother have?"}}. The problem has a simple common sense solution which assumes all sisters and brothers in the problem setting share the same parents. The correct response $C$ - number of sisters - is easily obtained by calculating $C = M + 1$ (Alice and her sisters), which gives the number of sisters Alice's brother has. To create problem variations, we choose to vary natural numbers $N,M \leq 7$, obtaining AIW variations 1-4 (see Suppl. Tab. \ref{tab:aiw_prompt_types}) that all pose same problem using variations irrelevant for problem solving. The resulting variations are as following:

\begin{quote}
    Variation 1. Alice has 3 brothers and she also has 6 sisters. [Correct answer: 7] \\
    Variation 2. Alice has 2 sisters and she also has 4 brothers. [Correct answer: 3]
    Variation 3. Alice has 4 sisters and she also has 1 brother.  [Correct answer: 5]
    Variation 4. Alice has 4 brothers and she also has 1 sister.  [Correct answer: 2]

    Question: How many sisters does Alice's brother have?
\end{quote}

We further use 3 prompt types, RESTRICTED, STANDARD and THINKING, to ensure we measure models across various prompt formulations and check the observations hold independent of employed prompt type (see also Sec. \ref{subsec:model_prompts_parsing} and Suppl. Sec. \ref{appendix:prompt types}). 



\subsubsection{Control AIW Light problems}
\label{subsec:aiw_light_control_problem}


To control for models struggling either with basic family relations structure handling or with executing arithmetic operations in frame of the posed AIW problem, we make various versions of AIW problem - AIW Light Family, AIW Light Arithmetic Siblings and AIW Light Arithmetic Total Girls. The AIW Light problems keep problem template close to the original, changing only the final question part such that the posed modified question tests particular operations. The variations 1-4 are created in the same way like in AIW original by varying natural numbers of brothers and sisters, while ensuring that the natural numbers for final correct answers in AIW original and AIW Light are matched across variations 1-4 (see also Suppl. Sec. \ref{appendix:prompt types}).

\textbf{AIW Light Arithmetic Siblings}. AIW Light Arithmetic Siblings has following problem template: \textit{\textbf{"Alice has $N$ brothers and she also has $M$ sisters. How many siblings does Alice have?"}}. Compared to AIW original, only question part is modified. To solve the problem, summing up already given numbers of brothers and sisters is sufficient - the correct answer is $C=N+M$. This requires basic grasping of relational family structure (realizing Alice's siblings are her sisters and brothers) and selection and execution of elementary arithmetic sum operation. In contrast to AIW original, it does not require execution of set operations nor binding sex attribute to Alice to properly assign her to correct sets. Should the issues with solving AIW original be rooted in selection and execution of elementary arithmetic operations in family frame, we should see models also failing here. Again, we create variations 1-4 by varying natural numbers $N,M$, such that correct responses $C$ are matched with AIW original variations 1-4 (Suppl. Tab. \ref{tab:aiw_light_arithmetic_siblings}). This gives following variations:

\begin{quote}
    Variation 1. Alice has 3 brothers and she also has 4 sisters. [Correct answer: 7] \\
    Variation 2. Alice has 2 sisters and she also has 1 brother. [Correct answer: 3]
    Variation 3. Alice has 4 sisters and she also has 1 brother.  [Correct answer: 5]
    Variation 4. Alice has 1 brother and she also has 1 sister.  [Correct answer: 2]

    Question: How many siblings does Alice have?
\end{quote}

\textbf{AIW Light Family}. AIW Light Family has following problem template: \textit{\textbf{"Alice has $N$ brothers and she also has $M$ sisters. How many brothers does Alice's sister have?"}}. Compared to AIW original, only question part is modified.
To solve the problem, reporting already given number of brothers is sufficient - the correct answer is $C=N$. This requires only basic grasping of relational family structure (understanding entity "Alice's sister", binding female attribute to Alice and realizing Alice and her sisters share same brothers). It does NOT require execution of any arithmetic or set operations, in contrast to AIW original. Should the issues with solving AIW original be rooted in handling basic family structure, we should see models also failing here. Again, we create AIW Light Family variations 1-4 by varying natural numbers $N,M$, such that correct responses $C$ are matched with AIW original variations 1-4. (Suppl. Tab. \ref{tab:aiw_light_family}). This gives following variations:

\begin{quote}
    Variation 1. Alice has 7 brothers and she also has 3 sisters. [Correct answer: 7] \\
    Variation 2. Alice has 4 sisters and she also has 3 brothers. [Correct answer: 3]
    Variation 3. Alice has 2 sisters and she also has 5 brothers.  [Correct answer: 5]
    Variation 4. Alice has 2 brothers and she also has 3 sisters.  [Correct answer: 2]

    Question: How many brothers does Alice's sister have?
\end{quote}

\textbf{AIW Light Arithmetic Total Girls}. AIW Light Arithmetic Total Girls has following problem template: \textit{\textbf{"Alice has $N$ brothers and she also has $M$ sisters. How many girls are there in total?"}}. Compared to AIW original, only question part is modified. To solve the problem, it is necessary to bind female attribute to Alice via the pronoun "she", to assign correct female attributes to the sisters and to execute the correct arithmetic sum operation adding all the obtained girls  - the correct answer is $C=M+1$ (note it is the same as for AIW original). This requires basic grasping of family structure (realizing who are the girls in the family) and selection and execution of elementary arithmetic sum operation. In contrast to AIW original, it does not require execution of set operations to properly assign Alice to sisters set. Should the issues with solving AIW original be rooted in binding correct sex attributes or counting total members of particular sex in family frame given its structure, we should see models also failing here. Again, we create variations 1-4 by varying natural numbers $N,M$, such that correct responses $C$ are matched with AIW original variations 1-4. (Suppl. Tab. \ref{tab:aiw_light_arithmetic_total_girls}). This gives following variations: 

\begin{quote}
    Variation 1. Alice has 6 sisters and she also has 3 brothers. [Correct answer: 7] \\
    Variation 2. Alice has 2 sisters and she also has 4 brothers. [Correct answer: 3]
    Variation 3. Alice has 4 sisters and she also has 1 brother.  [Correct answer: 5]
    Variation 4. Alice has 1 sister and she also has 4 brothers.  [Correct answer: 2]

    Question: How many girls are there in total?
\end{quote}






\subsection{Prompt types, providing input and response parsing}
\label{subsec:model_prompts_parsing}

\textbf{Model prompt types.} It is well known that so-called prompt engineering can heavily influence the model behavior and model response quality \cite{arora2022ask,wei2022chain,white2023prompt}. To check that our observations reflect model sensitivity to controlled, problem structure preserving variations in same manner independent of particular prompt type, we employed 3 various prompt types to provide model's input: STANDARD (prompt with instruction to format final answer output as a natural number), THINKING (prompt that in addition encourages thinking in spirit of CoT) and RESTRICTED (prompt with instruction to output nothing else but final answer as a natural number). THINKING v2 prompt type is a minor variation of THINKING type that just adds "step by step" after already existing "think carefully" phrasing (control experiments show that THINKING and THINKING v2 are equivalent in terms of observed performance, so we use both interchangeably, Suppl. Fig. \ref{subfig:aiw_thinking_v2_vs_thinking}). STANDARD and THINKING prompt types allow models to generate any text output before delivering the final answer, while RESTRICTED is used as control with restricted output to measure model behavior when the only output allowed is the final answer (see Fig. 1, Suppl. Fig. \ref{subfig:aiw_restricted_fluctuations} and Suppl. Tab. \ref{tab:aiw_prompt_types})

Furthermore, we make use of other prompt types (see Suppl. Sec.\ref{appendix:prompt types} for overview) to demonstrate various important properties and the different success or failure modes of the model behavior for the AIW problem. In those prompts, we re-use the main problem formulation as introduced in Sec. \ref{subsec:aiw_problem}, while adding various modifications. This allows us for instance to observe confabulations that contain clearly broken statements with reasoning-like convincing sound backing up wrong final answers, responses showing model overconfidence, or look into other function modes like in-context learning handling.

For each trial, models receive thus an input that has a form <\textit{instantiated-template}> <\textit{prompt-type}>, where <\textit{instantiated-template}> is template with substituted numbers instantiating one of problem variations 1-4 containing the question, followed by <\textit{prompt-type}> with task and output instructions corresponding to one of prompt types as described above.

\textbf{Parsing model responses.} To perform evaluations of model performance, it is necessary to parse and extract the final answer from the responses provided by the models. Each input to the model is combination of a AIW problem variation, followed by one of prompt types as described before. To keep the parsing procedure simple, each prompt type contains following output format instruction: \textit{"provide the final answer in following form: "\#\#\# Answer: ""}. We observed that all models we have chosen to test were able to follow such an instruction, providing a response that could be easily parsed. We also ran control experiments without such formatting instruction in the problem formulation, ensuring that behavior does not depend on it.




\subsection{Evaluating model responses}
\label{subsec:model_estimation_procedure}


The formatting instruction makes it possible to extract for each trial whether a model has provided a correct answer to the problem posed in the input. We can interpret then any number $n$ of collected responses as executing $n$ trials given a particular prompt for a given model ($n$ - number of Bernoulli trials), observing in each $i-$th trial a Bernoulli variable $X_i = \{0,1\}$. We interpret the number of correct responses $X=\sum_i X_i$ as random variable following a Beta-Binomial distribution with unknown probability $p$ of correct response that we also treat as random variable that comes from a Beta distribution, i.e. $p\sim Beta(\alpha,\beta)$, where $\alpha$ and $\beta$ are parameters of the Beta distribution. To obtain plots showing correct response ratios, we would like to estimate Beta distribution underlying $p$, and for that, we first estimate the mean of $p$ and its variance from the collected observations. To estimate $\hat{p}$, we use the formula for estimating the mean of $p$ for a binomial distribution: $\hat{p} = X/n$ (i.e. as a proportion of successes). We can report the estimate $\hat{p}$ as the estimate of the correct response rate of a given model and also, compare the correct response rates of various tested models. Moreover, we can estimate the variance of the probability of a correct response by using the following formula:
\begin{equation}
\textrm{var} \left(\frac {1} {n}\sum_{i=1}^{n} X_i \right)=\frac {1} {n^2}\sum_{i=1}^{n} \textrm{var}(X_i) = \frac {n \textrm{var}(X_i)} {n^2} = \frac{\textrm{var}(X_i)} {n} = \frac{p(1-p)} {n}
\end{equation}

The estimates of the variance and the standard deviation of $p$ can be thus obtained by using $\hat{p}$ as $\frac{\hat{p} (1-\hat{p})} {n}$ and $\sqrt{\frac{\hat{p} (1-\hat{p})} {n}}$   respectively. Using the estimated variance and mean of $p$, we can use the following relations for the variance: $ \left( \sigma^2=\frac {n\alpha \beta (\alpha +\beta +n)}{(\alpha +\beta )^{2}(\alpha +\beta +1)} \right)$ and the mean $ \left( \mu=\frac{\alpha} {\alpha + \beta} \right)$ in order to obtain $\alpha$ and $\beta$ parameters for the Beta distribution. To simulate data for the plots, we draw $N$ random samples corresponding to correct and incorrect responses using the estimated distribution of $p$ and obtain the plots showing performance on the task for various models of interest as a full distribution of the respective $p$.

\subsection{Selecting models for evaluation and conducting experiments}
\label{subsec:model_selection_experiments}


We are interested in testing state-of-the-art models that claim strong function, especially in generalization and reasoning, backed up by high scores shown on standardized benchmarks that are assumed to measure problem solving capabilities and specifically, generalization and reasoning. We therefore select models widely known and used in the ML community that also appear in the top rankings of the popular LLM leaderboards, like openLLM leaderboard by HuggingFace or ELO leaderboard by LMsys. We provide the overview of the selected models in Suppl. Tab. \ref{tab:aiw_models_overview} and list in Suppl. Tab. \ref{tab:bench_aiw} the corresponding standardized benchmarks where they obtain strong scores. Whenever possible, we were choosing models of various scales from the same family, ranging from small to large, in order to see how the capabilities to solve the posed task may vary with scale.

We expose selected SOTA LLMs, including advanced models at largest scales (see Suppl. Tab. \ref{tab:aiw_models_overview}) to AIW problem variations 1-4 (Suppl. Tab. \ref{tab:aiw_prompt_types}) and AIW Light control problems (Suppl. Tab. \ref{tab:aiw_light_arithmetic_siblings}, \ref{tab:aiw_light_family}, \ref{tab:aiw_light_arithmetic_total_girls}), and other problem versions using different prompt types as described above. For each combination of model, AIW problem variation and prompt type, at least 30 trials are collected to compute both correct response rates within each variation and average correct response rates across variations 1-4, Suppl. Fig. \ref{fig:num_of_responses_aiw}. For details on correct response rates estimation procedure, see Sec. \ref{subsec:model_estimation_procedure}

We use hosting platforms that offer API access or local deployment via vLLM~\cite{vllm_2024} for testing the models, and automatize the procedure by scripting the routines necessary to prompt models with our prompts set. The routines are simple and can be used by anybody with access to the APIs (we used liteLLM and TogetherAI for our experiments) or to locally hosted models to reproduce and verify our results. We protocol all the data from interactions with the models to enable community checking. We release all the collected raw response data, correct response rates estimates and routines used to conduct experiments as open-source for reproducibility and further usage.

\section{Results}
\label{sec:results}

\begin{figure}[tbh!]
    \centering
    \includegraphics[width=1.0\textwidth]{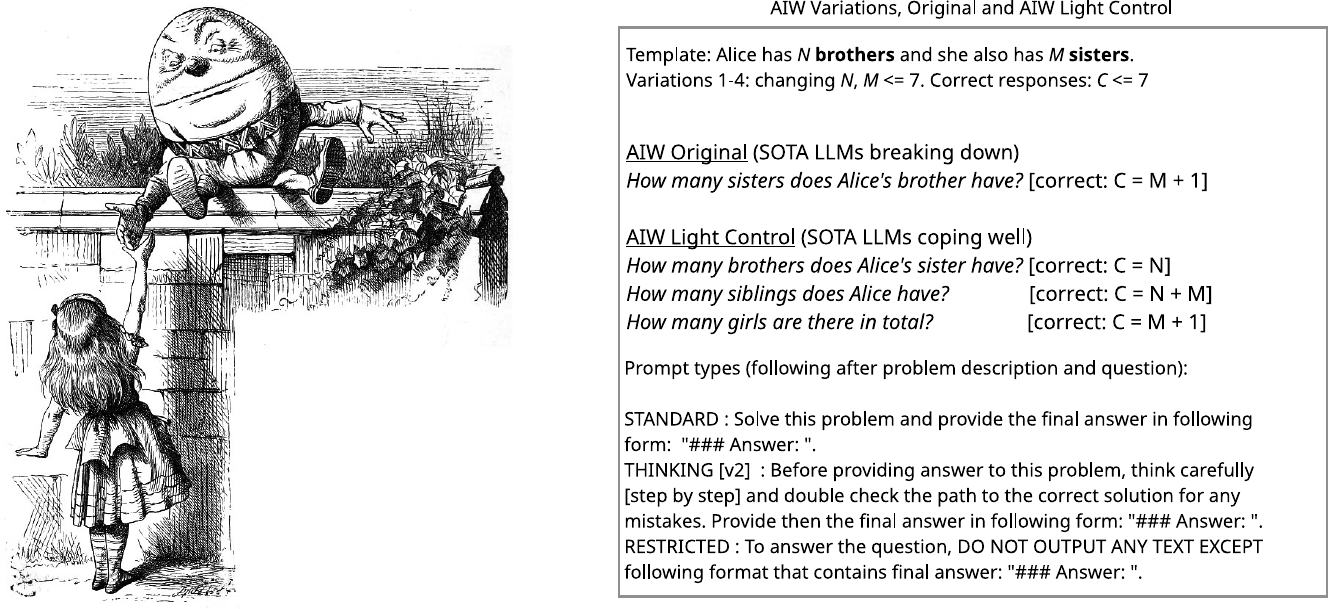}
    \caption{Alice is reasoning: will it break? Illustration of Humpty Dumpty from Through the Looking Glass~\cite{carroll1871through}, by John Tenniel, 1871. Source: Wikipedia.}
    \label{fig:alice_reasons_humpty}
\end{figure}


\subsection{Humpty Dumpty sat on a wall: breakdown of SOTA LLMs on the simple AIW problem}
\label{subsec:humpty}

\begin{figure}
    \centering
    \includegraphics[width=\textwidth]{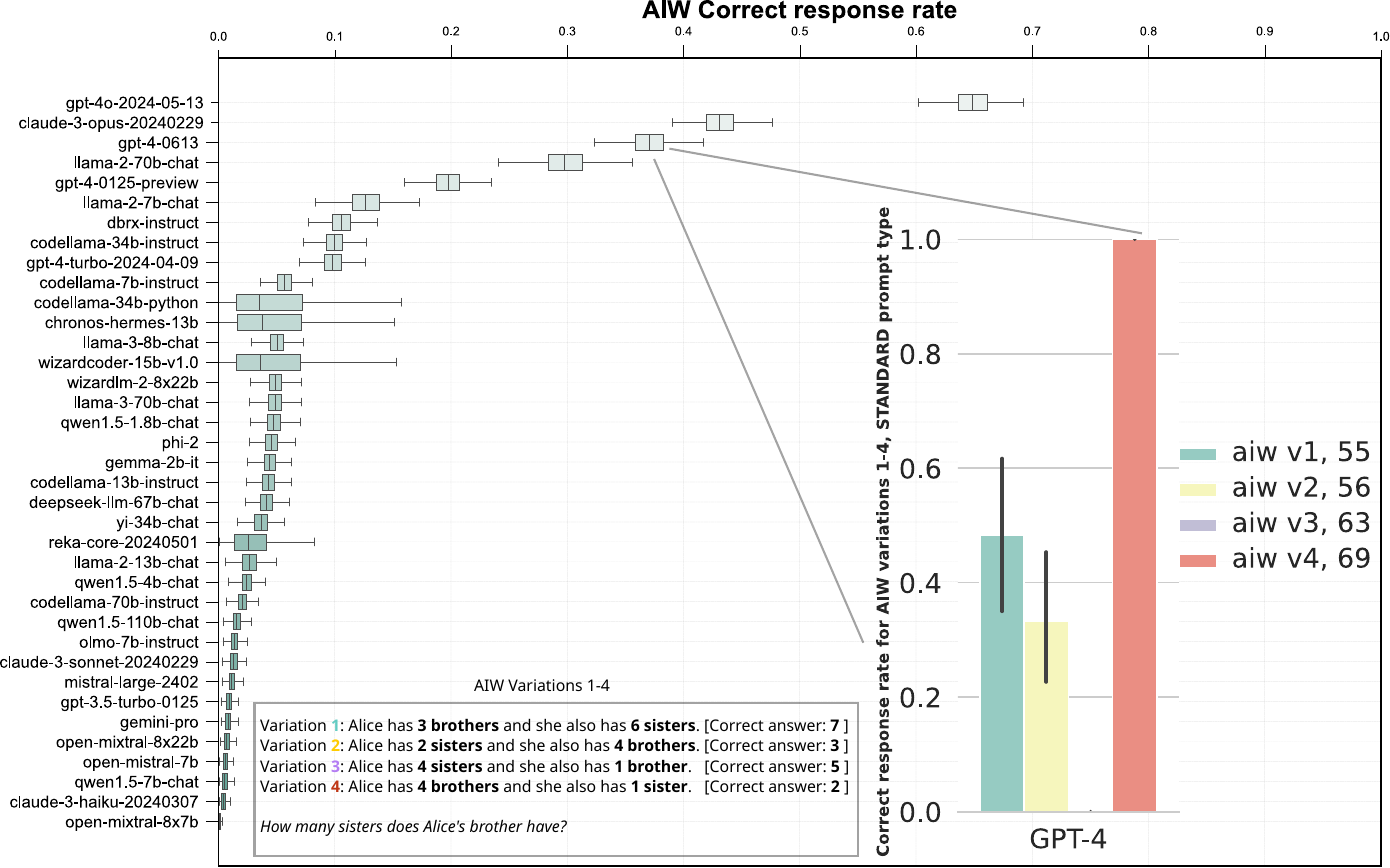}
    \vspace{-0.5cm}
    \caption{Collapse of SOTA LLMs on AIW problem. (\textbf{main}) Models with non-zero AIW correct response rate, average over STANDARD, THINKING, RESTRICTED prompt types and AIW variations 1-4. Omitted models score 0. \textbf{(inlay}) Strong fluctuations on AIW variations 1-4, despite problem structure and difficulty remaining entirely unchanged across variations. Note that overall correct response rate averaged across variations does not reveal these fluctuations (shown on example of GPT-4). Numbers in the legend are prompt IDs (Suppl. Tab. \ref{tab:aiw_prompt_types})}
    \label{fig:main_correct_response_rate}
    \vspace{-0.3cm}
\end{figure}

\textbf{AIW reveals severe generalization and reasoning deficits in SOTA LLMs}. Following the procedures described in Sec \ref{sec:methods_experiment}, we expose the selected models that claim strong function and reasoning capabilities (Suppl.  Tab. \ref{tab:aiw_models_overview}) and measure their correct response rate performance across and for each AIW variations 1-4 using various prompt types, executing $>30$ trials for each combination (see also Suppl. Tab. \ref{tab:aiw_prompt_types} and Suppl. Fig. \ref{fig:num_of_responses_aiw}). The results suggest that confronted with the AIW problem, models suffer a severe function breakdown. This breakdown has two main manifestations: 

\textbf{1. Low correct response rates.} Despite evident problem's simplicity, many models are not able to deliver a single correct response, and the majority stay well below correct response rate of $p = 0.2$.  We summarize the main results in the Fig. \ref{fig:main_correct_response_rate}. The only major exceptions from the observation of very low correct response rates are the largest scale closed models GPT-4 and Claude 3 Opus. These two models at largest scales obtain correct response rates well above $p = 0.3$, leaving the remaining large and smaller scales open-weights (e.g., Mistral-7B, Mixtral, Qwen, Command R+, and Dbrx Instruct) and closed-weights models (e.g., Gemini Pro, Mistral Large) far behind. Remarkably, many models that claim high scores on standardized benchmarks, show very low correct response rates close to 0, eg. Llama-3-8B, Mixtral-8x22B, Qwen1.5-110B, or exhibit even complete breakdown on AIW with correct response rate of zero across all variations, eg Command R+ or Qwen1.5-72B (Suppl. Tab. \ref{tab:bench_aiw})

The results presented in the Fig. \ref{fig:main_correct_response_rate} show estimates for correct response rates averaged across RESTRICTED, STANDARD and THINKING prompt types (Suppl. Tab. \ref{tab:aiw_prompt_types}, prompt IDs provided for reproducibility; Suppl. Fig. \ref{fig:aiw_all} with models scoring 0). RESTRICTED prompt type was used as further control that forces models into short outputs, restricting the compute for providing a solution and thus serving as low baseline for the performance (see Suppl. Sec. \ref{appendix:aiw_problem} and Suppl. Fig. \ref{fig:aiw_restricted}). Among the 4 models that are able to cross $p = 0.3$, two clear winners are the GPT-4o ($p = 0.649$) and Claude 3 Opus ($p = 0.431$). The only open-weights model in this set of better performers is the rather older Llama-2 70B Chat ($p = 0.3$). For better performers, when inspecting the responses with correct final answers, we see also correct reasoning backing up the final answers. For the poor performing models with low correct response rates, by inspecting those rare responses with correct answers we also in some cases still can see correct reasoning. In the poor performers, among the responses with a correct final answer we see however often responses where final answer, after careful inspection, turns out to be an accident of executing entirely wrong reasoning with various mistakes leading coincidentally to the final output number corresponding to the right answer. Such responses are encountered in models with low correct performance rates ($p < 0.4$) (see Suppl. Sec. \ref{appendix:responses} for response examples), and we correct via manual inspection the status of correct response for such cases.

\begin{figure}[bt!]
    \centering
    \includegraphics[width=1.0\textwidth]{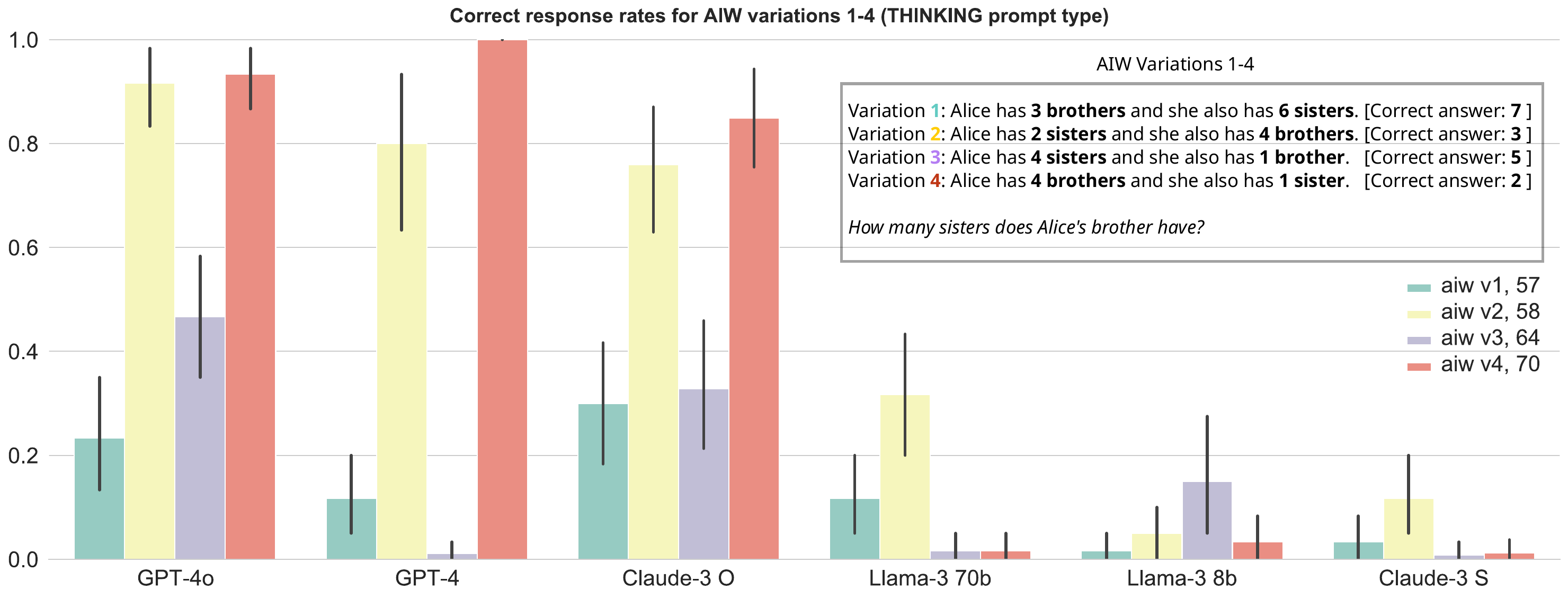}
    \caption{Strong fluctuations across AIW problem variations, THINKING prompt. For better performers, eg GPT-4o/4 and Claude Opus 3, correct response rates vary strongly from close to 1 to close to 0, despite AIW variations not affecting problem structure or difficulty (a color per each variation 1-4). This shows clear lack of model robustness, revealing generalization and basic reasoning deficits.}
    \label{fig:aiw_original_fluctuations}
\end{figure}

\textbf{2. Strong performance fluctuations across irrelevant AIW problem variations.} Importantly, we also observe strong fluctuation of correct response rates across AIW variations 1-4 as introduced in Sec. \ref{sec:methods_experiment}. Such fluctuations strongly affect better performers with higher average correct response rates like GPT-4/4o and Claude 3 Opus (as poor performers have often correct response rates across all variations close to 0, so that no room for fluctuations exist on that very low performance level). As shown in the Fig. \ref{fig:main_correct_response_rate} (inlay) for the STANDARD and Fig. \ref{fig:aiw_original_fluctuations} for the THINKING prompt type, the correct response rates can fluctuate between being close to 1 to being close to 0, depending on AIW variation. Remarkable is that such fluctuations appear despite AIW variations being all instances of the very same simple problem, as changes in numbers used across AIW variations do not change either the problem structure or its difficulty at all. This lack of robustness on such a simple problem hints on severe deficits in generalization. The strong fluctuations across variations appear independent of employed prompt types (Fig. \ref{fig:aiw_original_fluctuations_expl}), while correct response rate averaged across all variations also varies across prompt types, showing in addition expected prompt type dependency (Suppl. Fig. \ref{fig:aiw_restricted}, \ref{fig:aiw_thinking_v2_comparison})

\subsubsection{Control experiments using AIW Light problems}
\label{subsec:aiw_light_problems_exp_results}

As outlined in Sec. \ref{subsec:aiw_problem}, we make use of control experiments to rule out that the observed breakdown might be due to failures in executing low level operations specific to AIW, eg. parsing natural language, numbers, arithmetic operations, set operations, binding attributes to entities (eg female attribute via "she" pronoun to Alice), handling basic family structure and so on. Observed strong fluctuations across variations in better performing models make it already unlikely that same low level operation (or same set of low level operations) necessary to solve AIW are broken - if that would be the case, such failures would be manifested to equal extent across all variations, as they pose the very same problem requiring same operations to be performed, so that performance would suffer in same manner independent of variations. Still, to obtain direct evidence, we design control experiments such that they require low level operations necessary to solve AIW problem while also keeping problem structure close enough to the original (Sec. \ref{subsec:aiw_problem}).

In all following experiments, for each AIW variation, 60 trials were executed to estimate correct response rate and its variance.

\begin{figure}[t!]
    \centering
    \includegraphics[width=1.0\textwidth]{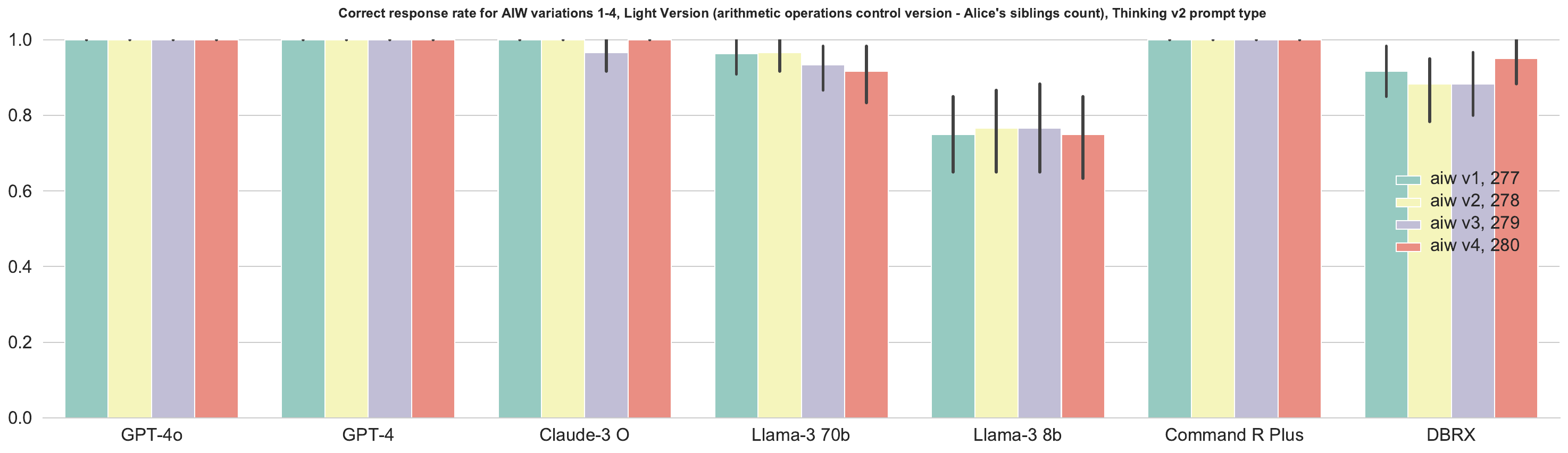}
    \caption{Correct response rates across AIW Light Arithmetic Siblings control problem variations 1-4 (THINKING v2 prompt type). Strong performance is observed across problem variations (a color per each variation 1-4; prompt IDs in the legend, Suppl. Tab. \ref{tab:aiw_light_arithmetic_siblings}). Models that entirely collapse on AIW, like Command R Plus and Dbrx Instruct, are clearly able to solve this version, with correct response rates going up to 1 or close to 1 across all problem variations. This shows that executing arithmetic operations or handling basic family setting is not an issue for the tested models.}
    \label{fig:aiw_light_arithmetic_siblings_thinking}
\end{figure}

\textbf{AIW Light Arithmetic Siblings}. We show tested models' performance in Fig. \ref{fig:aiw_light_arithmetic_siblings_thinking}. While all tested models clearly have struggled with AIW original (Fig. \ref{fig:main_correct_response_rate}, Suppl. Fig. \ref{fig:aiw_all}), we observe them successfully solving AIW Light Arithmetic Siblings. Correct response rates go high up close to 1 for most tested models across all variations 1-4. This is also the case for the models that show very low correct response rates close to 0 or 0 on AIW original, like Command R+ or Dbrx Instruct (Suppl. Fig. \ref{fig:aiw_all}, Suppl. Tab. \ref{tab:bench_aiw}). Importantly, strong fluctuations we observe across variations on AIW original (Fig.  \ref{fig:main_correct_response_rate}, \ref{fig:aiw_original_fluctuations}) disappear entirely. This clearly demonstrates that models neither struggle with basic grasping of relational family structure - eg., realizing Alice's siblings are her sisters and brothers, nor with selection and execution of elementary arithmetic sum operation.

\begin{figure}[t!]
    \centering
    \includegraphics[width=1.0\textwidth]{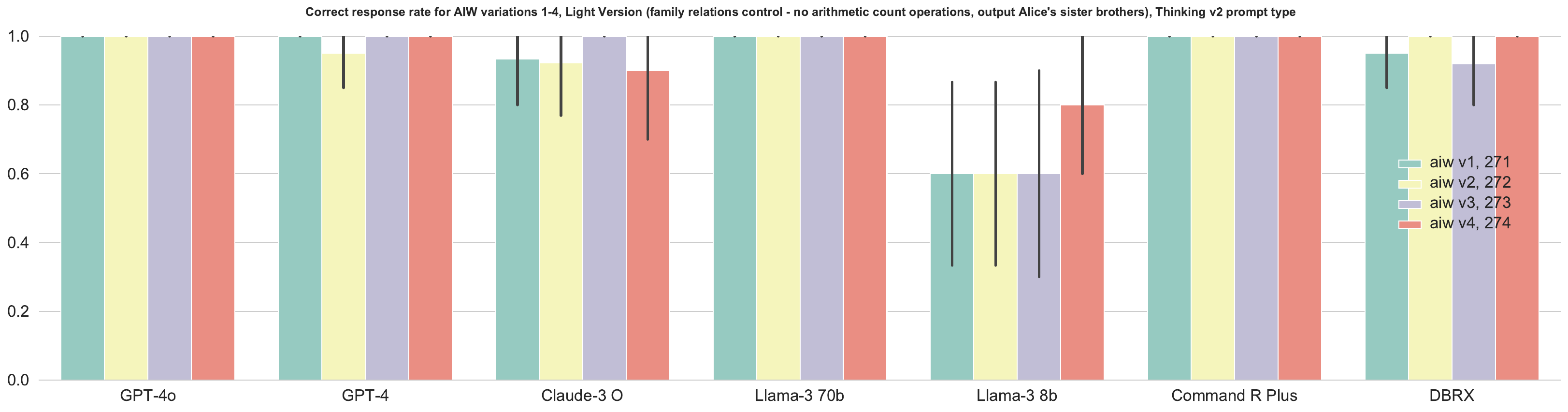}
    \caption{Correct response rates across AIW Light Family control problem variations 1-4 (THINKING v2 prompt type). Strong performance is observed across problem variations (a color per each variation 1-4). Models that entirely collapse on AIW, like Command R Plus and Dbrx Instruct, are clearly able to solve this version, with correct response rates going up to 1 or close to 1 across all problem variations. This shows that handling basic family relations and binding sex attributes to entities via pronouns is not an issue for the tested models.}
    \label{fig:aiw_light_family_thinking}
    \vspace{-0.3cm}
\end{figure}

\textbf{AIW Light Family}. We show tested models' performance in Fig. \ref{fig:aiw_light_family_thinking}. Also here we observe all the tested models that are struggling with AIW original successfully solving AIW Light Family. Correct response rates go high up close to 1 for most tested models across all variations 1-4. This is also the case for the models that show very low correct response rates close to 0 or 0 on AIW original. like Command R+ or Dbrx Instruct (Suppl. Fig. \ref{fig:aiw_all} \& Tab. \ref{tab:bench_aiw}). Also strong fluctuations that we observe across variations on AIW original (Fig.  \ref{fig:main_correct_response_rate}, \ref{fig:aiw_original_fluctuations}) disappear. This clearly demonstrates that models handle well basic grasping of relational family structure - understanding entity "Alice's sister", binding female attribute to Alice (which allows to determine correct number of brothers) and realizing Alice and her sisters share same brothers.

\begin{figure}[t!]
    \centering
    \includegraphics[width=1.0\textwidth]{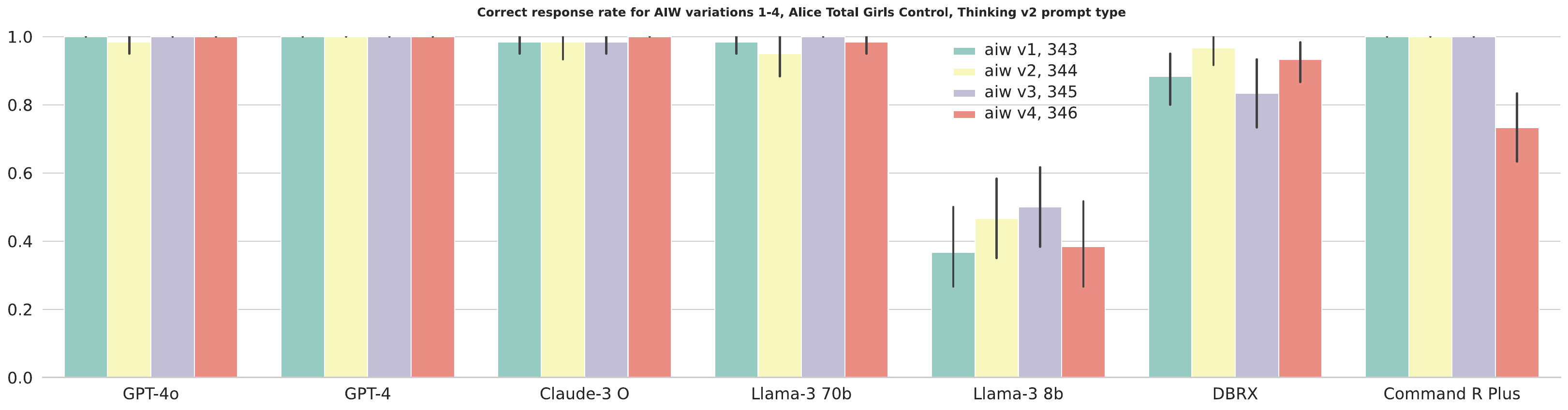}
    \caption{Correct response rates across AIW Light Arithmetic Total Girls control problem variations 1-4 (THINKING v2 prompt type). Strong performance is observed across problem variations (a color per each variation 1-4; prompt IDs in the legend, Suppl. Tab. \ref{tab:aiw_light_arithmetic_total_girls}). Models that entirely collapse on AIW, like Command R Plus and Dbrx Instruct, are clearly able to solve this version, with correct response rates going up to 1 or close to 1 across all problem variations. This rules out that either binding of female attributes to Alice and the sisters entities or selection and execution of arithmetic operations necessary to count total females is an issue for the tested models.}
    \label{fig:aiw_light_arithmetic_girls_thinking}
\end{figure}

\textbf{AIW Light Arithmetic Total Girls}. We show tested models' performance in Fig. \ref{fig:aiw_light_arithmetic_girls_thinking}. Again, we observe also here strong performance for all tested models that clearly have struggled with AIW original. Correct response rates go high up close to 1 for most tested models across all variations 1-4. This is also the case for the models that show very low correct response rates close to 0 or 0 on AIW original. like Command R+ or Dbrx Instruct. Also strong fluctuations that we observe across variations on AIW original (Fig.  \ref{fig:main_correct_response_rate}, \ref{fig:aiw_original_fluctuations}) are gone. This clearly demonstrates that models successfully cope with binding female attribute to entity of Alice, handle assignment of correct female attributes to the sisters and select and execute the correct arithmetic sum operation adding all the girls together.

From these control experiments, we are thus able to obtain strong evidence that all tested models do not suffer from low-level issues with tokenization and natural language or natural numbers parsing and can handle well basic family relations structure, binding of attributes to entities and selection and execution of elementary arithmetic operations necessary to solve AIW problem. This further strengthen the hypothesis that observed failures and strong fluctuations in all tested SOTA models on AIW problem (Fig.  \ref{fig:main_correct_response_rate}, \ref{fig:aiw_original_fluctuations}) are rooted in problem unspecific, generic deficits in generalization and basic reasoning about problem structure.

\subsubsection{Discrepancies between claimed strong function and observed breakdown}
\label{subsec:discrepancies}

\begin{figure}[t!]
    \centering
    \includegraphics[width=0.8\textwidth]{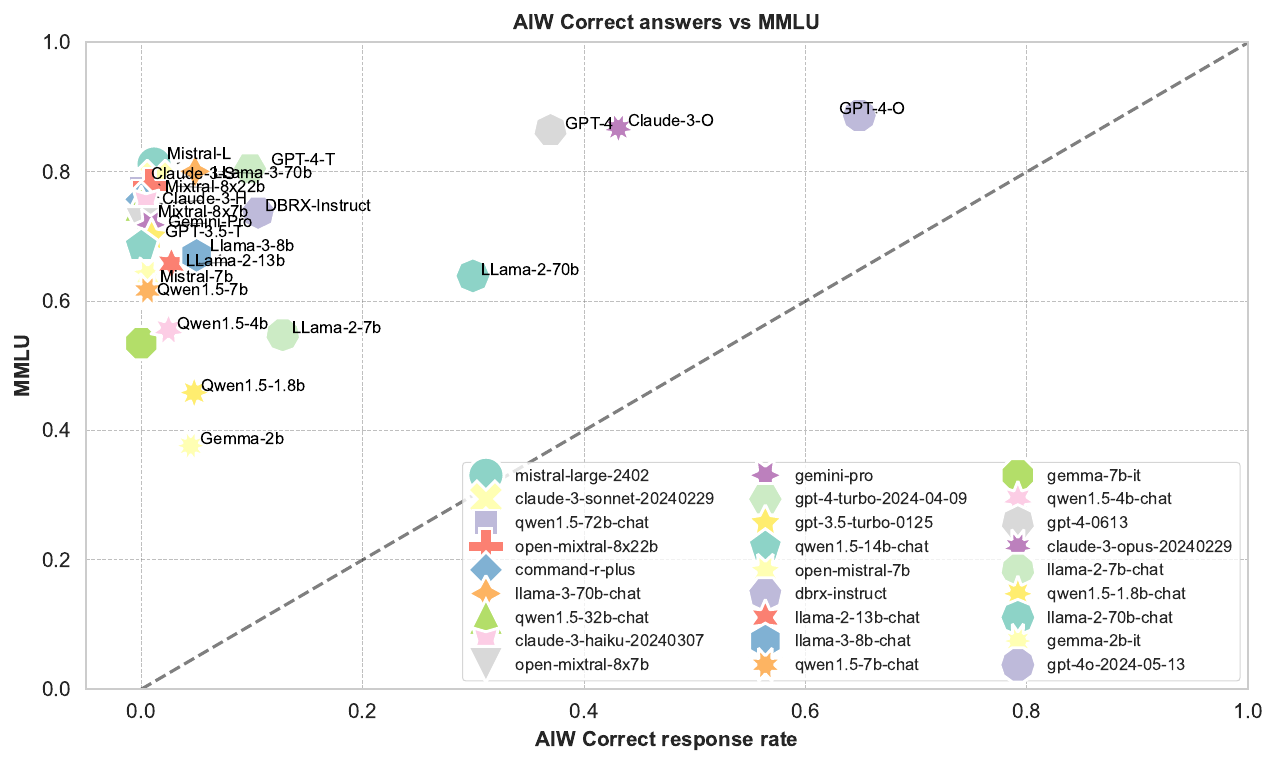}
    \vspace{-0.2cm}
    \caption{Failure of standardized benchmark MMLU to properly reflect and compare model basic reasoning capabilities as shown by strong discrepancy between AIW correct response rate vs MMLU average score. Many models, eg. Command R+, score 0 on AIW, but have high MMLU score. }
    \label{fig:mmlu_benchmark_vs_aiw}
    \vspace{-0.5cm}
\end{figure}

\textbf{Standardized benchmarks failure}. We observe failure of standardized reasoning benchmarks to properly reflect generalization and basic reasoning skills of SOTA LLMs by noting significant disparity between the model's performance on the AIW problem and the scores on conventional standardized benchmarks. All of the tested models report high scores on various standardized benchmarks that claim to test problem solving via reasoning, e.g. MMLU, ARC, Hellaswag. Our observations of SOTA models breaking down on the simple AIW problem hint that the benchmarks do not reflect deficits in generalization and basic reasoning of those models properly. We visualize this failure by plotting scores tested models obtain on wide-spread and accepted standardized benchmarks like MMLU versus the performance we observe on our proposed AIW problem. As strikingly evident from  Fig. \ref{fig:mmlu_benchmark_vs_aiw}, there is a strong mismatch between high scores on MMLU reported by the models and the correct response rates they obtain on AIW. This mismatch and lack of differentiation makes it impossible for a given model to predict from its score on MMLU whether it will suffer breakdown on a simple problem like AIW, making the score unreliable for measuring core capabilities. Also model ranking fails, as models with similarly high MMLU scores claiming similar function level can have dramatic difference on simple AIW problem. For instance, models like Llama-3-70B, Mistral Large or GPT-4-Turbo come close with their MMLU score to GPT-4/Claude 3 Opus, hinting comparable capabilities, while settled in the crowd of high MMLU - low AIW score region (left upper part of Fig. \ref{fig:mmlu_benchmark_vs_aiw}) with most other models that achieve very low AIW performance close to 0. This also demonstrates that MMLU, while containing problems of arguably higher difficulty, does not properly reflect deficits in basic model function, as revealed by much simpler AIW problem.  For similar evidence on other standardized benchmarks, see Suppl. Sec. \ref{appendix:benchs_failure}


\textbf{Go small, go home: breakdown of smaller scale models}. The few models capable of showing significant non-zero correct response rate for the AIW problem are residing on the larger scales. GPT-4o/4 and Claude 3 Opus have unknown scales, it is however reasonable to assume the model scale is well beyond 70B params and the tokens scale is well beyond 2T tokens. Observing the performance on the AIW problem across various models, we see evidence that in general, smaller scale models (known to have been overtrained on large token budgets of $>2T$ tokens) that have quite high scores on standardized reasoning benchmarks, some coming close to larger scale ones, suffer severe collapse on the AIW problem. No small scale model can even remotely approach the performance shown by better performers residing at larger scales which, despite observed strong fluctuations on problem irrelevant variations, do show significant correct response rates on some AIW problem variations (GPT-4o/4, Claude 3 Opus), while smaller scale models drop close to zero, staying far below $p=0.1$ (with the only exception of Llama 2 7B, which upon inspection reveals that its average correct response rate is substantially elevated due to one strong outlier on a single AIW variation, with correct response rate on others being close to 0 or 0, which cannot be taken as a signature of strong function and is a sign of severe lack of robustness, see also Fig. \ref{fig:llama_comparison_robustness}). An important note here is that while probability of random guess on AIW problem is not straight forward to calculate, as the answer might be any natural number, it is conceivable that any model capable of language parsing from the input can probably easily infer from the problem formulation that response should be a rather low natural number without ever grasping any essentials about actual problem structure. That might already restrict the candidate numbers to a narrow interval, eg 0-10, which would provide a random guess probability of about $p=0.1$. Following that consideration, any correct rate response that is around or below $p=0.1$ might considered just random guessing that does not require any reasoning. All the tested smaller scale models ($<13$B) would then fall into the category of random guessing according to Fig. \ref{fig:main_correct_response_rate},  Suppl. Fig. \ref{fig:aiw_all}. For further evidence on effect of scale, see Sec. \ref{subsec:model_comparison_ranking} and Fig. \ref{fig:llama_scaling}, \ref{fig:qwen_scaling}.




\subsection{Curiouser and curiouser}
\label{subsec:curious}

Following our observations of breakdown of SOTA LLM models that claim strong function described in Sec. \ref{subsec:humpty}, we investigated various properties and modes of the observed failures, reporting here the ones we find most remarkable. Investigating the AIW problem further, we find various other formulations that show same breakdown characteristics for all SOTA LLMs. We observe same lack of robustness on structure and difficulty preserving problem variations using various AIW extensions which also abandon original family problem framing, providing further evidence that observed breakdown of zero-shot generalization is of generic character and is not unique for the specific AIW problem setting, as also corroborated by control experiments in Sec. \ref{subsec:aiw_light_problems_exp_results}.



\begin{figure}[tb!]
    \centering
    \begin{subfigure}{1.0\textwidth}
        \includegraphics[width=1.0\textwidth]{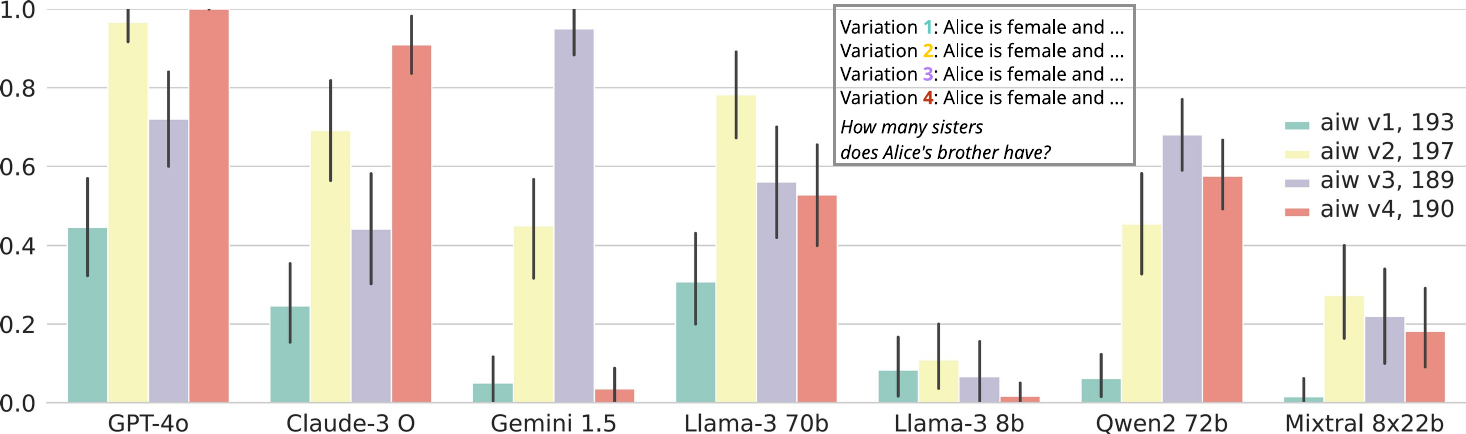}
        \caption{Correct response rates for AIW Alice Female Boost, THINKING v2 prompt type}
        \label{subfig:aiw_thinkingv2_alice_female_boost}
    \end{subfigure}
    \begin{subfigure}{1.0\textwidth}
        \includegraphics[width=1.0\textwidth]{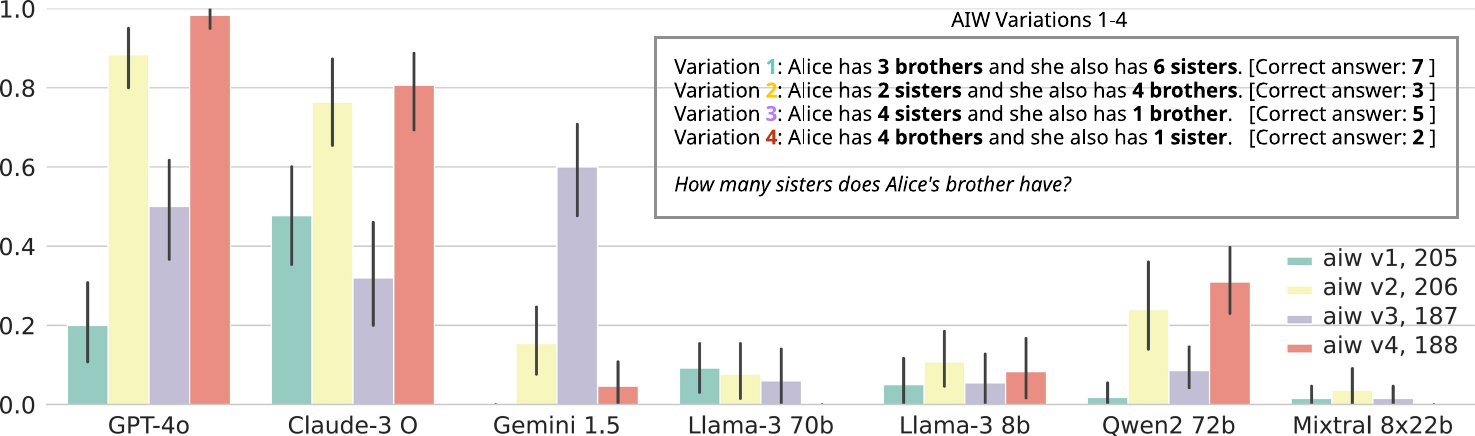}
        \caption{Correct response rates for AIW original, THINKING v2 prompt type}
        \label{subfig:aiw_thinkingv2}
    \end{subfigure}    
    \caption{Altering model performance by fully redundant information. Adding fully redundant information "Alice is female" leads to increase of average correct response rates in \textbf{(a)} compared to AIW original \textbf{(b)} (see also Suppl. Fig. \ref{fig:aiw_original_alice_female_power_boost_average_correct_response_rates}). For some models, eg Llama 3 70B or Qwen 2 72B, this boost via redundant info is especially pronounced and happens across all variations, resulting in clear overall improvement from \textbf{(b)} to \textbf{(a)}. Note performance of a smaller scale model like Llama 3 8B stays low close to 0. Strong fluctuations across variations 1-4 persist. This again shows lack of model robustness, hinting on severe generalization and basic reasoning deficits.}
    \label{fig:aiw_alice_thinkingv2_original_female_power_boost_fluctuations}
\end{figure}

\textbf{Performance fluctuations by adding fully redundant information: Alice female power boost}. One clear signature of generalisation and reasoning breakdown are the strong fluctuations we observe across AIW problem variations 1-4 that differ only in instantiated numbers (Fig. \ref{fig:aiw_original_fluctuations}). We investigate a further AIW problem version by adding \textit{"Alice is female"} to the original AIW problem formulation (see Suppl. Tab. \ref{tab:aiw_female_boost_variation_types} and Suppl. Sec. \ref{appendix:subsec:aiw_female_power_boost}). This is a fully redundant information, as Alice's gender is already unambiguously specified by the "she" pronoun used in original AIW problem. As evident from Fig. \ref{fig:aiw_alice_thinkingv2_original_female_power_boost_fluctuations} and Suppl. Fig. \ref{fig:aiw_original_alice_female_power_boost_average_correct_response_rates}, the average correct response rates are increasing, despite the provided "female boost" information being entirely redundant and not revealing anything new necessary for AIW problem solution. Altering performance by fully redundant information that should not affect problem solving reveals again deficits in generalization and basic reasoning. While average correct response rates increase, the strong fluctuations across AIW variations 1-4 remain (Fig. \ref{subfig:aiw_thinkingv2_alice_female_boost}). For instance, GPT-4o has on AIW variations 2,4 correct response rate close to 1, while dropping heavily for AIW variations 1,3, showing same lack of robustness despite the average boost.

\begin{figure}
    \centering
    \includegraphics[width=\textwidth]{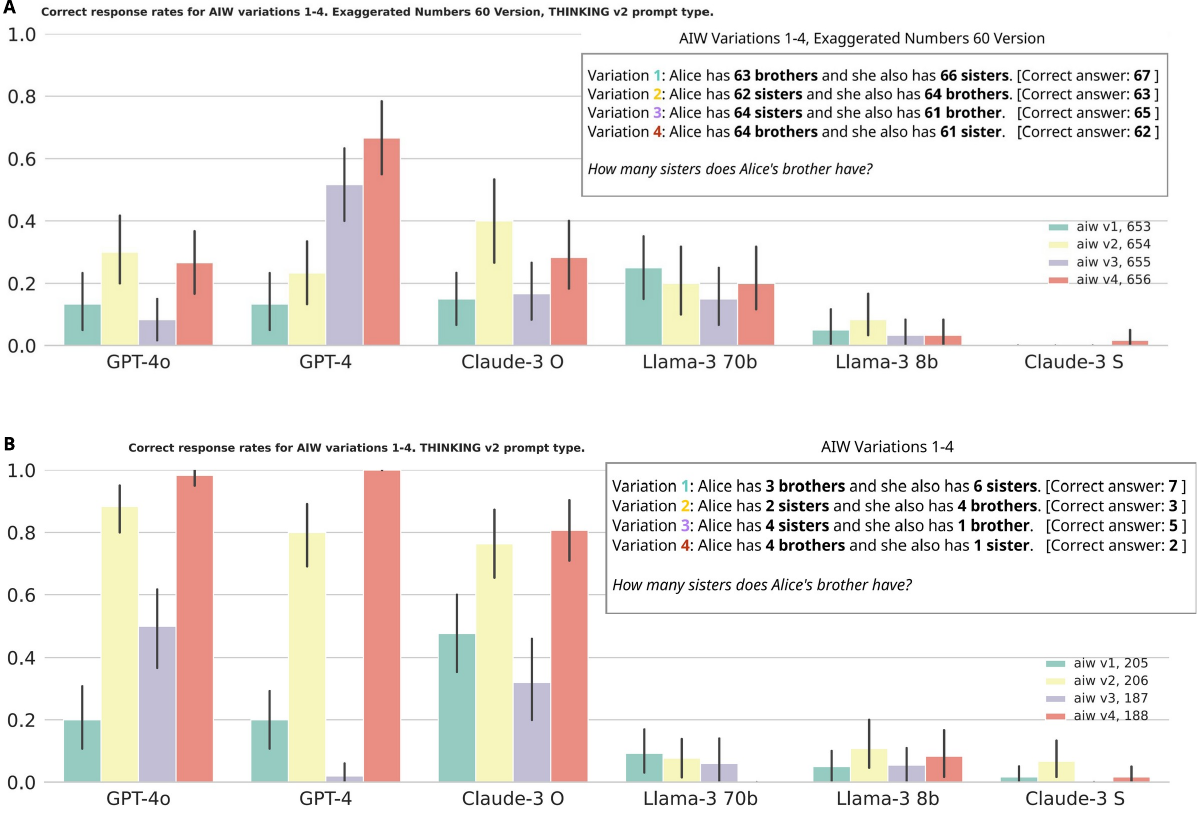}
    \caption{Strong fluctuations on AIW problem variations (a color per each variation 1-4) for AIW version with exaggerated numbers are used (\textbf{A}), created by adding 60 offset to numbers used in AIW original (\textbf{B}), using same THINKING v2 prompt type. Main pattern of model behavior remains the same - exhibiting strong fluctuations across variations and having breakdowns on some variations with low correct response rates. Variations correspond to different instantiations of numbers $N,M$ for brothers and sisters in the same problem template. Varying numbers should not affect problem solution at all, being a "natural" variation that does not affect problem structure and its difficulty. However, correct response rate varies strongly depending on the variation. Lack of robustness to irrelevant variations of such a simple problem points to severe generalization deficits. Correct response rates seem also to vary between the exaggerated number AIW version (\textbf{A}) and the original (\textbf{B}). Average across all 4 variations, it is lower for exaggerated numbers for most models. For Llama 3 70B as an exception, there is a overall increase, while performance stays still poor. Changes in correct response rates induced by mere number modification that leaves problem structure unchanged again provide evidence for lack of model robustness and hint on generic generalization deficits.}
    \label{fig:aiw_exaggerated_vs_orig_fluctuations}
    \vspace{-0.3cm}
\end{figure}

%
\textbf{Performance fluctuations by changing magnitude of instantiated numbers.}. Variations of numbers $N,M$ for brothers and sisters in AIW problem were so far deliberately chosen to be in the range of common sense, avoiding exceedingly large numbers. For the formal mathematical structure of the problem though, magnitude of numbers does not pose any change. We conduct experiment using AIW version with exaggerated numbers, created by adding offset of $60$ to numbers $N,M$ used in the AIW original. We see the same breakdown pattern as in AIW original - models exhibit strong fluctuations across problem structure and difficulty preserving variations, or break down to low performance level close to 0 across all variations (Fig. \ref{fig:aiw_exaggerated_vs_orig_fluctuations}) We also observe differences in correct response rates between the exaggerated number AIW version Fig. \ref{fig:aiw_exaggerated_vs_orig_fluctuations}(\textbf{A}) and the original Fig. \ref{fig:aiw_exaggerated_vs_orig_fluctuations} (\textbf{B}). As the number magnitude modification leaves the problem structure unchanged, these performance fluctuations provide further evidence for lack of model robustness to problem irrelevant perturbations and hint on generic generalization deficits.

\subsubsection{Persisting fluctuations and breakdown on various AIW versions}
\label{subsec:aiw_versions}


To confirm that the same observations hold for other problems of related kind, we construct further problem templates which pose problems of similar or higher difficulty than AIW original. Using same experimental procedure to create variations of these problem versions, we observe the same pattern as for the AIW original, especially the strong fluctuations across variations, confirming the existence of the same generalization deficits for further problem examples. The problems are as following:

\textbf{AIW extended (AIW Ext.)} In this problem version, Alice and Bob are employed as two entities, keeping the template close to the AIW original.  The resulting variations (for the version, where question is posed explicitly for Alice - AIW Ext Explicit - prompt IDs 264, 266, 268, 270) are as following:

\begin{quote}
    Alice and Bob are sister and brother.
    
    Variation 1. Alice has 3 sisters and Bob has 6 brothers. [Correct answer: 7] \\
    Variation 2. Alice has 2 sisters and Bob has 2 brothers. [Correct answer: 3] \\
    Variation 3. Alice has 1 sister and Bob has 4 brothers.  [Correct answer: 5] \\
    Variation 4. Alice has 3 sisters and Bob has 1 brother.   [Correct answer: 2]

    Question: How many brothers does Alice have?
\end{quote}

\textbf{AIW Friends} In this problem version, we abandon the family frame setting. Instead, we use male and female friends in problem formulation. Note the problem structure is still related to AIW original ( brothers and sisters are male and female siblings). We use an additional condition to ensure there is no common sense ambiguity in this problem version. The resulting variations are as following:

\begin{customquote}{0.2cm}
    Variation 1. Alice has 3 male friends and she also has 6 female friends. [Correct answer: 7] \\
    Variation 2. Alice has 2 female friends and she also has 4 male friends. [Correct answer: 3] \\
    Variation 3. Alice has 4 female friends and she also has 1 male friend. [Correct answer: 5] \\
    Variation 4. Alice has 4 male friends and she also has 1 female friend. [Correct answer: 2]

    All mentioned persons are friends with each other and have no other friends aside. \\
    Question: How many female friends does male friend of Alice have?
\end{customquote}

These two problem versions, AIW Ext and AIW Friends, have similar problem structure and difficulty to AIW original. They thus can be also seen as control to ensure our observations are consistent among various problem formulations that have similar problem difficulty (in contrast to AIW Light control problems which are designed to test low level operations involved, such that their templates pose problems of lower difficulty than AIW original).

\begin{figure}
    \centering
    \includegraphics[width=\textwidth]{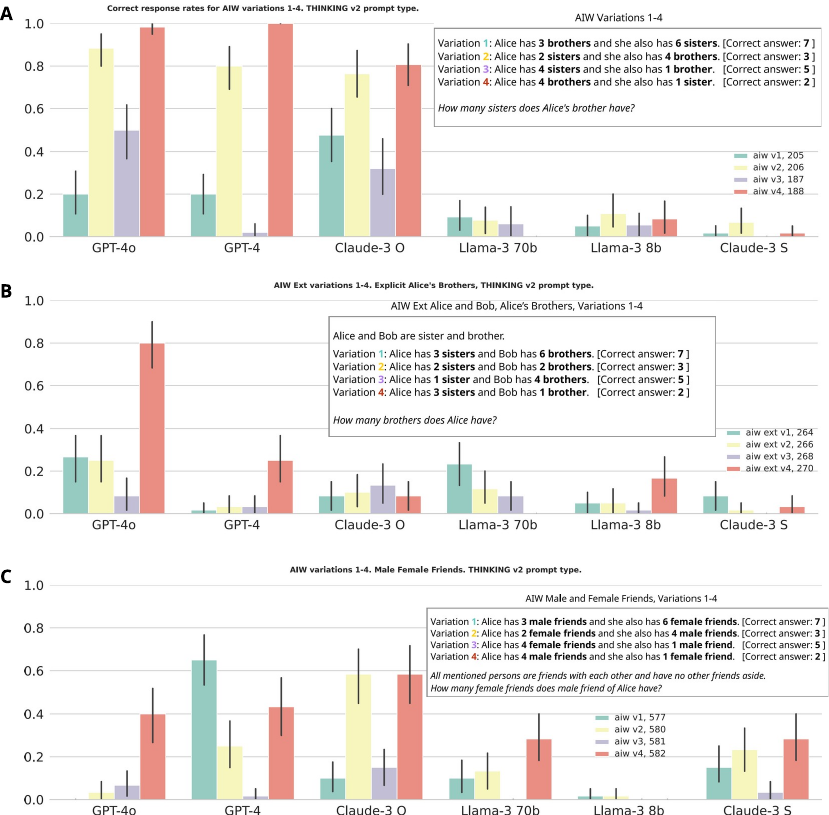}
    \caption{Strong fluctuations on problem variations (a color per each variation 1-4) comparing AIW original (\textbf{A}) with two further versions, AIW Extended (AIW Ext (\textbf{B}) and AIW Friends (\textbf{C}), using THINKING v2 prompt type. Overall, observed pattern remains the same - better performers (GPT-4/4o, Claude 3 Opus) that manage to get occasionally higher correct response rates ($p>0.3$) on some variations show strong fluctuations, with correct response rates dropping strongly on other variations, despite variations 1-4 leaving problem structure and difficulty unchanged by merely instantiating different numbers $N,M$ into problem templates. AIW Ext (\textbf{B}) and AIW Friends (\textbf{C}) show overall altered correct response rates (eg, for GPT-4/4o, Claude 3 Opus) compared to AIW original (\textbf{A})). Strong fluctuations are common phenomenon across the problem versions, again hinting on generic generalization deficits that are independent of specific problem formulation. Differences in performance between the problem versions provide additional evidence for lack of model robustness, as problem structure and difficulty is highly similar between the versions.}
    \label{fig:aiw_problem_versions}
    \vspace{-0.3cm}
\end{figure}


Results of experiments with AIW Ext and AIW Friends and comparison to AIW original are shown in Fig. \ref{fig:aiw_problem_versions}. We again see rather low correct response rates on the one hand, with better performers exhibiting  strong fluctuations on problem variations. While the breakdown pattern for each problem version might depend on a given model (eg., note Claude 3 Opus overall breakdown in correct response rates on AIW Ext in Fig. \ref{fig:aiw_problem_versions} (\textbf{B}), while performing comparably, though with lower correct response rates, on AIW Friends  Fig. \ref{fig:aiw_problem_versions} (\textbf{C}) ), the observed overall breakdown pattern is the same, confirming for tested SOTA LLMs same kind of severe lack of robustness and generalization deficit as observed on AIW original. Additional evidence for lack of model robustness and generalization deficit comes from the observation that performance also varies across problem versions, despite those being very similar in their structure and difficulty. Operations required to solve these various problem versions overlap strongly, although there are also differences. For instance, in AIW Ext  \ref{fig:aiw_problem_versions} (\textbf{B})), operations are required to handle Alice and Bob entities separately, to assign both properly to sisters and brothers sets, while in AIW original, only Alice has to be handled. On the other hand, AIW original has additional operations not required by AIW Ext. It requires binding of Alice via "she" pronoun to female gender before she can be properly assigned to sisters set and resolving the entity of Alice's brother in the posed question - which AIW Ext in turn does not require. In AIW Friends \ref{fig:aiw_problem_versions} (\textbf{C})), attributes female/male and friend have to be treated separately which requires additional binding operations. This is in contrast to sister and brother subsuming both the gender attribute (male, female) and the relationship (sibling) in one, which does not require additional binding. The problem structure and difficulty among the problem versions are thus similar enough so that if generalization were intact, such mild differences would not matter and no such pronounced difference in performance would have appeared, in contrast to our observations.



\subsubsection{Harder problem versions: further breakdown of advanced LLMs}
\label{subsec:harder_aiw_versions}

To test what happens if the simple AIW problem is further extended towards a substantially harder difficulty level, providing further challenge for the tested models, we construct further problem versions of higher difficulty while keeping same relational logic structure appeal and conduct experiments as described in following.


\textbf{AIW+ problem}. We constructed an AIW+ problem that features additional hierarchy and distractors when describing relational family structure (see Suppl. Sec \ref{appendix:prompt types} for full formulation). AIW+ problem template has following form: \textit{"Alice has \textbf{M} sisters and \textbf{N} brothers in total. Her mother has 2 brothers. She also has 1 sister who does not have children and who has \textbf{X} nephews and nieces in total. Alice's father has 2 sisters. He also has a brother who has \textbf{Y} nephews and nieces in total, and who also has \textbf{K} [sons/daughters]. How many cousins does Alice's sister have?"}. 

The solution to AIW+ problem is harder to obtain than the solution to AIW original with its much simpler structure. Solving AIW+ requires taking different paternal sides, that of mother and father, and calculating the number of cousins, taking care of subtracting Alice and her siblings, and summing up the total number of cousins from both sides. Still, this problem is arguably far from olympiad or university graduate level, as it requires just using provided numbers and careful execution of elementary arithmetic operations on straightforward path to solution. The correct solution is given by $C = (X - (M + N + 1)) + (Y - (M + N + 1) + K)$.  We follow again our approach to create variations by instantiating numbers $N,M,X,Y,K$ in problem template to obtain problem instances of same problem structure and difficulty (see also Suppl. Tab. \ref{tab:aiw_plus_prompt_types}).


\begin{figure}[tb!]
    \centering
    \includegraphics[width=\textwidth]{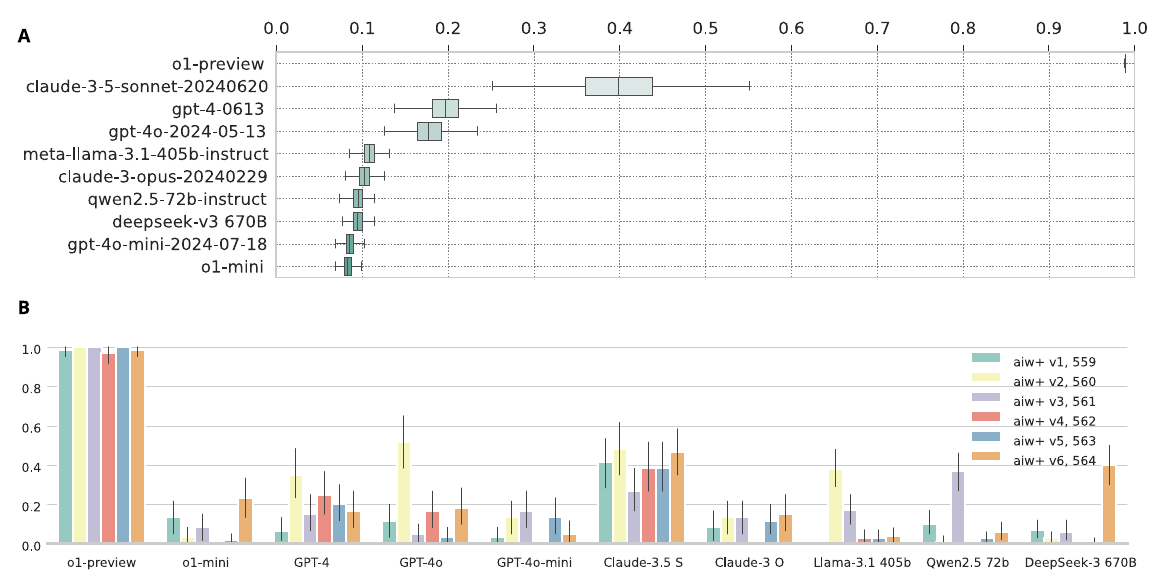}
    \caption{AIW+ correct response rates averaged over variations 1-6 (\textbf{A}) and fluctuations across variations (\textbf{B}). Most tested models undergo further collapse compared to AIW original. o1 preview as clear exception shows robust ability to solve AIW+ across all variations without fluctuations. o1-mini on contrary collapses close to 0, also showing fluctuations (eg on variation 6 vs others). AIW+ was made intentionally harder than simple AIW. However, models claiming strong function should be able to solve it, as it does not involve any higher level logic or math.}
    \label{fig:aiw_plus_all}
    \vspace{-0.3cm}
\end{figure}

We show a full example of correct solution with instantiated numbers for AIW+ variation 1 ($N=M=1, X=6, Y=5, K=2$), which corresponds to problem instance as following: 

\begin{customquote}{0.2cm}
    Variation 1. Alice has 1 sister and 1 brother in total. Her mother has 2 brothers. She also has 1 sister who does not have children and who has 6 nephews and nieces in total. Alice's father has 2 sisters. He also has a brother who has 5 nephews and nieces in total, and who also has 2 sons.  [Correct answer: 7] \\
    
    Question: How many cousins does Alice's sister have?
\end{customquote}

Here, we have on the mother side: 6 (total nephews and nieces)  - 3 (Alice and her siblings) = 3 cousins; on the father side: 5 (total nephews and nieces) + 2 (sons of the father's brother) - 3 (Alice and her siblings) = 4 cousins; summing up 3 + 4 = 7 cousins which Alice and any of her siblings share.


Exposing models to AIW+ following same methodology from Sec. \ref{sec:methods_experiment} (we add two more variations, so that AIW+ has variations 1-6), we observe further, even stronger collapse of performance also for those advanced models that were showing significant correct response rates for AIW problem (Fig. \ref{fig:aiw_plus_all}) For instance, for GPT-4/4o and Claude 3 Opus overall correct response rate averaged across variations stays below $p<0.2$ (Fig. \ref{fig:aiw_plus_all} (\textbf{A})). Large-scale open weight SOTA models Llama 3.1 405B, DeepSeek-v3 671B and Qwen 2.5 72B are settled around $p=0.1$ and below. Recent Claude 3.5 Sonnet is an outlier scoring higher up close to $p=0.4$, without showing strong fluctuations we usually observe (Fig. \ref{fig:aiw_plus_all} (\textbf{B}); see however Sec. on Claude 3.5 Sonnet breakdown - its performance on some versions of AIW might be due to exposure to AIW tasks in post training, as it appeared after first version of our public AIW release). To show that problem can be successfully handled, we also test here o1-preview that comes from the recent generation of reasoning models (which we treat as an exception; LLMs are conventionally understood as models pre-trained in purely autoregressive manner. It is still unknown for o1 class of models whether RL on unknown amounts of synthetic data presumably of math and logic type is executed during pre-training or is rather a part of post-training). o1-preview is a clear exception and has robust performance close to 1 across all AIW+ variations. Remarkably, o1-mini coming presumably from the same model class does not show same robustness - its performance is comparable to standard LLM generation far below o1-preview, settled close to 0 (Fig. \ref{fig:aiw_plus_all} (\textbf{A})) and exhibiting fluctuations as usually observed in our study (Fig. \ref{fig:aiw_plus_all} (\textbf{B})).

\begin{figure}[tb!]
    \centering
    \includegraphics[width=\textwidth]{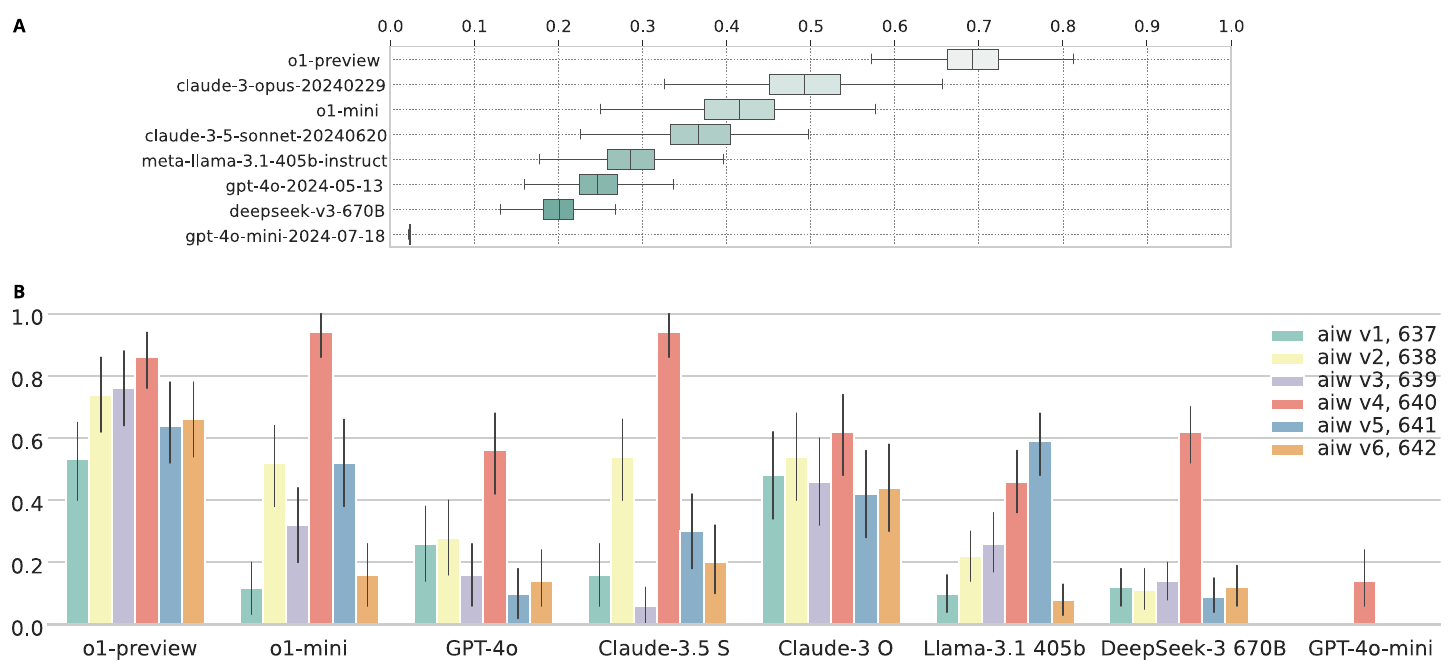}
    \caption{AIW Colleagues Circles correct response rates averaged over variations 1-6 (\textbf{A}) and fluctuations across variations (\textbf{B}). Same breakdown pattern is evident in the tested models - either overall low correct response rates or strong performance fluctuations across problem structure and difficulty preserving variations as observed on AIW original. Also o1 preview exhibits such fluctuations (eg. variation 4 vs variation 1, 5 or 6), providing evidence that also strongest models from o1 class are not robust, pointing to generalization deficits. o1-mini, which collapsed strongly on AIW+, exhibits here fluctuations much stronger than o1 preview, again providing evidence that o1-mini is much weaker than o1 preview, contrary to claims relying on standardized benchmarks. Strong fluctuations affect most models that obtain significant correct response rates on at least one of the variations, eg. GPT-4o, Claude 3.5 Sonnet, Llama 3.1 405B, DeepSeek v3 671B. GPT-4o-mini collapses entirely.  AIW Colleague Circles is harder than simple AIW, but it does not involve any high difficulty logic or math. Models claiming strong function on level of graduate or olympiad math tasks should be able to solve it without fluctuations with higher correct response rates across variations.}
    \label{fig:aiw_cc_average_fluctuations}
    \vspace{-0.3cm}
\end{figure}

\textbf{AIW Colleague Circles}. Evidencing strong performance of o1-preview on AIW+, we design a further problem with harder difficulty level than AIW original to test whether we can observe same breakdown patterns on the strongest model. AIW like problems can be understood as problems on graphs, featuring entities, properties and relationships that define sets. To increase problem difficulty, we depart from simple connectivity that was characteristic of the AIW original and introduce circles of colleagues where all-to-all connectivity defines a circles, while some entities have connections to outside, make those entities hubs connecting circles.

AIW Colleague Circles problem template has following form: \textit{"Alice has 3 male colleagues and she also has \textbf{M} female colleagues in total. All these mentioned persons in the circle around Alice are colleagues of each other. Bob has 2 female colleagues and 1 male colleague in total. All these mentioned persons in the circle around Bob are colleagues of each other. The people in the circle around Bob do not have other colleagues aside - with the only exception of Matilda. She is colleague of Bob, being part of Bob's circle, and she is also colleague of Alice, being part of Alice's circle. All the mentioned persons have no colleagues beyond the already described group of people. How many female colleagues does Matilda have?"}.

Again, the solution to AIW Colleague Circles problem is harder to obtain than the solution to AIW original with its much simpler structure. Same as AIW+, the problem is though still far from olympiad or graduate level, which advanced SOTA LLMs often claim to master robustly. The correct solution is given by $C = M + 1$.  We follow again our approach to create variations by instantiating number $M$ in problem template to obtain problem instances of same problem structure and difficulty (note we use only alteration of a single number here to create variations 1-6, corresponding to instantiating $M=6,2,4,1,3,5$ with corresponding correct answers $C=7,3,5,2,4,6$, see also Suppl. Tab. \ref{tab:aiw_circles_prompt_types}).

We conduct experiments to test models' sensitivity to perturbations of AIW Colleague Circle problem. We again observe the already familiar breakdown pattern for all tested models - including o1-preview (Fig. \ref{fig:aiw_cc_average_fluctuations}). While we saw o1-preview solving all the posed AIW versions so far robustly with high correct response rates close to 1 across all variations, here we observe rates below (p=0.8) and more importantly, strong fluctuations, eg. $p<0.6$ on variation 1 vs $p>0.8$ on variation 4 (Fig. \ref{fig:aiw_cc_average_fluctuations} (\textbf{B})). With only one single natural number varied in the problem template, this provides evidence that also larger scale models of o1 class are not robust to problem structure and difficulty preserving variations, hinting on generalization deficits in rather simple setting (as compared to claimed capabilities to robustly solve problems in graduate and olympiad level math setting, which are far above the level here). Interestingly, o1-mini does not undergo such a strong collapse as on AIW+ (Fig. \ref{fig:aiw_plus_all}), which might be further evidence for AIW Colleague Circles difficulty being rather moderate, as we observe o1-mini usually lagging far behind o1-preview on AIW problem versions tested here.

\subsection{Characteristics of the observed breakdown}
\label{subsec:breakdown_characteristics}


\subsubsection{Dominance of wrong responses assessed by frequency distribution of natural numbers on output.}
\label{subsec:frequency_response_distribution_results}

To shed more light on modes of correct or wrong responses provided by the models when confronted with AIW problem variations, we show here frequency distribution for natural numbers on the output for AIW variations with higher and lower correct response rates.

\begin{figure}[t!]
    \centering
    \includegraphics[width=0.9\textwidth]{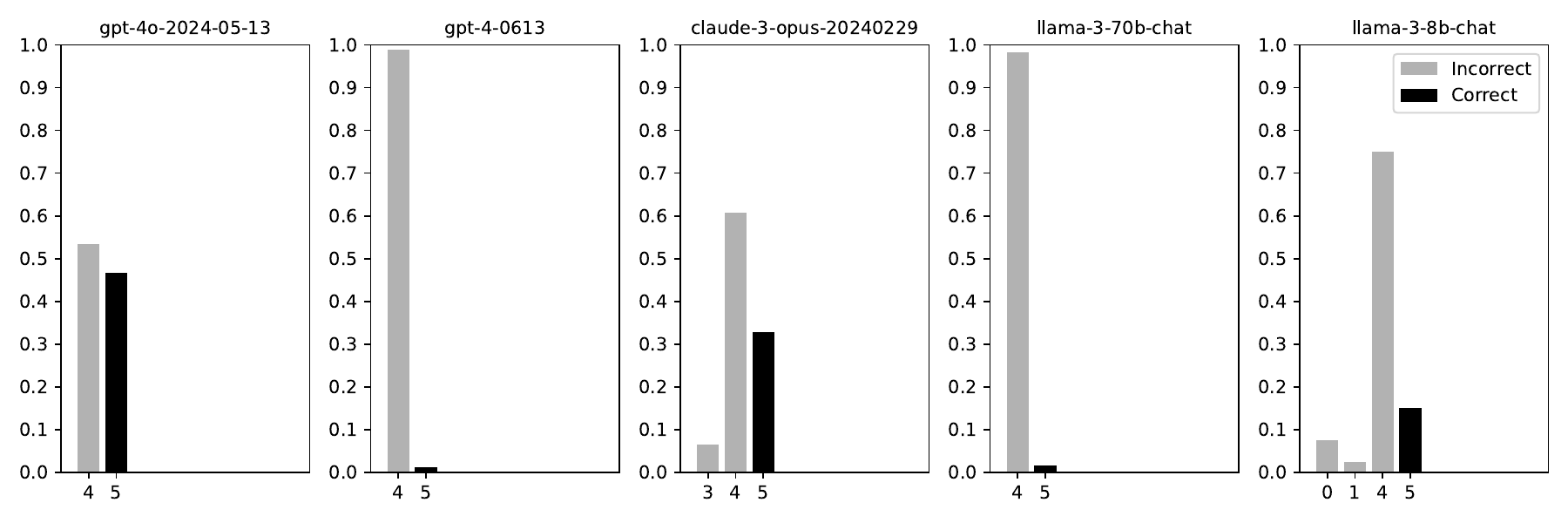}
    \caption{Frequency distribution of output numbers in models' responses. Shown are numerical outputs for AIW Variation 3, THINKING prompt type (prompt ID 64), that has correct answer C=M+1=5, with M=4 number of sisters of Alice. For this AIW variation, models have low performance (Fig. \ref{fig:aiw_original_fluctuations}). Correspondingly, peaks are on the dominant wrong response, R=M=4. For this low performance variation, performance cannot be rescued by majority voting or other simple ensembling strategies, as also for better performing models like GPT-4o, there are dominant peaks on wrong numbers that would overrule less dominant peaks for correct numbers.  Weaker models, eg Llama 3 8B, show also broader distribution. Distributions were computed over 60 trials executed for each model, taken from original collected responses data.}
    \label{fig:aiw_numbers_id_64_thinking}
\end{figure}

\begin{figure}[t!]
    \centering
    \includegraphics[width=0.9\textwidth]{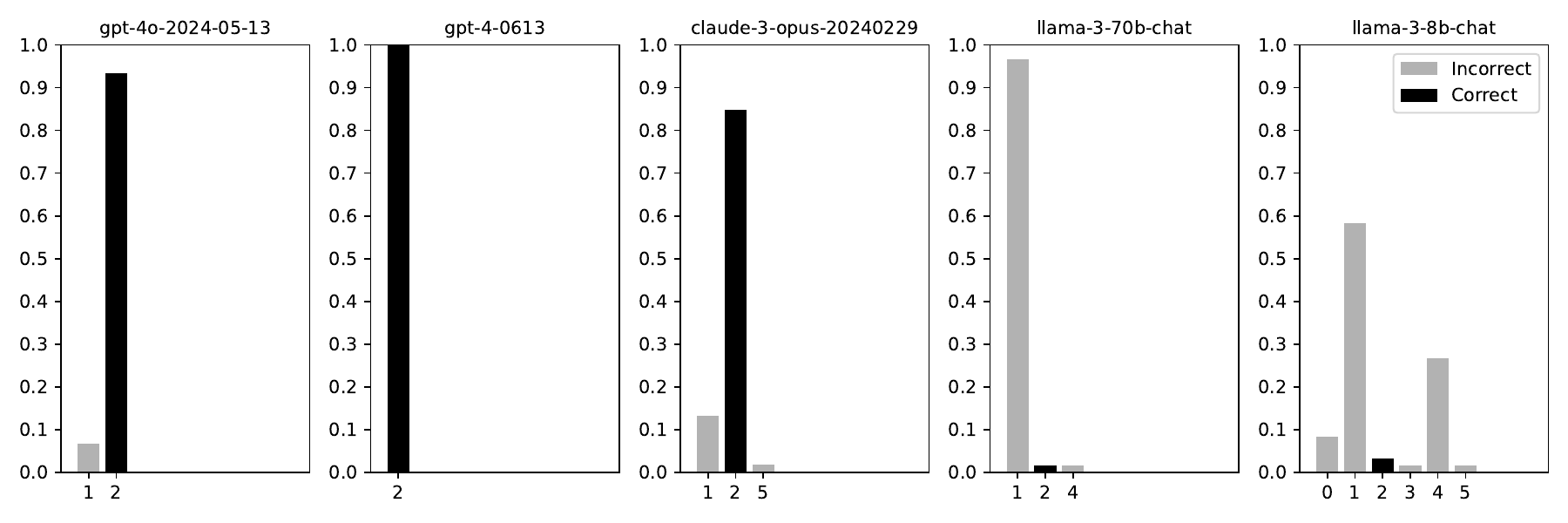}
    \caption{Frequency distribution of output numbers in models' responses. Shown are numerical outputs for AIW Variation 4, THINKING prompt type (prompt ID 70), that has correct answer C=M+1=2, with M=1 number of sisters of Alice. For this AIW variation, models have higher performance (Fig. \ref{fig:aiw_original_fluctuations}). Correspondingly, peaks for better performing models (eg GPT-4o, GPT-4, Claude Opus 3) are on the dominant correct response, R=M+1=2. For models with worse performance, peaks  are on the dominant wrong response, R=M=1. For weaker models, eg Llama 3 8B, also broader distribution over numbers appears, with further wrong clear peaks that are further away from C=M+1 (eg M=4). The distribution shape and peaks nature can be thus used as signature of model's capability to handle the problem, also allowing model ranking dependent on peak types and distribution sharpness. Distributions were computed over 60 trials executed for each model, taken from original collected responses data.}
    \label{fig:aiw_numbers_id_70_thinking}
\end{figure}

As evident from the plots, in higher performance AIW variations (Fig. \ref{fig:aiw_numbers_id_70_thinking}), dominants peaks are often positioned on correct answer C=M+1, while for lower performance AIW variations (Fig. \ref{fig:aiw_numbers_id_64_thinking}), dominant peaks fall on wrong answer M. Further, for weaker models, distribution broadens, covering more numbers (eg in Llama 3 8b), while for better performers, responses concentrate on M and M+1, peaking on correct or wrong answer on depending on AIW variation. Remarkably, for lower performance AIW variations (Fig. \ref{fig:aiw_numbers_id_64_thinking}), performance cannot be rescued by major voting or by similar ensemble like strategies, as peaks on wrong response numbers dominate clearly peaks on numbers for correct responses, which would still correspond to committing wrong answer when performing majority voting.

\begin{figure}[t!]
    \centering
    \includegraphics[width=0.9\textwidth]{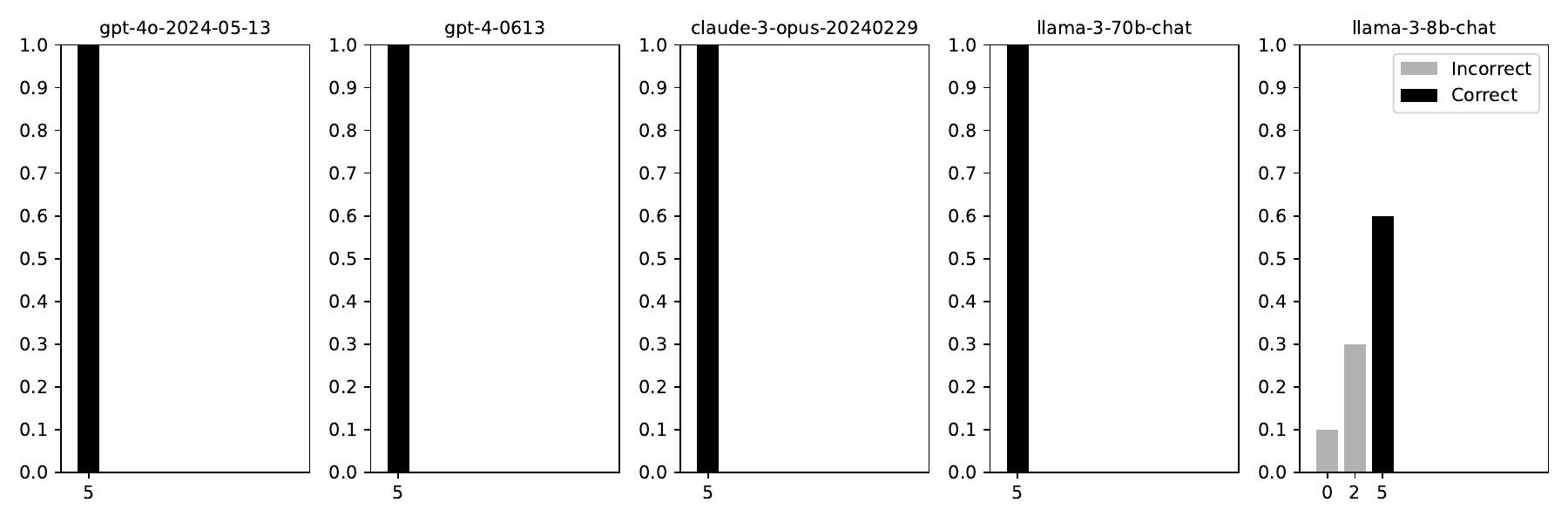}
    \caption{Frequency distribution of output numbers in models' responses. Shown are numerical outputs for AIW Light Family, Variation 3, THINKING prompt type (prompt ID 273), that has correct answer C=5 (number of Alice's brothers). For this AIW Light version, all models have high performance. Correspondingly, peaks are on the dominant correct response, R=5. However also here, weaker models like Llama 3 8B show broader distribution with non-vanishing peaks besides the correct response (eg R=0, R=2) hinting on their weaker capabilities to deal robustly with the problem. Distributions were computed over 60 trials executed for each model.}
    \label{fig:aiw_numbers_id_273_thinking}
\end{figure}

\begin{figure}[t!]
    \centering
    \includegraphics[width=0.9\textwidth]{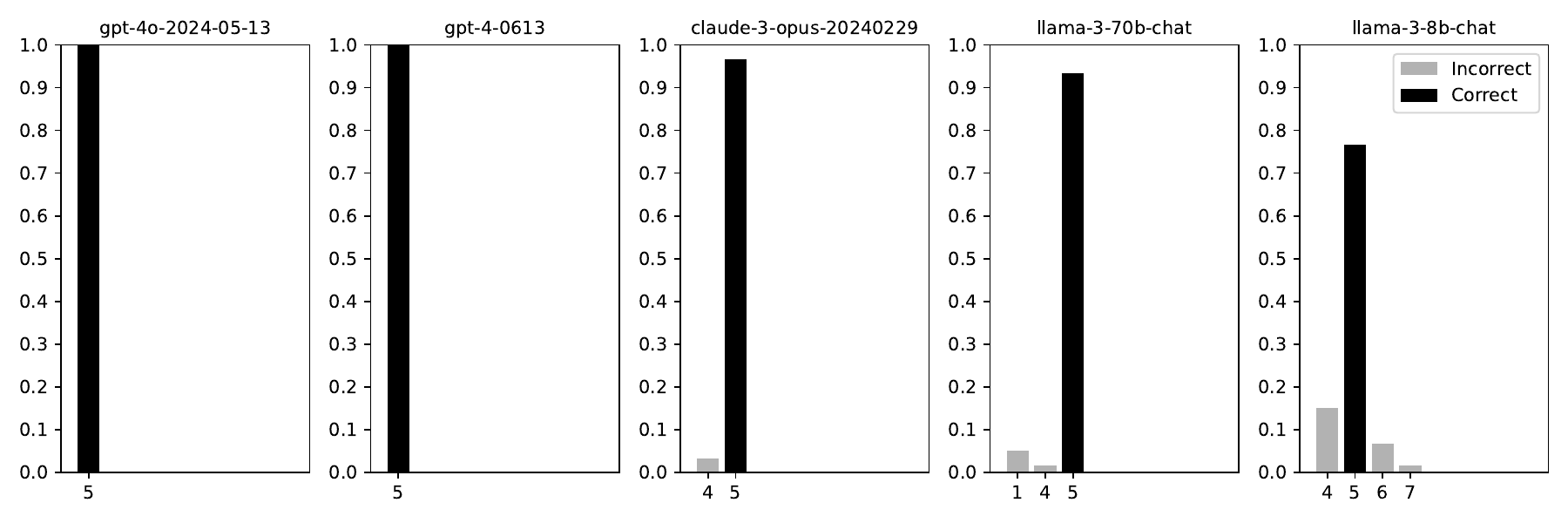}
    \caption{Frequency distribution of output numbers in models' responses. Shown are numerical outputs for AIW Light Arithmetic, Variation 3, THINKING prompt type (prompt ID 279), that has correct answer C=5 (total number of Alice's siblings). For this AIW Light version, all models have high performance. Correspondingly, peaks are on the dominant correct response, R=5. However also here, weaker models like Llama 3 8B show broader distribution with non-vanishing peaks besides the correct response (eg R=4, R=6) hinting on their weaker capabilities to deal robustly with the problem. Distributions were computed over 60 trials executed for each model.}
    \label{fig:aiw_numbers_id_279_thinking}
\end{figure}

For the AIW Light problem versions used in control experiments, we observe as expected clear dominant peaks on the numbers corresponding for correct responses across all tested models (Fig. \ref{fig:aiw_numbers_id_273_thinking}, \ref{fig:aiw_numbers_id_279_thinking}), as AIW Light problems are successfully solved across all their variations.

We note that distribution characteristics, eg concentration on numbers around the correct answer, height of the peaks, can be a further signature that reflects model's capability to handle the problem. More capable models retain dominant peaks on number corresponding to correct answer with smaller peaks on neighboring numbers, while weak models have large peaks on numbers corresponding to wrong answers or in general broad distribution across all natural numbers below 10. Computing scores from distribution shape can thus also enable model ranking, similar to looking at distribution shape of model's correct response rates across variations.

\subsubsection{Overconfident tone and confabulations accompanying wrong answers}
\label{subsec:confabulation_overconfidence}

\textbf{Overconfident tone in wrong solutions.} Observing strong failures, we were curious to see how models explain their generated solutions. For the THINKING prompt type, where prompt contains request to double check the solution, we encounter examples where models spontaneously provide assessment of the solution quality and their confidence into the solution. Remarkably,  we see that in many cases of the observed responses with wrong reasoning and wrong final answers, the models claim high quality for their provided solution and are also strongly confident that the provided wrong solution is correct. For instance, Claude 3 Opus uses expressions like ``logic holds up; double-checking the solution; no mistakes in the reasoning; solution is correct.", and Command R+ reports ``This conclusion is straightforward and clear" for the wrong answers they provide.  

We further use variations of the prompt types to make the models generate estimates of the solution quality and their confidence on it, like the SCIENTIST prompt or the CONFIDENCE prompt (see Suppl. Sec. \ref{appendix:prompt types}). With those customized prompt types, we again observe strong overconfidence in the solution quality across the tested models. For the SCIENTIST prompt type, we see for instance Llama 2 70B using persuasive expressions like ``carefully analyzing; use logical reasoning; provide a precise and accurate solution; conclusion might seem counterintuitive at first, but it's actually correct" to back up its wrong solutions. For the CONFIDENCE prompt type, we see for instance for the wrong responses given by Command R+ accompanying statements like ``The solution is clear and unambiguous, and I am highly confident that it is correct."; ``I am confident in this answer, as it logically follows from the provided information.". See Suppl. Sec. \ref{appendix:overconfidence} for full examples.



\textbf{Confabulations to back up wrong solutions.} We observe that many models that show reasoning breakdown and produce wrong answers generate at the same time persuasive explanations that contain reasoning-like or otherwise plausible sounding statements to back up the often non-sensical solutions they deliver. We call here such phenomena \textit{confabulations}. Such confabulations may contain for instance calculations or logic-like statements that make no sense. Confabulations can also refer to reasoning about social norms or structures. For instance, in Command R+ we observe many confabulations that use concepts of gender identity such as non-binary gender or concepts related to inclusion or to cultural context dependent family identification as additional backup for the provided wrong reasoning and incorrect answers. Another type of confabulation that we observe is complete refusal to answer due to invented ethical concerns about the nature of the posed AIW problem, such as violation of privacy or lack of inclusion (for instance in CodeLLama-70B-instruct), or by expressing incorrect concerns about supposedly ill-posed problem formulation. See Suppl. Sec. \ref{appendix:confabulation} for more details.

\subsubsection{Further relevant observations.}
\label{subsec:further_relevant_observations}


We outline here further experiments that provide hints to the nature of the observed failures and incapabilities. 1. \textit{Inability to revise wrong solutions.} We experiment with customized prompts to enable multi-turn self-verification and also conduct multi-turn interactions with selected models to encourage those to revise their wrong solutions. In the majority of those attempts, while eagerly agreeing to revise the solutions and check those for possible mistakes, models show failure to properly detect mistakes and to revise wrong solutions (Suppl. Sec. \ref{appendix:revision_inability}). 2. \textit{Reformulation of AIW as relational SQL database problem.} We make use of relational logic underlying the AIW problem structure and prompt models to reformulate AIW into a correct relational SQL database format, using a customized SQL-FORM prompt type. We observe that smaller scale models, and also some larger scale ones, consistently fail to generate correct a relational SQL form. Some models are on the contrary able to do so frequently, e.g., Mistral/Mixtral (Suppl. Sec. \ref{appendix:reformulation_sql}) 3. \textit{Parameterized version AIW-Param}. We use a parameterized AIW formulation containing variables ($N$,$M$; or $X$,$Y$) for brothers and sisters, and we use variables instead of instantiating natural numbers into the template. We again observe a failure of the majority of the models to cope with this generic form, being unable to produce a correct generic solution, e.g. $C=M+1$ or $C=Y+1$ (Suppl. Sec. \ref{appendix:aiw_param_problem}).  4. \textit{Base model performance}. We inspect the performance of base model selection that claim strong function, e.g. Llama 2, Llama 3, Mistral/Mixtral by using AIW-base prompt types compatible with base model function. We observe the same behavior as with instruct models, measuring strong collapse on AIW problem (Suppl. Sec. \ref{appendix:base_models}. 5. \textit{In-context-learning (ICL)} As a shortcut solution in form of $M+1$ exists for AIW problem, we do not expect from ICL to install robust generalization when confronted with solved examples of AIW problem instances. We execute the experiment to check whether showing problem instances with natural numbers would enable solving of a more general AIW-Param problem. We observe failure of models to do so (Suppl. Sec. \ref{appendix:icl}) 

\subsection{AIW problems as measurement tool}


\subsubsection{Debunking strong function claims}
\label{subsec:debunking_strong_claims}

\begin{figure}[t!]
    \centering
    \includegraphics[width=1.0\textwidth]{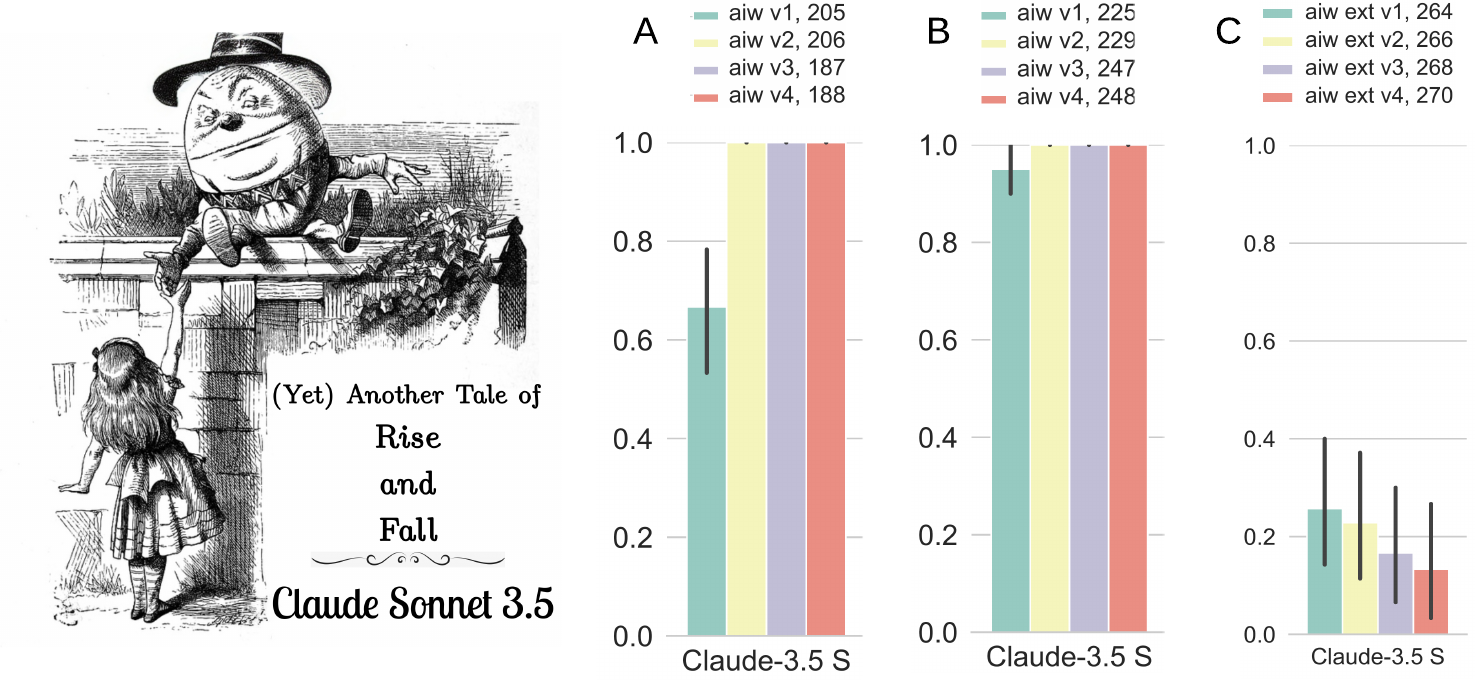}
    \caption{A Tale of Rise and Fall of Claude 3.5 Sonnet. While correct response rates go up close to 1 on \textbf{(A)} AIW original and also \textbf{(B)} AIW Original Bob version, strong breakdown of corrrect response rates is observed on AIW extension \textbf{(C)} (AIW Ext), accompanied with fluctuations across variations 1-4. Strongly elevated correct respones rates on AIW original might hint on exposure of Claude 3.5 Sonnet to AIW problem data for tuning. Collapse on AIW Ext, which has same problem structure as AIW original, shows though again clearly lack of robustness and hints on same basic reasoning deficits as suspected for other tested models.}
    \label{fig:aiw_orig_ext_claude35}
\end{figure}

\textbf{A Tale of Rise and Fall, 1: Claude 3.5 Sonnet}. After publication of our first manuscript version presenting AIW original problem (May 2024), Claude 3.5 Sonnet was released (June 2024), followed by wide acclaim as a strong frontiers LLM, as also confirmed by standardized benchmarks. We tested the model on AIW original problem variations 1-4. As the first among all tested models (at the time point, o1 model family was not existing yet), we observed for Claude 3.5 Sonnet correct response rates being 1 on all but AIW variation 1 (Fig. \ref{fig:aiw_orig_ext_claude35} \textbf{A}). This behavior is also unchanged when using Bob AIW Variation, replacing Alice with Bob and adapting formulation correspondigly to keep problem's original structure, Fig. \ref{fig:aiw_orig_ext_claude35} \textbf{B}. Claude 3.5 Sonnet appeared thus from that evidence to have solved AIW problem. We were cautious making conclusions from AIW original formulation only, as those problems were already publicly available and it could not be excluded that data was used for training of models appearing thereafter. We thus used AIW extended (AIW Ext, Sec. \ref{subsec:aiw_versions}) as it was not appearing in the initial version of our study. Like other AIW versions, AIW Ext has similar problem structure and difficulty as AIW original, AIW Ext variations 1-4 being also aligned with AIW original variations 1-4 wrt. correct final answers (Sec. \ref{subsec:aiw_versions} Suppl. Tab. \ref{tab:aiw_ext_prompts} ). We tested Claude 3.5 Sonnet auf AIW Ext variations 1-4 in same way we tested models on AIW variations 1-4. Claude 3.5 Sonnet revealed then strikingly different behavior on AIW Ext, Fig. \ref{fig:aiw_orig_ext_claude35} \textbf{C} - it shows strong breakdown to much lower correct response rates and again fluctuations across variations 1-4, despite AIW Ext problem structure being highly similar to the AIW original. This is in line to Claude 3.5 Sonnet behavior on other AIW versions, eg AIW+ (Fig. \ref{fig:aiw_plus_all}) and AIW Colleague Circles (Fig. \ref{fig:aiw_cc_average_fluctuations}), where it also exhibits lower correct response rates and strong fluctuations across variations.

\begin{figure}[t!]
    \centering
    \includegraphics[width=1.0\textwidth]{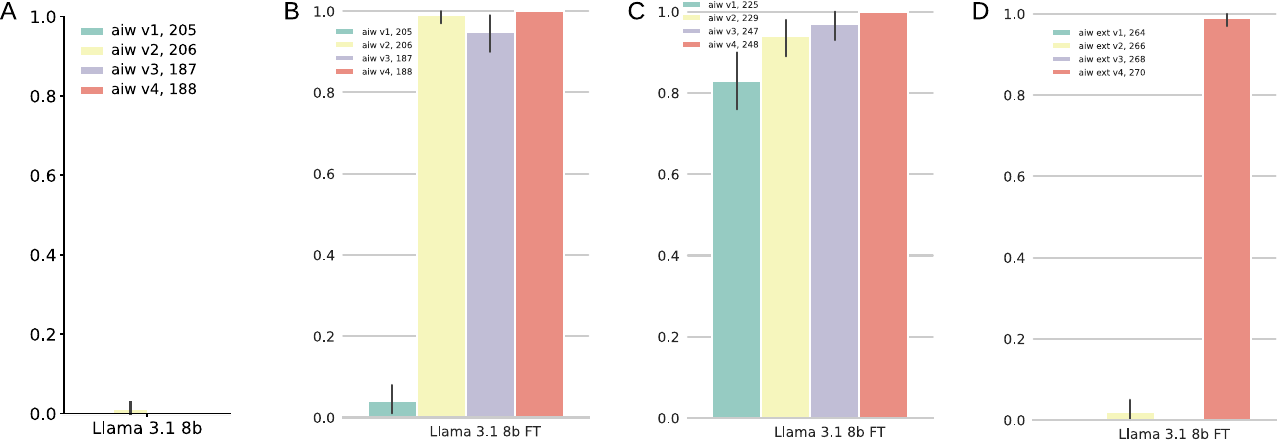}
    \caption{A possible explanation for Claude 3.5 Sonnet Tale of Rise and Fall, using Llama 3.1 8B, fine-tuned on generated samples containing AIW original variations 1-4, mixed with Alpaca instruction data. Original Llama 3.1 8B has correct response rates on AIW original close to 0 \textbf{(A)}. Following fine-tuning, we observe same pattern as in case of Claude 3.5 Sonnet in Fig. \ref{fig:aiw_orig_ext_claude35}. While correct response rates shoot up close to 1 on AIW original \textbf{(B)} and also on AIW original Bob version (replacing Alice with Bob) \textbf{(C)} for most variations, strong breakdown of correct response rates is observed on AIW Ext \textbf{(D)}, with strong fluctuations appearing. Strongly elevated correct responses rates on AIW original are caused by the exposure of Llama 3.1 8B to AIW problem data during instruction tuning, while original Llama 3.1 8B has correct response rates o or close to 0 across variations on AIW original \textbf{(A)}. Collapse on AIW Ext \textbf{(D)}, which has same problem structure as AIW original, shows though clearly still existing lack of robustness, revealing that generalization was not repaired by the fine-tuning on problem specific data.}
    \label{fig:aiw_orig_ext_llama8B_FT}
\end{figure}

After initial observation of high performance on AIW original, testing Claude 3.5 Sonnet on other AIW versions reveals same picture as observed for other tested models - strong fluctuations across AIW variations, testifying lack of model robustness to problem structure preserving variations. Claude 3.5 Sonnet behavior is strikingly different from other stronger performers on AIW original, but has same character on AIW Ext and other versions. This might be a hint that data similar to AIW original problem formulation was used for tuning Claude 3.5 Sonnet, which might have fixed its performance for that particular problem version, but did not endow it with robust generalisation, which still remains broken, as evident from breakdown on other AIW versions. We provide supporting evidence that such phenomena might indeed happen by our preliminary experiments, where we take open weights model Llama 3.1 8B and fine-tune it on the data containing various problem instances generated from AIW original problem template. As evident in Fig. \ref{fig:aiw_orig_ext_llama8B_FT} , similar picture emerges - we observe strong performance improvement of the fine-tuned Llama 3.1 8B on AIW original. It scores close to 1 on most of AIW variations, while still poorly performing on AIW Ext. Models like Claude 3.5 Sonnet report high scores on standardized benchmarks, however strong function claims cannot be derived from those, as we see from strong collapse and performance fluctuations on variations of rather simple problems as AIW Ext.

\textbf{A Tale of Rise and Fall, 2: NuminaMath-7B \& claim of olympiad level problem solving.} Another example of debunking overblown claims is a case of NuminaMath-7B that was ranked 1st at the AIMO competition in July 2024, solving 29/50 private set problems of olympiad math level. The claim was widely put forward that the model is capable of solving high school olympiad math problems. AIW has arguably average elementary school level and does not require any advanced math knowledge. We tested NuminaMath-7B on AIW and observed a strong collapse of this model on AIW problem, with correct response rates close to 0 across AIW variations 1-4 (Fig. \ref{}). Using AIW Light control problems, we can also see that NuminaMath-7B can handle all the low level operations (elementary arithmetic, attribute binding, etc) and knowledge required to deal with family structure, ruling out that those are the issues. Using the AIW problem setting, we thus can contradict the strong claim of being capable to deal with olympiad level high school math

\begin{figure}[thb!]
    \centering
    \includegraphics[width=1.0\textwidth]{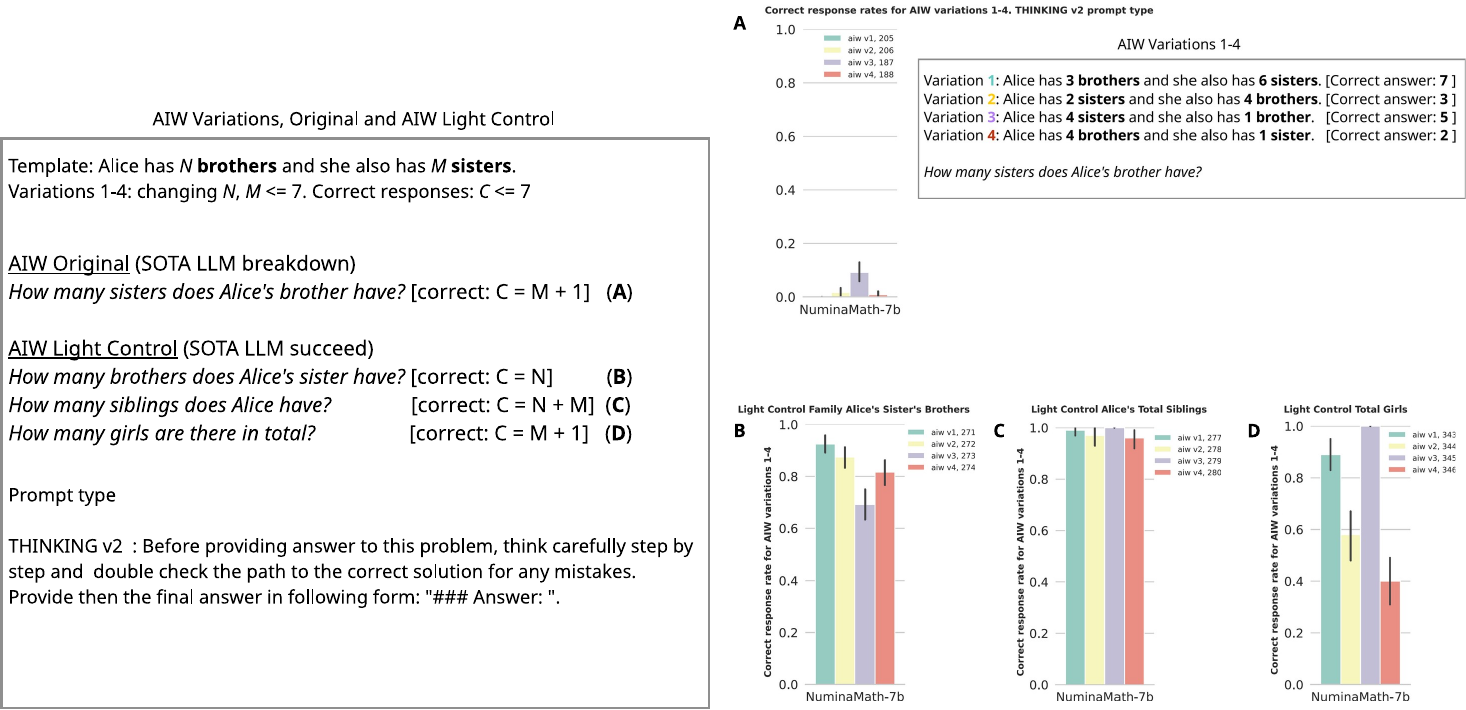}
    \caption{Testing NuminaMath-7B, which claimed olympiad high school level math problem solving via 1st rank in AIMO competition, by using AIW original and AIW Light control problems. (\textbf{A}) Very low correct response rates across AIW problem variations 1-4 (THINKING v2 prompt type). NuminaMath-7B suffers strong collapse on simple AIW problem that has average elementary school level. This reveals clear deficits in generalization and even basic reasoning, refuting the claim of strong function on special domain of math problems. For each AIW variation, 100 trials were executed to estimate correct response rate and its variance. AIW Light experiments test various operations and knowledge required for solving AIW (\textbf{B}) Asking for Alice's sister's brothers number (requires understanding entity "Alice's sister", binding female attribute to Alice and realizing Alice and her sisters share same brothers) (\textbf{C}) Asking for Alice's siblings number (requires understanding entity "siblings", accessing numbers of Alice's brothers and sisters, executing addition operation) (\textbf{D}) Asking for total girls number (requires binding female attribute to Alice via pronoun "she" and to her sisters, selecting and executing the correct arithmetic sum operation to count all the obtained girls). Across all AIW Light control problems, NuminaMath-7B obtains correct response rates much higher than for AIW original, some being close to 1. This proves that handling language, basic family structure, parsing numbers, and handling elementary arithmetics like counting are all intact and not the cause for failures in AIW (\textbf{A}). In (\textbf{D}), strong fluctuations despite only differences being instantiated numbers across variations of the same simple problem hint again on severe generalization deficits. The reason for the collapse on the AIW original problem is thus failure in inferring the problem structure, pointing to generalization and reasoning deficits, which is in contrast to claims made for NuminaMath-7B as a strong high school olympiad level math problem solver, based on AIMO competition benchmark, which did not reveal such flaws.}
    \label{fig:numina-7B-aiw_original_light_control_expl}
\end{figure}


\textbf{A Tale of Rise and Fall, 3: o1-mini \& claim of matching larger scale with smaller ones.} o1-mini was announced recently together with o1 and o1-preview as a smaller scale member of the new class of reasoning models. o1-mini was reported to obtain very strong scores on standardized benchmarks. Based on this, claims of strong function were put forward, specifically in comparison to the larger scale o1-preview. In original openAI announcement (Sep. 2024), o1-mini was reported to outperform larger-scale o1-preview on high school AIME math competition and coding tasks. This led to speculations that o1-mini can match or even outperform o1-preview as robust problem solver, a claim that is often put forward for models on smaller scales trained with substantial compute or obtained via distillation or other compression techniques from their larger scale counterparts.


\begin{figure}[thb!]
    \centering
    \includegraphics[width=0.7\textwidth]{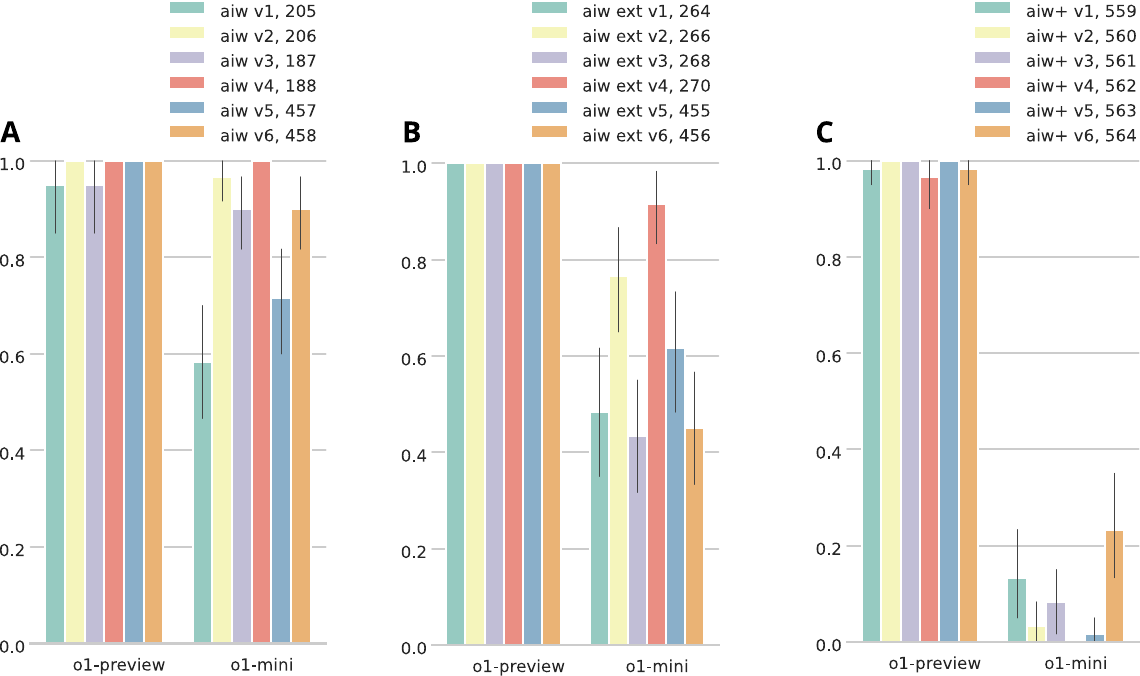}
    \caption{Testing o1-mini on the subject of the claims to match or outperform its larger scale counterpart, o1-preview. Comparing o1-preview and o1-mini using variations of AIW original (\textbf{A}), AIW Ext (\textbf{B}) and the harder AIW+. o1-mini shows progressive signs of breakdown, with fluctuations already visible in (\textbf{A}), becoming more apparent in (\textbf{B}) and strong collapse in (\textbf{C}), with correct response rates going down. o1-preview shows in contrast robust problem solving without fluctuations, keeping high correct response rates close to 1 across variations. This clear demonstrates o1-preview superiority contrary to claims of o1-mini strong function that were relying on various standardized benchmarks. See also Fig. \ref{fig:aiw_cc_average_fluctuations} for comparison on AIW Colleague Circles problem.}
    \label{fig:o1-mini_comparison}
\end{figure}

We tested o1-mini and o1-preview using various AIW problem templates introduced here and the problem structure and difficulty preserving variations. As evident from conducted experiments, o1-mini shows signs of breakdown starting with AIW original, that become more severe on AIW Ext, where fluctuations become stronger, culminating with collapse on AIW+, where correct performance rate on most variations stays well below $p=0.2$. o1-preview on contrary demonstrates robust problem solving with high correct response rates close to 1 without performance fluctuations across variations on all problem versions (see Fig. \ref{fig:aiw_cc_average_fluctuations} for AIW Colleague Circles problem where o1-preview also exhibits breakdown pattern - while still clearly outperforming o1-mini) This is in line with our observations reported here wrt. to smaller scale models and also specifically for o1-mini in comparison to other models (Fig. \ref{fig:aiw_plus_all}, \ref{fig:aiw_cc_average_fluctuations}) , where o1-mini stays far beyond o1-preview, and importantly, showing much higher sensitivity to problem structure preserving variations, pointing to severe generalization deficits, in contrast to o1-preview. 


Thus, AIW problem templates and the problem structure preserving variations offer a measurement technique that can reveal lack of robustness and model weaknesses in generalization and problem solving that remain undiscovered by standardized benchmarks. We think that conducted experiments can also serve as a vivid warning that many of the claims put forward for strong core functions of SOTA LLMs cannot be trusted. Such claims often rely on benchmarks that overlook clear function deficits, and simple AIW problem templates with their variations offers a tool for systematic, reproducible stress testing and debunking of such claims.





\subsubsection{Model comparison and ranking}
\label{subsec:model_comparison_ranking}
In previous experiments, we have shown that it is possible to use simple AIW problems as tool for falsification of strong function claims - robust generalization and strong problem solving - made for SOTA LLMs, revealing model breakdowns manifested in low correct response rates or strong fluctuations across problem structure preserving variations.

We also have seen that measuring overall correct response rates averaged across AIW variations (Fig. \ref{fig:main_correct_response_rate}, \ref{fig:mmlu_benchmark_vs_aiw}, \ref{fig:aiw_plus_all} (\textbf{A}), \ref{fig:aiw_cc_average_fluctuations} (\textbf{A})),  and looking at shape of correct response rate distribution across AIW variations (Fig. \ref{fig:aiw_original_fluctuations}, \ref{fig:aiw_problem_versions}, \ref{fig:aiw_plus_all} (\textbf{B}), \ref{fig:aiw_cc_average_fluctuations} (\textbf{B}), \ref{fig:o1-mini_comparison}) can provide ordering of models with regard to their capability to handle AIW problems. Here we demonstrate on selected scenarios that measuring capability to handle simple AIW problems \& their variations can provide model comparison and model ranking that might reflect true model generalization and problem solving capabilities better than standardized benchmarks.


\begin{figure}[tb!]
    \centering
    \includegraphics[width=1.0\textwidth]{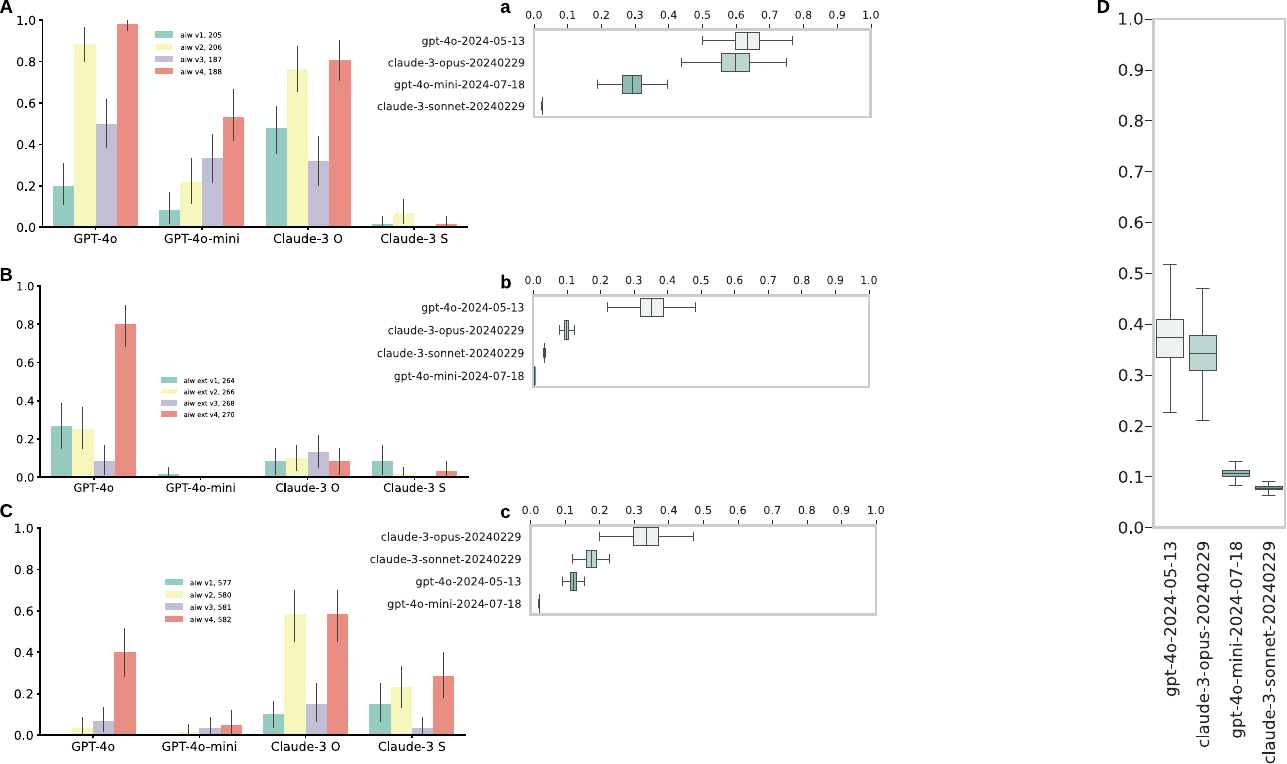}
    \caption{Model comparison and ranking for GPT-4o/GPT-4o-mini and Claude 3 Opus / Claude 3 Sonnet smaller-larger scale counterparts using performance measurement on AIW original (\textbf{A}), (\textbf{a}), AIW Ext (\textbf{B}), (\textbf{b}) and AIW Friends (\textbf{C}) (\textbf{c}) problems. Correct response rates distribution across variations 1-4 depicted in (\textbf{A}), (\textbf{B}) and (\textbf{C}) reveals strong performance fluctuations, across variations 1-4 in the problem template (a color per each variation) and across problem versions. While amount of fluctuations reveal various degree of model robustness, simple model comparison and ranking is possible via averaging correct response rates across variations as shown in (\textbf{a}), (\textbf{b}) and (\textbf{c}) per problem version, or by averaging correct response rates across problem versions as shown in (\textbf{D}). Similar to standardized benchmarks, AIW problems lead to ranking that puts larger scale GPT-4o/Claude 3 Opus into higher performance level than their smaller scales counterparts GPT-4o-mini/Claude 3 Sonnet. Contrary to standardized benchmarks, eg MMLU, GSM8k, HellaSwag, performance gap in model comparison larger/smaller scale is strongly pronounced, also placing smaller scale models to much lower performance level, indicating their strong breakdown. Considered AIW problems are simple, thus the observed breakdown of smaller scale models and strong gap to larger scale is not due to increasing problem difficulty compared to standardized benchmarks.}
    \label{fig:gpt-claude-mini_comparison}
\end{figure}

\textbf{Comparing models of various scales.} Similar to model comparison between o1-preview and o1-mini (Fig. \ref{fig:o1-mini_comparison}), where we were falsifying the hypothesis the smaller scale model being equal or even superior to its larger counterpart in problem solving, we continue with establishing model comparison using AIW problems.


As first example, we examine GPT-4o/GPT-4o-mini and Claude 3 Opus/Claude 3 Sonnet, which are also pairs of smaller and larger scale closed models counterparts. We take results from the experiments conducted with AIW original, AIW Ext and AIW Friends and plot both distributions of correct response rates for variations 1-4 of each corresponding AIW problem template (Fig. \ref{fig:gpt-claude-mini_comparison} (\textbf{A}), (\textbf{B}), (\textbf{C})), together with correct response rates averaged across variations for each problem (Fig. \ref{fig:gpt-claude-mini_comparison} (\textbf{a}), (\textbf{b}), (\textbf{c})) and overall correct response rates averaged across all problem versions (Fig. \ref{fig:gpt-claude-mini_comparison} (\textbf{D})). Similar to model comparison via standardized benchmarks, we observe larger scale GPT-4o/Claude 3 Opus ranking higher than their smaller scale counterparts GPT-4o-mini/Claude 3 Sonnet. In contrast to standardized benchmarks, there is clear breakdown of smaller scale GPT-4o-mini/Claude 3 Sonnet (below $p=0.1$) if measuring correct response rates averaged across all problems and variations (Fig. \ref{fig:gpt-claude-mini_comparison} (\textbf{D})), and gap to their larger scale counterparts GPT-4o/Claude 3 Opus (above $p=0.3$) is strongly pronounced. As a comparison to MMLU as standardized benchmark, GPT-4o/GPT-4o-mini with 0.887/0.82 and Claude 3 Opus/Claude 3 Sonnet with 0.882/0.815  exhibit much smaller gap and are all settled at higher performance levels. Importantly, as the considered AIW problems are simple, the strong breakdown of smaller scale models and their wide gap to larger scale cannot be explained by increasing problem difficulty compared to standardized benchmarks. As evident from fluctuations shown in Fig. \ref{fig:gpt-claude-mini_comparison}, models show various degree of robustness, both to variations of same problem template and to problem versions. Only if a model shows sufficient robustness to variations, it can exhibit higher overall correct response rates accumulated over variations and problem versions. Standardized benchmarks do not make use of such controlled problem variations, and are thus not sensitive to lack of model robustness, which is picked up by testing on AIW problems, despite their low difficulty level.

\begin{figure}[tb!]
    \centering
    \includegraphics[width=1.0\textwidth]{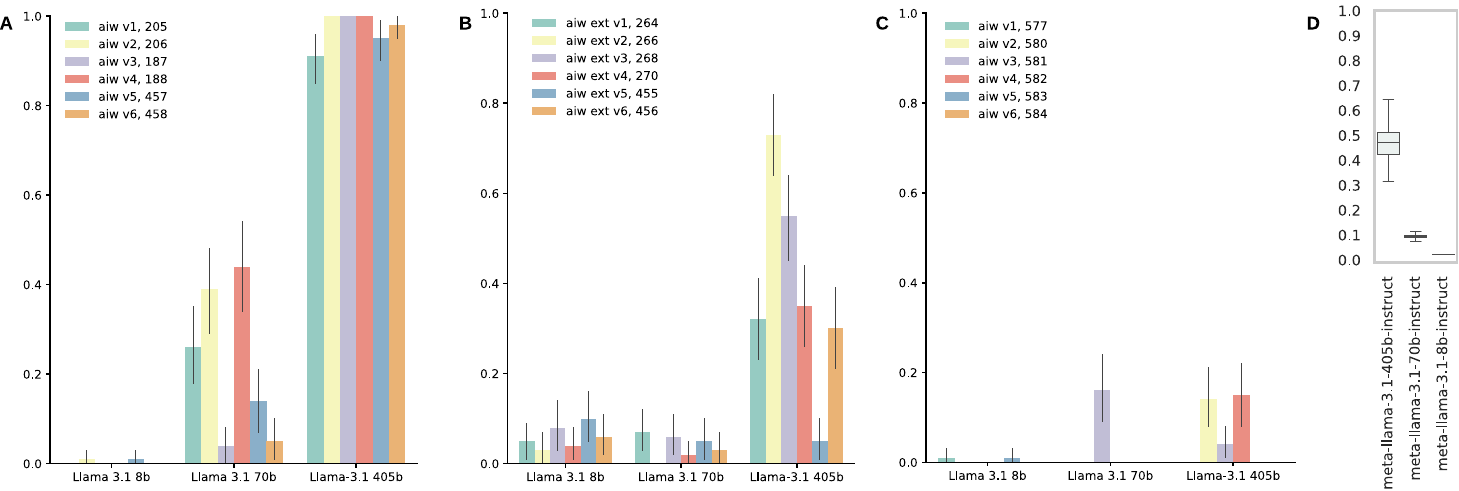}
    \caption{Model comparison and ranking, example of various scales of Llama 3.1 family. Models were tested on (\textbf{A}) AIW, (\textbf{B}) AIW Ext and (\textbf{C}) AIW Friends. Effect of scale is evident when comparing distribution of correct response rates across problem variations and correct response rate averaged across all problem versions in (\textbf{D}), hinting advantage of larger scale pre-training. Lack of model robustness is evident from strong collapse models suffer on AIW Friends, which is structurally similar to AIW and AIW ext.}
    \label{fig:llama_scaling}
\end{figure}

\begin{figure}[tb!]
    \centering
    \includegraphics[width=1.0\textwidth]{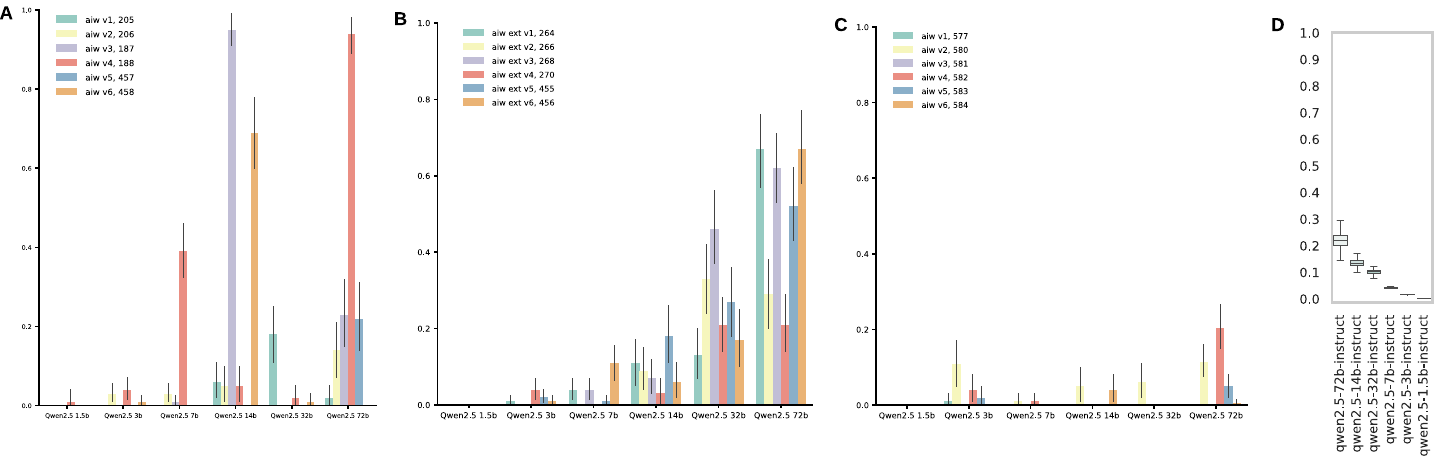}
    \caption{Model comparison and ranking, example of various scales of Qwen 2.5 model family. Models were tested on (\textbf{A}) AIW, (\textbf{B}) AIW Ext and (\textbf{C}) AIW Friends. Effect of scale is evident when comparing distribution of correct response rates across problem variations and correct response rate averaged across all problem versions in (\textbf{D}), hinting advantage of larger scale pre-training. Lack of model robustness is evident from strong collapse models suffer on AIW Friends, which is structurally similar to AIW and AIW ext.}
    \label{fig:qwen_scaling}
\end{figure}

As another example. we compare open weights model families Qwen 2.5 (Fig. \ref{fig:qwen_scaling} and Llama 3.1 (Fig. \ref{fig:llama_scaling}), which offer a broad range of pre-training scales (Qwen 2.5 1.5B-72B and Llama 3.1 8B-405B model scales). From results obtained from testing on AIW original and AIW Ext, we observe expected dependence of performance on pre-training scale. In line with previous observations, small scale models show strong breakdown with correct response rates close to 0 across variations. With larger scale, models exhibit higher correct response rates, while showing strong performance fluctuations across problem structure \& difficulty preserving variations. We see again strong performance fluctuation also across problem versions, for instance for Llama 3.1 405B having much higher correct response rate for AIW original compared to AIW Ext (Fig. \ref{fig:llama_scaling}), although both problems have highly similar structure. Similar to our observations on Claude 3.5 Sonnet (Fig. \ref{fig:aiw_orig_ext_claude35}), this might be again the case of training data containing AIW original instances, with strong diminished performance on AIW Ext pointing to generalization deficits. 

\begin{figure}[tb!]
    \centering
    \includegraphics[width=1.0\textwidth]{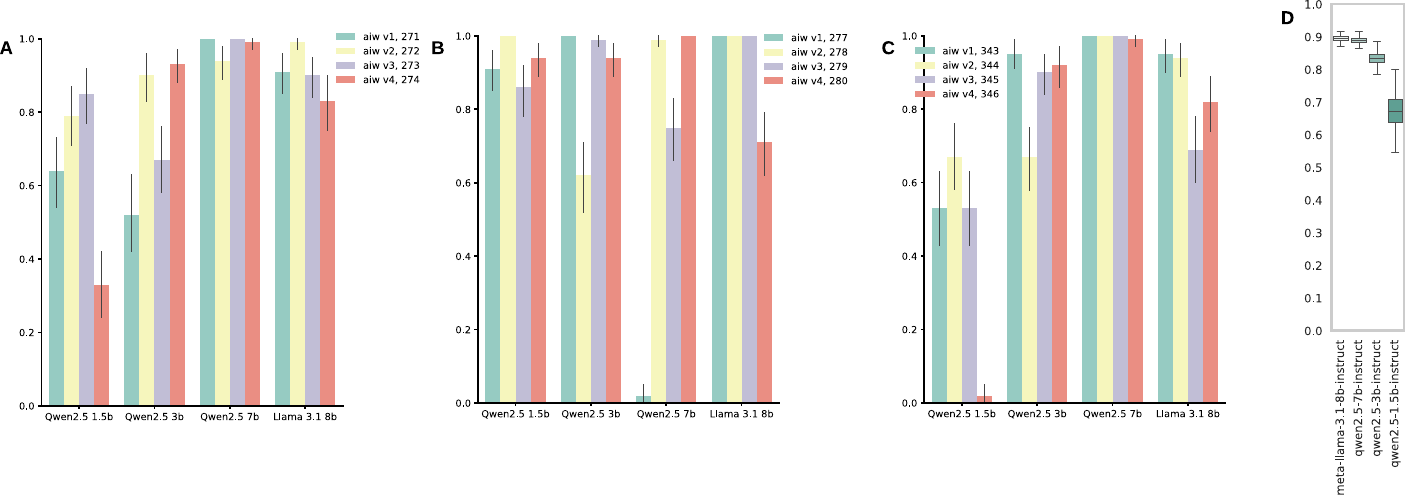}
    \caption{Model comparison and ranking on smaller scales using AIW Light problems, on example of smaller scale Qwen 2.5 and Llama 3.1 models. Models were tested on (\textbf{A}) AIW Light Family, (\textbf{B}) AIW Light Arithmetic Siblings and (\textbf{C}) AIW Arithmetic Total Girls. While smaller scale models lose signal on AIW, AIW Light makes it possible to see effect of scale also at smaller scales. Comparing distribution of correct response rates across problem variations and correct response rate averaged across all problem versions in (\textbf{D}), advantage of larger scale pre-training is visible - higher average correct response rates and less pronounced fluctuations.}
    \label{fig:aiw_light_llama_qwen_smaller_scales}
\end{figure}

For smaller scale models below 8B, the breakdown on AIW original and AIW Ext is so strong that correct response rates do no reflect differences in scale, most models at that small scales being close to 0. We can still measure differences on smaller scales by using AIW Light control problems, that are even simpler than already simple AIW original and AIW Ext. Using variations of problem templates of various difficulty level with upper level being still simple, we can thus compare models both at smaller scales and larger scales, obtaining model ranking without losing signal and differentiation at poorer or higher capability levels (Fig. \ref{fig:aiw_light_llama_qwen_smaller_scales}). 







\textbf{An unified score for measuring  model robustness.} Correct response rates averaged across variations of an AIW problem or averaged across various AIW problems with all their variations alone do not reveal the degree of present fluctuations across variations, which is important signature of model robustness or lack thereof, reflecting its generalization capability. Using average correct response rate alone can also hide the existing fluctuations. For instance, a model having very high correct response rate on only few problem variations, while others being close to 0, showing strong fluctuation, and a model having lower, but uniformly distributed correct response rates, showing no fluctuations, can show same overall correct response rate if averaged across variations. 

While it is possible to report eg. variance together with average correct response rate and avoid such ambiguity, we introduce here an unified score $\mathcal{R}$ that reflects both fluctuations degree and average correct response rate in one single number. To calculate $\mathcal{R}$ for any set of $K$ variations of one given AIW problem or a set of AIW problems (with total number of $K$ variations), we normalize correct response rates $p_i, i=1,\dots,K$ to obtain a probability distribution $x_i = \frac{p_i}{\sum_j^K p_j + \epsilon}$ (with $\epsilon=10^{-4}$ a small constant to avoid division by zero), computing then standard entropy re-scaled by log of number of variations, $H(X) = - \frac{1}{log K} \sum_i^K x_i \log x_i$. We also compute the average of correct response rates $\hat{p} = \frac{1}{K} \sum_i^K p_i$ and obtain $\mathcal{R} = H(X) \cdot \hat{p}$. $\mathcal{R}$ is close to 0 for models that either have overall low correct response rates or high peaks at few variations (model that have generalization deficits), while it is close to 1 for models that have uniformly high correct response rates across all variations (models with high robustness and strong generalization).

\begin{figure}[tb!]
    \centering
    \includegraphics[width=1.0\textwidth]{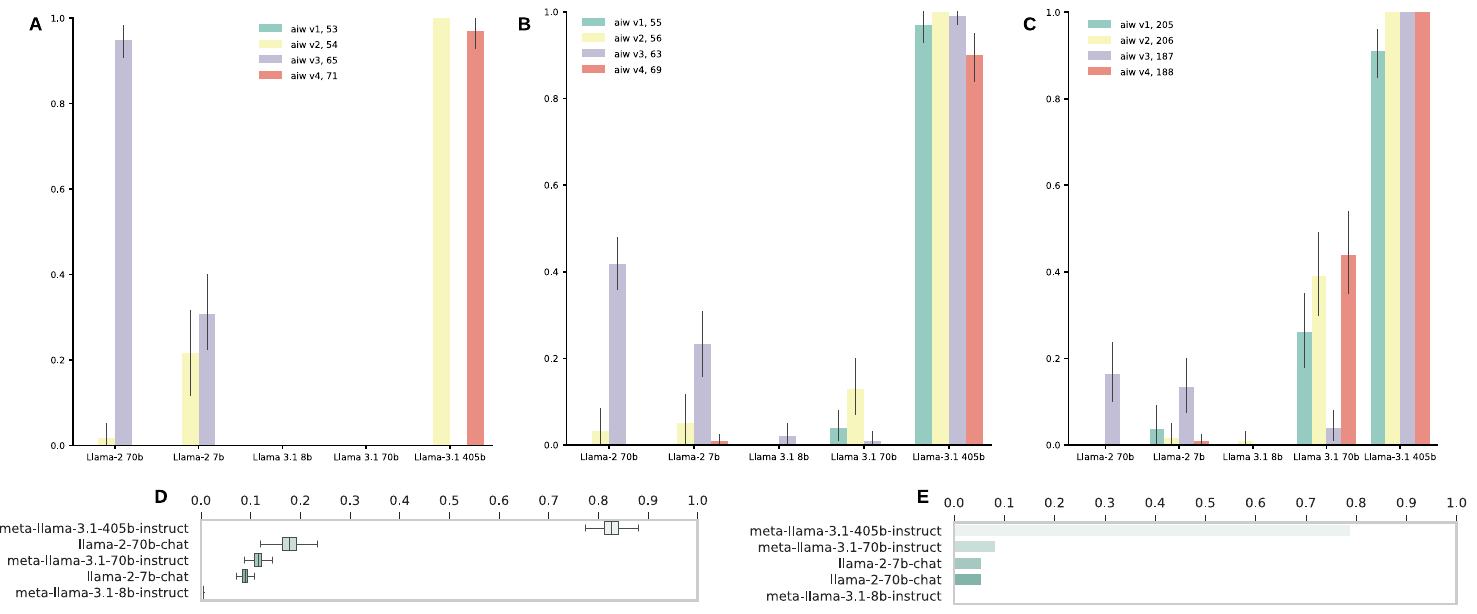}
    \caption{Model comparison via unified robustness score $\mathcal{R}$ on example of various Llama models. Correct response rates fluctuations across variations (a color per each variation) measured for tested models on AIW original problem variations 1-4 using (\textbf{A}) RESTRICTED, (\textbf{B}) STANDARD and (\textbf{C}) THINKING v2 prompt type. Regard strong outlier - variation 3 - for Llama 2 70B, which is the only of 4 variations showing significant non-zero correct response rate, others being 0 or close to 0. This outlier is the only contributor to the overall average correct response rate of Llama 2 70B as shown in (\textbf{D}). While average correct response rate ranks Llama 2 70B higher than Llama 3.1 70B, this ranking does not account for severe lack of robustness. Unified robustness score $\mathcal{R}$ punishes Llama 2 70B as evident in (\textbf{E}), putting it on same level as Llama 2 7B that despite having lower correct response rates, does not exhibit such extreme fluctuations. Llama 3.1 70B is ranked higher, reflecting its more consistent behavior - either low correct rates across all variations in (\textbf{A}) and (\textbf{B}), or consistently increased correct response rates across many variations in (\textbf{C}), indicating better robustness.}
    \label{fig:llama_comparison_robustness}
\end{figure}

We show how unified score $\mathcal{R}$ can be helpful in providing a more accurate view on model robustness and model ranking than one based average correct response rate and its variance alone in Fig. \ref{fig:llama_comparison_robustness}. As evident from Fig. \ref{fig:llama_comparison_robustness} (\textbf{D}), average correct response rate ranks Llama 2 70B higher than Llama 3.1 70B. This score is however driven by  only one outlier variation, where the model manages to show high performance, while other variations have either 0 or close to 0 correct response rate, as visible in full correct response rate distribution across variations in Fig. \ref{fig:llama_comparison_robustness} (\textbf{A}), (\textbf{B}), (\textbf{C}). The ranking via correct response rate averaged across variations thus does not reflect this severe lack of model robustness. Ranking via unified robustness score $\mathcal{R}$ on the other hand drops Llama 2 70B on the same level as Llama 2 7B (Fig. \ref{fig:llama_comparison_robustness} (\textbf{E})), punishing its strong fluctuations. Llama 3.1 70B is ranked higher, reflecting its more consistent behavior - either low correct rates across all variations, or increased correct response rates across consistently many variations, which indicates better robustness than Llama 2 70B. This advantage is not visible in Fig. \ref{fig:llama_comparison_robustness} (\textbf{D}) - when measuring average correct response rates only, Llama 3.1 70B falls behind less robust Llama 2 70B. Combining both rankings allows thus to avoid preferring less robust models over the more robust ones only due to extreme high performance outliers exhibited only on few variations out of many.

\begin{figure}[tb!]
    \centering
    \includegraphics[width=1.0\textwidth]{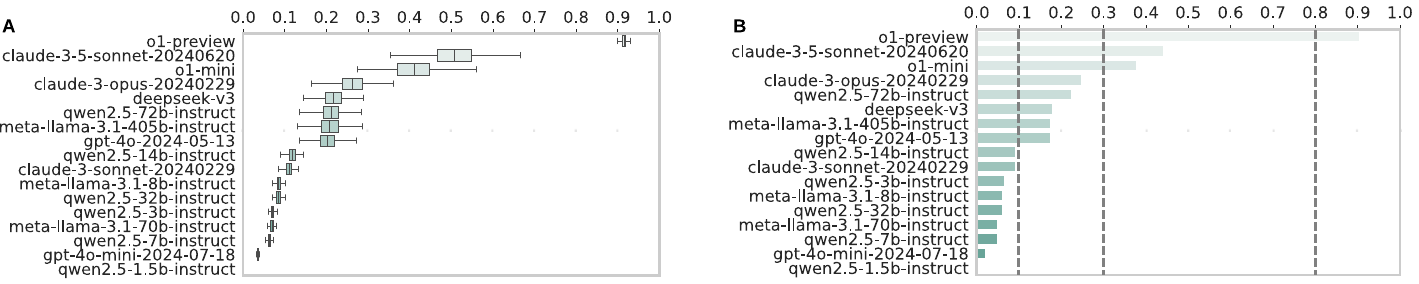}
    \caption{Model comparison and ranking using average correct response rates (\textbf{A}) and unified robustness score $\mathcal{R}$ (\textbf{B}). We use test set containing AIW Ext, AIW Friends, AIW Plus and AIW Circles Colleagues with their corresponding variations 1-6. Both rankings identify same model groups. o1-preview is the clearly strongest model, outperforming by far all other models. o1-mini and Claude 3.5 Sonnet settle in the mid performance region. The lower performance region still showing rudimentary robustness is occupied by larger scale SOTA LLMs - Claude 3 Opus, Qwen 2.5 72B, DeepSeek v3, GPT-4o and Llama 3.1 405B. In the lowest region are mostly smaller scale LLMs but also some larger ones like Qwen 2.5 32B and Llama 3.1 70B.}
    \label{fig:overview_ranking_comparison_robustness}
\end{figure}

In Fig. \ref{fig:overview_ranking_comparison_robustness} we show the overview for model comparison and ranking using both average correct response rates and unified robustness scores. We use set of all AIW problems presented here so far (AIW Ext, AIW Friends, AIW Plus, AIW Circles Colleagues) with exception of AIW original, as we see evidence of data leakage for more recent models like Claude 3.5 Sonnet (Sec. \ref{subsec:debunking_strong_claims}, Fig. \ref{fig:aiw_orig_ext_claude35}) or Llama 3.1 405B (Sec. \ref{subsec:model_comparison_ranking}, Fig. \ref{fig:llama_scaling}), and would like to avoid giving those models unfair advantage. Both measures deliver same picture for model ranking. We see that with exception of o1-preview ($\mathcal{R}=0.9$), all models exhibit strong lack of robustness, staying well below $\mathcal{R} = 0.5$. We indicate in Fig. \ref{fig:overview_ranking_comparison_robustness} (\textbf{B}) a possible separation into performance classes corresponding to high (o1-preview), mid (Claude 3.5 Sonnet, o1-mini - which settle in lower range of mid region), low (Claude 3 Opus, DeepSeek v3, Llama 405B, Qwen 2.5 72B, GPT-4o) and very low (Llama 3.1 70B, Qwen 2.5 32B, GPT-4o-mini and other smaller scale models) levels, which hold both with respect of robustness and average correct response rates. 

\subsubsection{Model comparison: advanced reasoning models}
\label{subsec:reasoning_model_comparison}
Recent works have demonstrated how open-weights reasoning models can be created by using open-weights LLMs as a base. Models like DeepSeek R1~\cite{guo2025deepseek} were created from base LLM DeepSeek v3, using simple reinforcement learning based losses (Group Relative Policy Optimization, GRPO) and conventional supervised fine-tuning (SFT), employing as training data real and synthetically generated reasoning traces in multiple training stages. Further models were distilled using reasoning traces, either synthetically generated from DeepSeek R1 (S1.1 32B~\cite{muennighoff2025s1}, OpenThinker 32B~\cite{openthinker32B,openthoughts}, based on Qwen 2.5 32B), DeepSeek R1 Zero (R1-Distilled-Qwen-32b and R1-Distilled-Llama-70b~\cite{guo2025deepseek}) or curated from mix of real and synthetic reasoning data (LIMO-32B~\cite{ye2025limo}). These reasoning models show strong increase compared to conventional LLMs in scores on standardized benchmarks related to problem solving at olympiad or graduate difficulty levels (AIME24~\cite{AIME}, MATH500~\cite{hendrycks2021math}, GPQA Diamond~\cite{rein2024gpqa}).

We measure the robustness and generalization performance of these models using AIW problems, to test the strong claims behind the high scores achieved on reasoning benchmarks (see also Sec. \ref{subsec:debunking_strong_claims}) and check how these models compare to conventional LLMs. To reduce probability of test data leakage (see also Sec. \ref{subsec:debunking_strong_claims}, Fig. \ref{fig:aiw_orig_ext_claude35} and Sec. \ref{subsec:model_comparison_ranking}, Fig. \ref{fig:llama_scaling}), we exclude AIW original and AIW ext from AIW test set for the experiments, as reasoning models released very recently might have been exposed to data from those or very similar problems due to their public availability. We take thus AIW Friends, AIW Plus and AIW Circles Colleagues as problem test set.

\begin{figure}[tb!]
    \centering
    \includegraphics[width=1.0\textwidth]{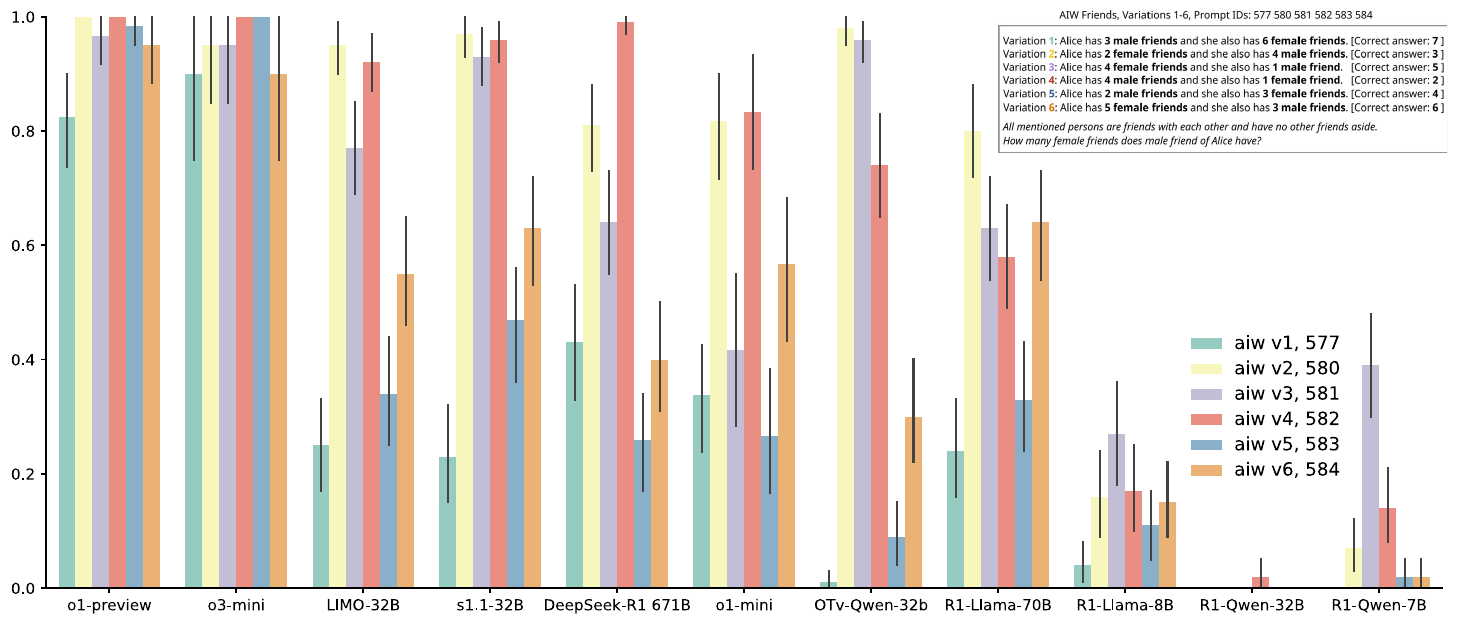}
    \caption{Strong fluctuations of correct response rates on variations of AIW Friends problem (a color per each variation 1-6) exhibited by recent reasoning models. With exception of o1-preview and o3-mini, most reasoning models that show strong performance on standardized reasoning benchmarks reveal inability to cope with the problem robustly when facing slight variations in problem template. Eg, correct response rate drops for DeepSeek R1 671 from nearly 1.0 on variation 4 down to 0.2 on variation 1, despite differences between the two being just instantiated numbers. R1-Qwen-32B collapses entirely. Distilled reasoning models (S1.1 32B, LIMO 32B, OpenThinker-Qwen-32B) perform on par with DeepSeek-R1 671B and o1-mini, despite using for distillation SFT only. Distilled models at larger scales (32B, 70B) perform significantly better than smaller scale 7B/8B models (with exception of R1-Qwen-32B). Larger scale distilled models, DeepSeek-R1 and o1-mini show remarkable similarity in the distribution of correct response rates across problem variations.} 
    \label{fig:reasoning_AIW_Friends}
\end{figure}


\begin{figure}[tb!]
    \centering
    \includegraphics[width=1.0\textwidth]{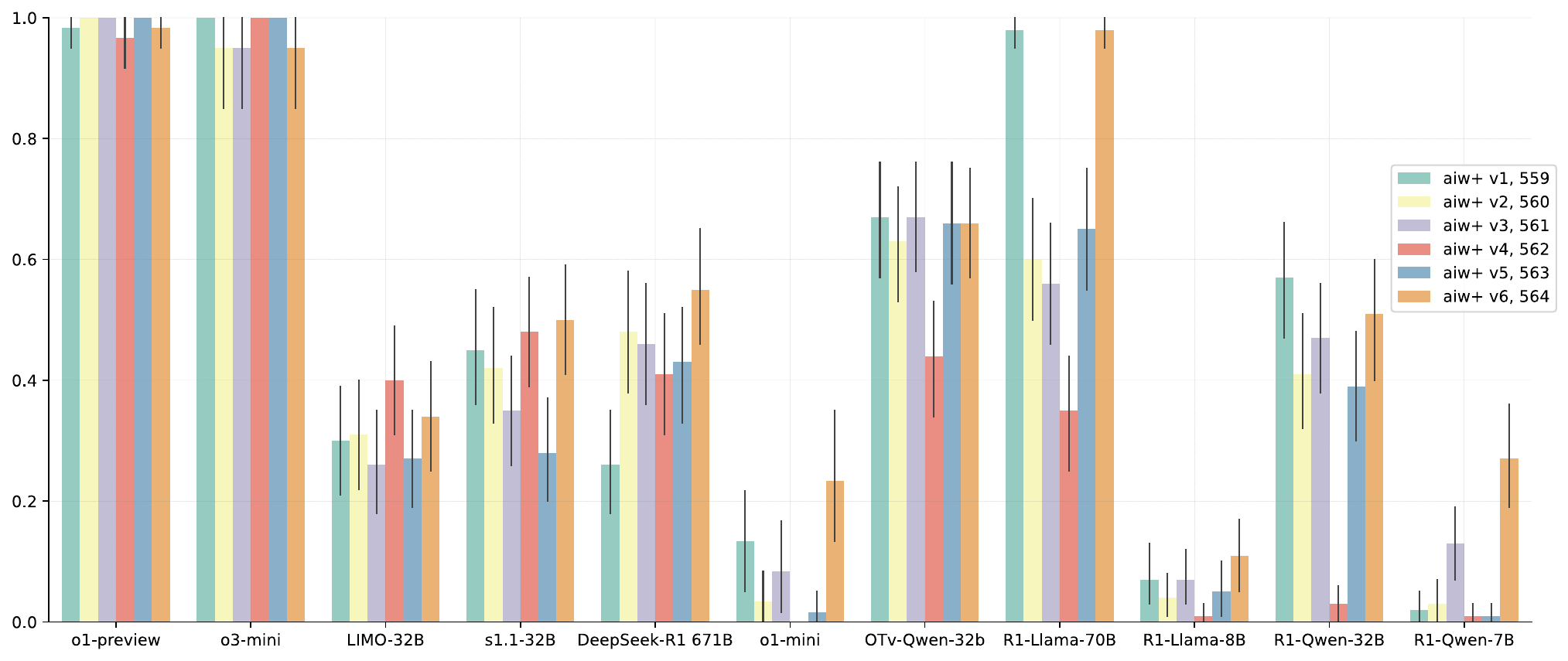}
    \caption{Strong fluctuations of correct response rates on variations of AIW Plus problem (a color per each variation 1-6) exhibited by recent reasoning models. With exception of o1-preview and o3-mini, most reasoning models that show strong performance on standardized reasoning benchmarks reveal inability to cope with the problem robustly when facing slight variations in problem template. Eg, correct response rate drops for R1-Llama 70B from close to 1.0 on variation 1 or 6 below 0.4 on variation 4, despite differences between the those being just instantiated numbers. o1-mini undergoes strong overall collapse. Distilled reasoning models (S1.1 32B, LIMO 32B, OpenThinker-Qwen-32B) perform on par with DeepSeek-R1 or outmatch it, despite using for distillation SFT only. Distilled models at larger scales (32B, 70B) perform significantly better than smaller scale 7B/8B models.}
    \label{fig:reasoning_AIW_Plus}
\end{figure}

\begin{figure}[tb!]
    \centering
    \includegraphics[width=1.0\textwidth]{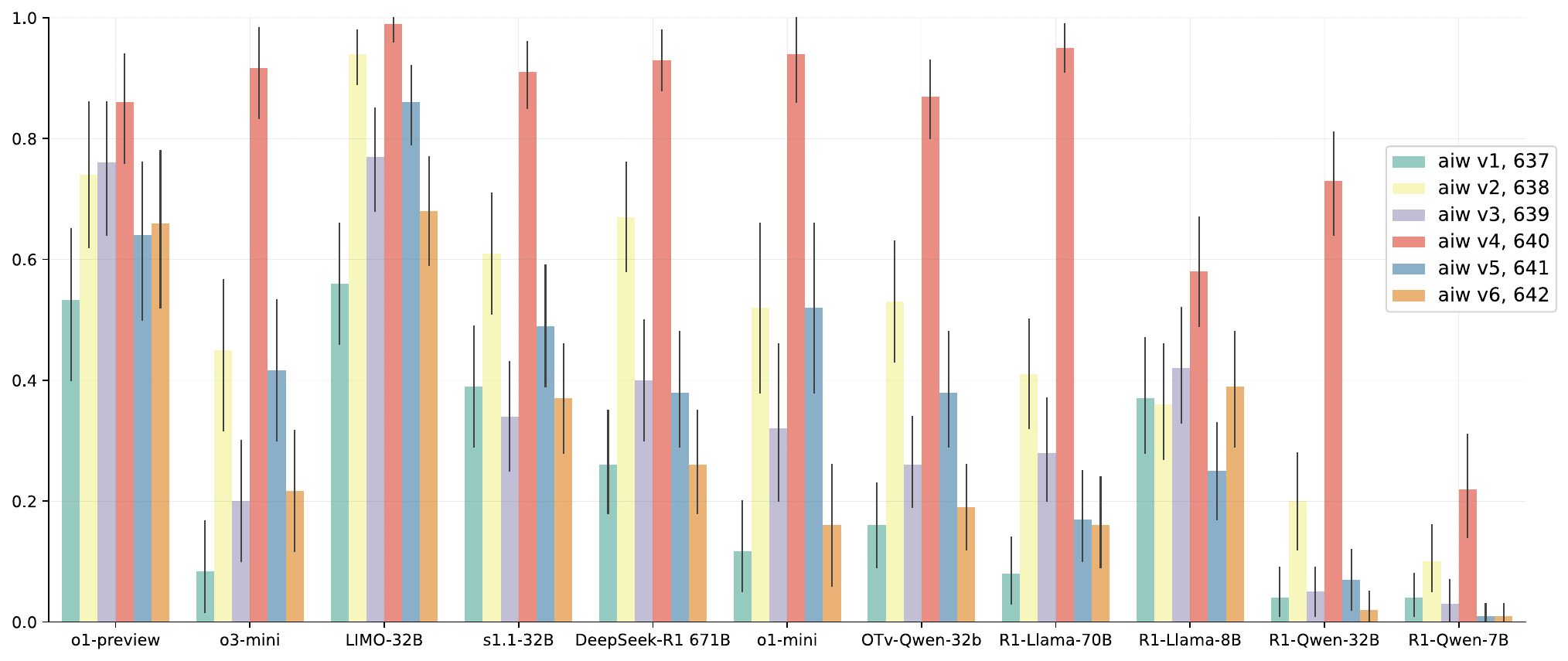}
    \caption{Strong fluctuations of correct response rates on variations of AIW Circles Colleagues problem (a color per each variation 1-6) exhibited by recent reasoning models, including top frontier models like o1-preview and o3-mini (medium). Fluctuations reveal that all reasoning models with strong performance on standardized reasoning benchmarks are unable to cope with the problem robustly, being sensitive to slight variations in problem template. Eg, correct response rate drops for o3-mini-medium from above 0.9 on variation 4 to below 0.1 on variation 1, despite differences between those being just instantiated numbers. Distilled reasoning models (S1.1 32B, LIMO 32B, OpenThinker-Qwen-32B) perform on par with DeepSeek-R1 or outmatch it, despite using for distillation SFT only. Distilled models at larger scales (32B, 70B) perform significantly better than smaller scale 7B/8B models (with exception of R1-Llama-8B vs R1-Llama-70B). Larger scale distilled models, DeepSeek-R1, o3-mini and o1-mini show remarkable similarity in the distribution of correct response rates across problem variations.}
    \label{fig:reasoning_AIW_Circles_Colleagues}
\end{figure}

\begin{figure}[tb!]
    \centering
    \includegraphics[width=1.0\textwidth]{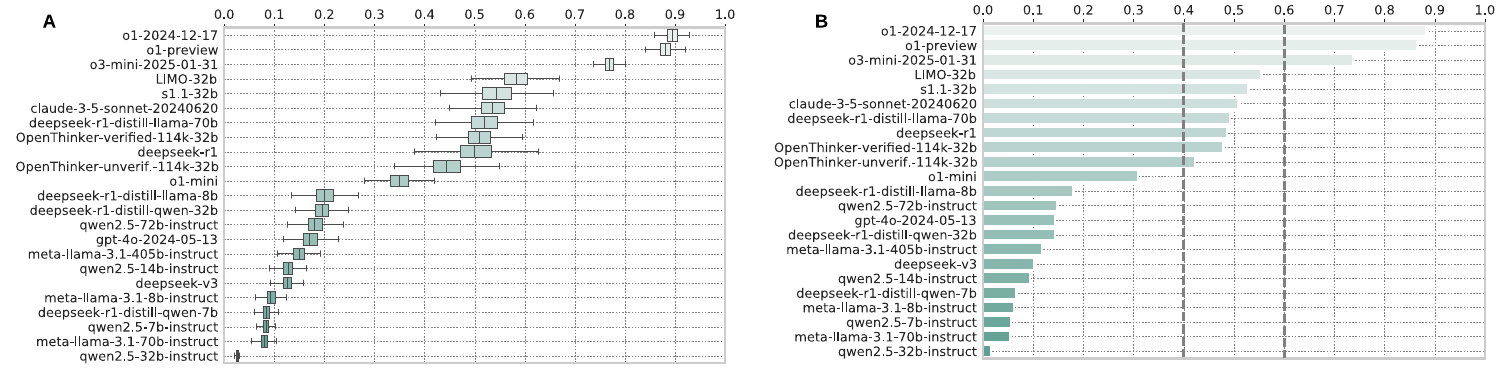}
    \caption{Comparing and rankining recent reasoning models on AIW test set (AIW Friends, AIW Plus, AIW Circles Colleagues) using average correct response rates (\textbf{A}) and unified robustness score $\mathcal{R}$ (\textbf{B}). Evident is a strong performance boost for distilled larger scale (32B, 70B) reasoning models (with exception of R1-Qwen-32B), settling in mid range $0.4 < p, \mathcal{R} < 0.6$ for average correct response rates $p$ and robustness $\mathcal{R}$, leaving standard SOTA LLMs far behind. Standard SOTA LLMs and smaller scale distilled reasoning models reside below $0.2$. Larger scale distilled reasoning models match or outperform DeepSeek-R1 and o1-mini. This provides evidence that using only SFT on reasoning traces for the fine-tuning can be sufficient to reach performance level comparable with methods that employ RL. Interestingly, OpenThinker-Unverified-32B, trained only on unverified traces, while having lower performance than OpenThinker-Verified-32B, also manages to gain substantial boost, hinting that also unverified reasoning traces might have strong value as reasoning boosting data.}
    \label{fig:reasoning_AIW_Model_Ranking_Comparison}
\end{figure}

As evident from Fig. \ref{fig:reasoning_AIW_Friends}, \ref{fig:reasoning_AIW_Plus}, and \ref{fig:reasoning_AIW_Circles_Colleagues}, despite their high scores on standardized reasoning benchmarks, most reasoning models still suffer from strong fluctuations across AIW problem variations. Notable exception is again o1-preview, although also this model exhibit significant fluctuations and decreased average correct response rates on AIW Circles Colleagues. This shows that claims of robust problem solving of olympiad or graduate level as signalled by strong performance on reasoning benchmarks are not sustainable, as AIW problems where models show lack of robustness are far below these levels. On the other hand, reasoning models indeed strongly improve in robustness and average correct response rates compared to tested conventional SOTA LLMs as evident from Fig \ref{fig:reasoning_AIW_Model_Ranking_Comparison}. Remarkable is that this improvement in robustness is also achieved by distilled reasoning models that use only SFT on reasoning traces in single stage post-training. Models like R1-Distilled-Llama-70B~\cite{guo2025deepseek} (distilled on 800k closed data), OpenThinker-32B~\cite{openthinker32B,openthoughts} (distilled on OpenThoughts-114k, 3 epochs), S1.1-32B~\cite{muennighoff2025s1} (distilled on 1k DeepSeek-R1 data), LIMO-32B~\cite{ye2025limo} (distilled on 0.8k samples mix from real and DeepSeek-R1 generated data), having either Llama 3.3 or Qwen 2.5 as a base, are increasing their performance strongly compared to their base instruct models, reaching up to levels comparable with DeepSeek R1 671B, which uses both SFT and RL during multi stage training, also outperforming closed reasoning models like o1-mini. We observe remarkable similarity in the distribution shape of correct response rates between larger scale SFT only distilled models and DeepSeek-R1 or o1-mini and o3-mini that use RL. Further, as seen on Fig \ref{fig:reasoning_AIW_Model_Ranking_Comparison},  OpenThinker-Unverified-32B manages also to gain substantial boost despite using unverified traces for distillation (in contrast to OpenThinker-Verified-32B and other distilled reasoning models). This provides a hint that also unverified reasoning traces might be already sufficient to induce strong performance boost in problem solving compared to standard SOTA LLMs, while verified traces add further boost.



\section{Related work \& limitations}
\label{sec:related_limitations}

\textbf{Measuring LLMs capabilities.} Since the seminal breakthroughs in language modelling~\cite{devlin2018bert,raffel2020exploring, brown2020language}, measuring LLM capabilities became indispensable for evaluations and model comparison. To measure how well a language model performs on reasoning, there exists a plethora of different standardized reasoning benchmarks. These benchmarks can be roughly divided into categories by what exact reasoning capability we want to test such as ARC \cite{clark2018think}, PIQA \cite{bisk2020piqa}, GSM8K \cite{cobbe2021training}, HellaSwag \cite{zellers2019hellaswag}, MMLU \cite{hendrycks2020measuring} or WinoGrande \cite{DBLP:journals/corr/abs-1907-10641}. Multiple works aim on improving reasoning performance of LLMs as measured by those standardized benchmarks in various ways \cite{wei2022chain,yao2024tree,zhou2022least,wang2022self,pfau2024let}. Recently, increasingly challenging benchmarks have been developed, aiming at testing of olympiad and graduate level problem solving, for instance AIME24/AIME25~\cite{AIME}, MATH500~\cite{hendrycks2021math,lightman2023let}, or GPQA Diamond~\cite{rein2024gpqa}, which contain problems of very high difficulty and are also used to test recent reasoning models~\cite{guo2025deepseek, muennighoff2025s1,openthinker32B,ye2025limo}.





\textbf{Finding weak spots in LLMs capabilities.} Paralleling impressive progress shown by LLM research, cautious voices have been raising concern about discrepancy between claimed capabilities as measured by standardized benchmarks and true model reasoning skills by presenting carefully selected evidence for model failures \cite{mitchell2023we}. Number of works were investigating test data leakage into training set as possible cause of observed discrepancy between claimed and actual capabilities, pointing out high test dataset contamination due to large-scale pre-training on web-scale data \cite{golchin2023time,li2024task}.  In response, the research community has been undertaking attempts to create benchmarks like HELM~\cite{liang2023holistic}, BIG-bench~\cite{srivastava2023beyond} or LiveCodeBench~\cite{jain2025livecodebench} that are supposed to circumvent data leakage issues and properly test generalization capabilities beyond memorization.







\begin{figure*}[t!]
    \centering
    \includegraphics[width=1.0\textwidth]{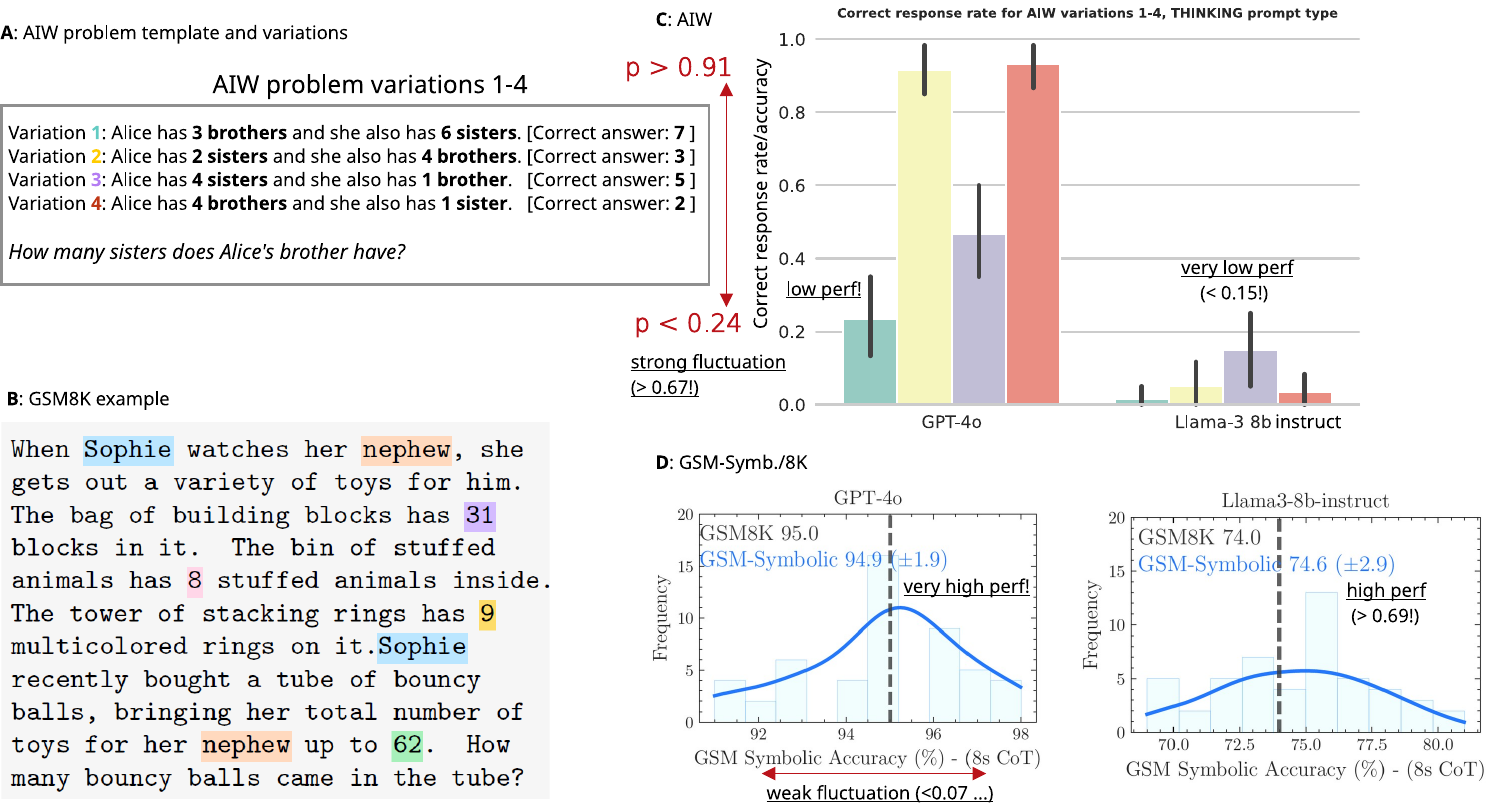}
    \caption{ Both AIW (\textbf{A}) and GSM-Symbolic (GSM-S) (\textbf{B}) use variations in problem templates to measure sensitivity of model performance to variations and draw conclusions about generalization and reasonining capabilities. AIW (\textbf{C}) provides strong evidence for generalization deficits by observing 1) strong fluctuations across variations of simple AIW problem (a color for each AIW variation 1-4) and 2) low average correct response rates for most models, eg. Llama 3 8B on the right. This provides convincing falsification of strong function hypothesis. In contrast, GSM-S (\textbf{D}), while using more sophisticated and content overloaded (eg see (\textbf{B})) problems, cannot offer such conclusive evidence. The observed fluctuations are weak (eg 0.07 vs 0.67 on AIW for GPT-4o), and average performance is high (eg > 0.69 vs < 0.15 on AIW for Llama 3 8B), while GSM8k falls well within the measured response distribution. While AIW variations thus reveal GPT-4o and Llama-3 8B generalization deficits, they stay hidden for GSM-S. We hypothesize that inability to observe strong fluctuations on GSM-S might be due to leakage of GSM8k type of problems into training data. Fig. (\textbf{B}) and (\textbf{D}) are adapted from \cite{mirzadeh2024gsm}}
    \label{fig:aiw_vs_GSMSymb_expl}
\end{figure*}

Multiple studies \cite{wu2023reasoning,dziri2024faith,lewis2024using,berglund2023reversal,moskvichev2023conceptarc,huang2023large} have shown breakdowns of language models reasoning capabilities in different scenarios and also lack of robustness to variation of problem formulation~\cite{zongfool2024,zheng2024large}. Other works were looking into particular reasoning failures like deficits in causality inference \cite{jin2023can,jin2023cladder}. These works operate often with formalized problems of high difficulty that does not have simple common sense character expressible in natural language. Similar in spirit to our work, simple math word problems were used in \cite{patel2021nlp} to show model breakdowns, but models on current frontiers that claim strong generalization and advanced capabilities were not tested. Recently, Mirzadeh et. al. (2024) made in their work~\cite{mirzadeh2024gsm} use of similar approach to create variations from templates of GSM8K problems. This work however does not provide for strong models any conclusive evidence of function breakdown on simple problems, in fact showing evidence that models like GPT-4o or Llama 3 8B can handle those well (Fig. \ref{fig:aiw_vs_GSMSymb_expl}). In contrast, we are able to measure lack of model robustness and strong fluctuations also for SOTA LLMs and advanced reasoning models pretrained at largest scales.  A key limitation of our current approach is the lack of sufficient diversity and number of AIW problem variations. This can be addressed in future work by systematic procedural instance generation for evaluation of model behavior on more diverse problem set with larger variation number.



\section{Discussion \& Conclusion}
\label{sec:discussion_conclusion}


In our work, using a simple AIW problem (Sec. \ref{sec:methods_experiment}) 
that requires only elementary set and arithmetic operations and can be easily solved by adults and arguably even children, we observe a striking breakdown of SOTA LLMs performance when confronted with the AIW problem and its variations (Sec. \ref{subsec:aiw_problem}, Suppl. Tab. \ref{tab:aiw_prompt_types}). The breakdown is manifested in \textbf{(i)} Low average correct response rates (Fig. \ref{fig:main_correct_response_rate}) and \textbf{(ii)} Strong performance fluctuation across structure and difficulty preserving natural variations of the same problem, which hints at fundamental issues with the generalization capability of the models (Fig. \ref{fig:aiw_original_fluctuations}). 

The observed breakdown is in strong contrast with claims about SOTA LLMs being robust solvers of higher difficulty problems at graduate or olympiad level. Any system claiming even basic robust reasoning should be able to obtain high correct response rates close to $100\%$ on problems as simple as AIW across all its variations. The obtained evidence also falsifies the claim of strong zero-shot generalization - strong performance fluctuations across variations that keep problem structure and difficulty unchanged reveal severe generalization deficits in all tested SOTA LLMs (Fig. \ref{fig:aiw_original_fluctuations}, \ref{fig:aiw_problem_versions}, \ref{fig:aiw_plus_all}, \ref{fig:aiw_cc_average_fluctuations}, \ref{fig:reasoning_AIW_Friends}, \ref{fig:reasoning_AIW_Plus}, \ref{fig:reasoning_AIW_Circles_Colleagues}). Given simplicity of the AIW problem, one can speak of complete reasoning breakdown in the most of the tested models, as observed strong fluctuations clearly prove the inability to infer robustly simple common problem structure behind all AIW variations. 

By executing experiments on control AIW Light problems, we provide evidence that the observed failures are not rooted in low-level issues like inability to parse crucial problem related information from natural language or execute elementary arithmetic operations required to solve it (Sec. \ref{subsec:aiw_light_problems_exp_results}). We also confirm similar breakdown patterns in further AIW problem versions like AIW Ext, AIW Friends, AIW+ and AIW Circles Colleagues (Fig. \ref{fig:aiw_problem_versions}, \ref{fig:aiw_plus_all}, \ref{fig:aiw_cc_average_fluctuations}, \ref{fig:reasoning_AIW_Friends}, \ref{fig:reasoning_AIW_Plus}, \ref{fig:reasoning_AIW_Circles_Colleagues}), corroborating hypothesis that observed deficits in generalization and basic reasoning are generic and not specific to the specific problem type. Our study also clearly points to failure of standardized benchmarks to properly measure core model functionality such as generalization or reasoning (Fig. \ref{fig:mmlu_benchmark_vs_aiw}, Suppl. Sec. \ref{appendix:benchs_failure}, Suppl. Tab. \ref{tab:bench_aiw}). Standardized benchmarks assigning high scores to SOTA LLMs fail to reveal severe model weaknesses made evident by breakdown on simple AIW problem. 

It has to be noted that despite observed breakdown with low average correct response rates and strong fluctuations, the reasoning is not entirely absent. Better performing larger scale models like GPT-4/4o,  Claude 3 Opus, Claude 3.5 Sonnet do show examples of fully correct reasoning (see Suppl Fig. \ref{fig:aiw_ex_gpt4_turbo_right}, \ref{fig:aiw_ex_claude_opus_right}). As our results show, this reasoning capability is however fragile and cannot be accessed robustly, even in such a simple scenario as posed by AIW problem variations. The same is valid for the very recent advanced reasoning models like open-weights DeepSeek R1 671B or closed o1-preview, o3-mini and o1-mini. While they do show clear improvement over standard SOTA LLMs, strong fluctuations on problem structure and difficulty preserving variations still appear (Fig. \ref{fig:o1-mini_comparison}, \ref{fig:reasoning_AIW_Friends}, \ref{fig:reasoning_AIW_Plus}, \ref{fig:reasoning_AIW_Circles_Colleagues}, again debunking claims of robust graduate or olympiad level problem solving put forward for these models based on standardized reasoning benchmarks (AIME24, MATH500, GPQA Diamond).

Our findings raise the question how it is possible that standardized benchmarks - that contain much harder problems than AIW - attest seemingly strong function capabilities to the tested models, while testing via simple AIW problems reveal such clear breakdown. In line with previous works~\cite{golchin2023time,li2024task,dominguez2024training}, we hypothesize that testing on standardized benchmarks suffers either from direct benchmark test data leakage or from training on test tasks that slip into training and post-training datasets of models created after benchmark appearance. We confirm with our experiments such scenario (Fig. \ref{fig:aiw_orig_ext_llama8B_FT}). Simple supervised fine-tuning (SFT) on a set of AIW problems is sufficient to strongly increase performance of Llama 3.1 8B from very poor to very high across variations of the same problem as used in fine-tuning. The fine-tuned models undergoes though severe collapse when going to closely related, but slightly different AIW problem versions. Similar phenomena is observed when testing Claude 3.5 Sonnet (Fig. \ref{fig:aiw_orig_ext_claude35}) or Llama 3.1 405B (Fig. \ref{fig:llama_scaling}) which appeared after public data on AIW problems was released - performing very strong on one set of AIW problems while collapsing on closely related versions. It becomes thus again clear that in order to properly measure generalization in models trained on large web-scale data, it is crucially to design benchmarks that are immune at least against direct test data leakage.  

The appraisal for the smaller scale models (e.g. Mistral-7B; LLama 2/3 7/8B; Qwen 2.5 1.5B, 3B, 7B) might be thus also based to large extent on illusion of strong function mediated by standardized benchmarks that suffer from test data leakage. We observe a severe breakdown for smaller scale models on AIW, with a clear large gap to better performing models that all reside at larger scales (Fig. \ref{fig:main_correct_response_rate}, \ref{fig:llama_scaling} ,  Fig. \ref{fig:qwen_scaling}, Fig. \ref{fig:llama_scaling}, Suppl. Fig. \ref{fig:aiw_all}, Suppl. Tab. \ref{tab:bench_aiw}). We hypothesize here that the claimed strong functions of smaller scale models might be a mere illusion corroborated by broken benchmarks that in their current state cannot offer a proper model comparison and thus also cannot be used as downstream tasks for measuring important scaling laws.

To perform basic research in direction of improving generalization and reasoning of the models, it is thus important that whole pipeline of model creation - dataset composition and dataset itself, source code for training, the trained model itself, the standardized benchmarking procedure and test data - is fully open and reproducible. Models that has open weights only do not allow for proper analysis whether their performance on test data is due to strong core function and generalization, or rather due to test data being in the training dataset. Closed models accessible via API only often do not even allow proper evaluation, as for instance default settings such as system prompt and inference hyperparameters may remain invisible to independent evaluation parties. We think therefore that proper progress in studying how to measure and install strong generalization in the future models necessarily requires full pipeline of model creation - especially the often neglected dataset composition - to be open-source (see works on FineWeb \cite{FineWeb2024_Blog} and DCLM-baselines \cite{li2025datacomp} for encouraging examples of such open dataset composition pipelines), as otherwise the claims about generalization and reasoning capabilities will stay unsubstantiated. We show here a way to debunk such claims in the closed models using AIW problems. This however still relies on creating a problem set that is not yet a part of the training data, which is hard to ensure if training dataset is unknown.

Relying on claims backed up by standardized benchmarks, it is commonly-held position to attribute to SOTA LLMs advanced functions like zero-shot reasoning~\cite{kojima2022large}, and in general to put high expectations of strong core functionality on released SOTA LLMs~\cite{ achiam2023gpt,touvron2023llama,touvron2023llama2,jiang2023mistral}. Such claims extend beyond basic research artifacts and become pervasive in applied industry, where SOTA LLMs are advertised as robust problem solvers for various real world settings, explicitly emphasizing their value as robust reasoners, coders and problem solvers, attesting "key business-critical capabilities" or suitability for "real-world enterprise use cases" (see announcements by Cohere on Command R-Plus~\cite{CommandR2024, Cohere2024A}, or by Mosaic on DBRX~\cite{Dbrx2024}, as only few selected representative examples out of many). Our study provides clear evidence for model breakdown on simple problems, and importantly, lack of model robustness in face of natural problem variations that do not change problem structure or its difficulty.

Observed breakdown of model robustness, coupled with public claims based on standardized benchmarks present an inherent safety problem. Models with insufficient generalization and lack of robustness in basic reasoning are inherently unsafe. As we observe sensitivity to problem variations in such a simple scenario, it is highly likely that similar lack of robustness will manifest on variations of real world problems to solve, resulting in unexpected performance changes. Those will be hard to detect for users if the problem solution is not trivial to check as it in case of simple AIW problems. 

It should be thus clear that SOTA LLMs cannot be trusted to deal robustly with real world tasks, or to robustly solve problems of graduate or olympiad level as claimed by high scores on benchmarks containing those. Performance on current standardized reasoning benchmarks creates illusion of robust reasoning capabilities, and making it even worse - when doing mistakes, models are overconfident, insisting on their wrong answers being correct, and producing highly persuasive and suggestive explanations for their wrong responses, which might further obscure errors for the end-users due to partly plausible sounding text (see Suppl. Sec. \ref{appendix:confabulation} for examples of such confabulations). To ensure safety, public claims about model function should be based only on those scientific evaluations that properly measure the models' capabilities to generalize and reason, while basic research has to be performed guided by such benchmarks to endow future models with sufficient generalization and robust basic reasoning skills. 

Our study should thus serve as a vivid warning that current SOTA LLMs are not yet capable of strong generalization and robust reasoning, and enabling those is still subject of basic research. We see in the recent reasoning models a promising direction to enable stronger model function than SOTA LLMs can offer, as our initial experiments with AIW problem testing show. To properly measure generalization and guide further progress, novel benchmarks are though required. AIW problem and its variations offer a starting point and a measurement technique that can reveal lack of robustness and model weaknesses in generalization and core functions that remain undiscovered by current benchmarks. Variations built into problem templates can serve as method for creating new benchmarks that are, in contrast to current common benchmarks, no longer static and can provide much more reliable measurement and testing of model generalization and reasoning, avoiding test set leakage issues by instantiating any amount of diverse problem instances through combinatorial power. New benchmarks should follow Karl Popper's principle of falsifiability \cite{popper1934logic}, attempting everything to break model's function, highlighting its deficits, and thus showing possible directions for model improvement, while also offering protection from overblown claims about model's capabilities.

\section*{Acknowledgments}

We would like to express gratitude to all the people who are working on making code, models and data publicly available, advancing community based research and making research more reproducible. Specifically, we would like to thank all the members of the LAION Discord server\footnote{\url{https://discord.gg/BZqhreFazY}} community and Open-$\Psi$ (Open-Sci) Collective\footnote{\url{https://discord.gg/GsKh4mBVcv}} for providing fruitful ground for scientific exchange and open-source development. 

MN, LCK, MC and JJ acknowledge funding by the Federal Ministry of Education and Research of Germany (BMBF) under grant no. 01IS22094B (WestAI - AI Service Center West), under grant no. 01IS24085C (OPENHAFM) and under grant 16HPC117K (MINERVA), as well as co-funding by EU from EuroHPC Joint Undertaking programm under grant no. 101182737 (MINERVA) and from Digital Europe Programme under grant no. 101195233 (openEuroLLM).

LCK acknowledges the Helmholtz Information \& Data Science Academy (HIDA) for providing financial support enabling a short-term research stay at Juelich Supercomputing Center (JSC), Research Center Juelich (FZJ) to conduct research on foundation models.

\bibliographystyle{unsrt}
\bibliography{references}

\clearpage
\begin{appendix}

\begin{center}
{\Large\bf Supplementary.}
\end{center}

\section{Additional details on performed experiments}
\label{appendix:exp_details}

Here we give further details on the procedures around the executed experiments. 

\subsection{Models selected for experiments}
\label{subsec:model_tested_selection}

To provide overview over origin of core tested models used for the AIW experiments, we list those in Suppl Tab. \ref{tab:aiw_models_overview}. All tested models use same default inference hyperparameters, $T = 0.1$, $\textit{top-p} = 1.0$ (we executed control experiments to check that various settings do not change the main pattern in the observed behavior). The output of standard LLMs was limited to $2048$ tokens, and as evident from Suppl Fig. \ref{fig:avg_len_of_responses_aiw}, most observed responses stayed well below this limit. When testing recent reasoning models, the limit was raised to 32k or 43k, depending on model type.

\begin{table*}[ht]
    \caption{Names, origin and versioning of core test models used in the experiments.}
    \label{tab:aiw_models_overview}
    \vskip 0.15in
    \begin{center}
        \begin{small}
            \begin{tabular}{p{4.5cm}p{1.5cm}p{2.5cm}p{1.5cm}p{3.9cm}}
                \toprule
                Name & Origin & Released & Open Weights & Sources \\
                \midrule
                GPT-4o-2024-05-13 & OpenAI & 13.05.2024 & No & \cite{achiam2023gpt, GPT4o, HelloGPT4o} \\
                GPT-4-turbo-2024-04-09 & OpenAI & 09.04.2024 & No & \cite{achiam2023gpt, GPT4} \\
                GPT-4-0125-preview & OpenAI & 25.01.2024 & No & \cite{achiam2023gpt, GPT4} \\
                GPT-4-0613 & OpenAI & 13.06.2023 & No & \cite{achiam2023gpt, GPT4} \\
                GPT-3.5-turbo-0125 & OpenAI & 24.01.2024 & No & \cite{OpenAI2022, OpenAIGPT35,OpenAIGPT35_Turbo_0125} \\
                Claude-3-5-sonnet-20240620 & Anthropic & 21.06.2024 & No & \cite{Anthropic2024B} \\
                Claude-3-opus-20240229 & Anthropic & 04.03.2024 & No & \cite{Anthropic2024A, anthropic2024claude} \\
                Claude-3-sonnet-20240229 & Anthropic & 04.03.2024 & No & \cite{Anthropic2024A, anthropic2024claude} \\
                Claude-3-haiku-20240307 & Anthropic & 04.03.2024 & No & \cite{Anthropic2024A, anthropic2024claude} \\
                Gemini 1.0 Pro & Google & 06.12.2023 & No & \cite{Pichai2023, team2023gemini} \\
                Gemini 1.5 Pro & Google & 16.02.2024 & No & \cite{Gemini1_5_announce, team2024gemini} \\
                gemma-7b-it & Google & 05.04.2024 (v1.1) & Yes & \cite{Gemma2024_announce, Gemma2024_modelcard} \\
                gemma-2b-it & Google & 05.04.2024 (v1.1) & Yes & \cite{Gemma2024_announce, Gemma2024_modelcard} \\
                Mistral-large-2402 & Mistral AI & 26.02.2024 & No & \cite{MistralAI2024_API, MistralAI2024_Large} \\
                Mistral-medium-2312 & Mistral AI & 23.12.2023 & No & \cite{MistralAI2024_API, MistralAI2024_Large} \\ 
                Mistral-small-2402 & Mistral AI & 26.02.2024 & No & \cite{MistralAI2024_API, MistralAI2024_Large} \\
                open-mixtral-8x22b-instruct-v0.1 & Mistral AI & 17.04.2024 & Yes & \cite{MistralAI2024_API, MistralAI2024_7x22B} \\
                open-mixtral-8x7b-instruct-v0.1 & Mistral AI & 11.12.2023 & Yes & \cite{MistralAI2024_API, MistralAI2024_7x8B} \\
                open-mistral-7b-instruct-v0.2 & Mistral AI & 11.12.2023 & Yes & \cite{jiang2023mistral, MistralAI2024_API, MistralAI2024_7B} \\
                Command R+ & Cohere & 04.04.2024 & Yes & \cite{Cohere2024A, Cohere2024B} \\
                Dbrx Instruct & Mosaic & 27.03.2024 & Yes & \cite{Dbrx2024} \\
                Llama 2 70B Chat & Meta & 18.07.2023 & Yes & \cite{Meta2023, touvron2023llama2} \\
                Llama 2 13B Chat & Meta & 18.07.2023 & Yes & \cite{Meta2023, touvron2023llama2} \\    
                Llama 2 7B Chat & Meta & 18.07.2023 & Yes & \cite{Meta2023, touvron2023llama2} \\
                Llama 3 70B Chat & Meta & 18.04.2024 & Yes & \cite{Meta2024_Llama3, Meta2024_Llama3_modelcard} \\    
                Llama 3 8B Chat & Meta & 18.04.2024 & Yes & \cite{Meta2024_Llama3, Meta2024_Llama3_modelcard} \\
                Qwen 1.5 1.8B - 72B Chat & Alibaba & 04.02.2024 & Yes & \cite{bai2023qwen, Qwen2024_announce} \\
                Qwen 2 72B Instruct & Alibaba & 07.06.2024 & Yes & \cite{Qwen2_announce} \\
                \bottomrule
            \end{tabular}
        \end{small}
    \end{center}
    \vskip -0.1in
\end{table*}

\section{Problem versions, prompt types and variations}
\label{appendix:prompt types}

Here we provide overview of all AIW problem versions as introduced in Sec. \ref{subsec:aiw_problem} with their corresponding variations and prompt types used. 

For the full inputs provided to the tested models during experiments, see Suppl. Tab. \ref{tab:aiw_prompt_types} for AIW problem, Suppl. Tab. \ref{tab:aiw_light_arithmetic_siblings} (AIW Light Arithmetic Siblings), \ref{tab:aiw_light_family} (AIW Light Family), \ref{tab:aiw_light_arithmetic_total_girls} (AIW Light Arithmetic Total Girls) for AIW Light control problems, Suppl. Tab. \ref{tab:aiw_female_boost_variation_types} for AIW Alice Female Boost, Suppl. Tab. \ref{tab:aiw_exaggerated_num_prompts} for AIW Exaggerated Numbers,  Suppl. Tab. \ref{tab:aiw_ext_prompts} for AIW Extended (AIW Ext), Suppl. Tab.  \ref{tab:aiw_friends_prompts} for AIW Friends, Suppl. Tab.  \ref{tab:aiw_plus_prompt_types} for AIW Plus and  Suppl. Tab.  \ref{tab:aiw_circles_prompt_types} for AIW Circles Colleagues.




\begin{table*}
\centering
\caption{AIW main variations 1-4, prompt types and correct answers overview.}
\label{tab:aiw_prompt_types}
\begin{tabular}{p{0.5cm}p{11cm}p{2cm}p{1cm}}
\toprule
Var. & Prompt & Type/Answer & ID \\
\midrule
1 & Alice has 3 brothers and she also has 6 sisters.
 How many sisters does Alice's brother have? Solve
 this problem and provide the final answer in
 following form: "\verb|#|\verb|#|\verb|#| Answer:
 ". & STANDARD / 7 & 55 \\
1 & Alice has 3 brothers and she also has 6 sisters.
 How many sisters does Alice's brother have? Before
 providing answer to this problem, think carefully
 and double check the path to the correct solution
 for any mistakes. Provide then the final answer in
 following form: "\verb|#|\verb|#|\verb|#| Answer:
 ". & THINKING / 7 & 57 \\
1 & Alice has 3 brothers and she also has 6 sisters.
 How many sisters does Alice's brother have? To
 answer the question, DO NOT OUTPUT ANY TEXT EXCEPT
 following format that contains final answer:
 "\verb|#|\verb|#|\verb|#| Answer: ". & RESTRICTED / 7 & 53 \\
2 & Alice has 2 sisters and she also has 4 brothers.
 How many sisters does Alice's brother have? Solve
 this problem and provide the final answer in
 following form: "\verb|#|\verb|#|\verb|#| Answer:
 ". & STANDARD / 3 & 56 \\
2 & Alice has 2 sisters and she also has 4 brothers.
 How many sisters does Alice's brother have? Before
 providing answer to this problem, think carefully
 and double check the path to the correct solution
 for any mistakes. Provide then the final answer in
 following form: "\verb|#|\verb|#|\verb|#| Answer:
 ". & THINKING / 3 & 58 \\
2 & Alice has 2 sisters and she also has 4 brothers.
 How many sisters does Alice's brother have? To
 answer the question, DO NOT OUTPUT ANY TEXT EXCEPT
 following format that contains final answer:
 "\verb|#|\verb|#|\verb|#| Answer: ". & RESTRICTED / 3 & 54 \\
3 & Alice has 4 sisters and she also has 1 brother.
 How many sisters does Alice's brother have? Solve
 this problem and provide the final answer in
 following form: "\verb|#|\verb|#|\verb|#| Answer:
 ". & STANDARD / 5 & 63 \\
3 & Alice has 4 sisters and she also has 1 brother.
 How many sisters does Alice's brother have? Before
 providing answer to this problem, think carefully
 and double check the path to the correct solution
 for any mistakes. Provide then the final answer in
 following form: "\verb|#|\verb|#|\verb|#| Answer:
 ". & THINKING / 5 & 64 \\
3 & Alice has 4 sisters and she also has 1 brother.
 How many sisters does Alice's brother have? To
 answer the question, DO NOT OUTPUT ANY TEXT EXCEPT
 following format that contains final answer:
 "\verb|#|\verb|#|\verb|#| Answer: ". & RESTRICTED / 5 & 65 \\
4 & Alice has 4 brothers and she also has 1 sister.
 How many sisters does Alice's brother have? Solve
 this problem and provide the final answer in
 following form: "\verb|#|\verb|#|\verb|#| Answer:
 ". & STANDARD / 2 & 69 \\
4 & Alice has 4 brothers and she also has 1 sister.
 How many sisters does Alice's brother have? Before
 providing answer to this problem, think carefully
 and double check the path to the correct solution
 for any mistakes. Provide then the final answer in
 following form: "\verb|#|\verb|#|\verb|#| Answer:
 ". & THINKING / 2 & 70 \\
4 & Alice has 4 brothers and she also has 1 sister.
 How many sisters does Alice's brother have? To
 answer the question, DO NOT OUTPUT ANY TEXT EXCEPT
 following format that contains final answer:
 "\verb|#|\verb|#|\verb|#| Answer: ". & RESTRICTED / 2 & 71 \\
4 & Alice has 4 brothers and she also has 1 sister.
 How many sisters does Alice's brother have? Solve
 the problem by taking care not to make any
 mistakes. Express your level of confidence in the
 provided solution as precisely as possible. & CONFIDENCE / 2 & 11 \\
3 & Alice has 4 sisters and she also has 1 brother.
 How many sisters does Alice's brother have? To
 solve the problem, approach it as a very
 intelligent, accurate and precise scientist
 capable of strong and sound reasoning. Provide the
 solution to the problem by thinking step by step,
 double checking your reasoning for any mistakes,
 and based on gathered evidence, provide the final
 answer to the problem in following form:
 "\verb|#|\verb|#|\verb|#| Answer: ". & SCIENTIST / 5 & 40 \\
\bottomrule
\end{tabular}
\end{table*}

\begin{table*}
\centering
\caption{AIW Light Arithmetic Siblings variations  1-4}
\label{tab:aiw_light_arithmetic_siblings}
\begin{tabular}{p{0.5cm}p{11cm}p{2cm}p{1cm}}
\toprule
Var. & Prompt & Type/Answer & ID \\
\midrule
1 & Alice has 3 brothers and she also has 4 sisters. How many siblings does Alice have? Before providing answer to this problem, think carefully step by step and double check the path to the correct solution for any mistakes. Provide then the final answer in
 following form: "\verb|#|\verb|#|\verb|#| Answer:". & THINKING v2 / 7 & 277 \\
2 & Alice has 2 sisters and she also has 1 brother. How many siblings does Alice have? Before providing answer to this problem, think carefully step by step and double check the path to the correct solution for any mistakes. Provide then the final answer in
 following form: "\verb|#|\verb|#|\verb|#| Answer:". & THINKING v2 / 3 & 278 \\
3 & Alice has 4 sisters and she also has 1 brother. How many siblings does Alice have? Before providing answer to this problem, think carefully step by step and double check the path to the correct solution for any mistakes. Provide then the final answer in
 following form: "\verb|#|\verb|#|\verb|#| Answer: ". & THINKING v2 / 5 & 279 \\
4 & Alice has 1 brother and she also has 1 sister. How many siblings does Alice have? Before providing answer to this problem, think carefully step by step and double check the path to the correct solution for any mistakes. Provide then the final answer in following form: "\verb|#|\verb|#|\verb|#| Answer:". & THINKING v2 / 2 & 280 \\
\bottomrule
\end{tabular}
\end{table*}

\begin{table*}
\centering
\caption{AIW Light Family variations  1-4}
\label{tab:aiw_light_family}
\begin{tabular}{p{0.5cm}p{11cm}p{2cm}p{1cm}}
\toprule
Var. & Prompt & Type/Answer & ID \\
\midrule
1 & Alice has 7 brothers and she also has 3 sisters. How many brothers does Alice's sister have? Before providing answer to this problem, think carefully step by step and double check the path to the correct solution for any mistakes. Provide then the final answer in
 following form: "\verb|#|\verb|#|\verb|#| Answer:". & THINKING v2 / 7 & 271 \\
2 & Alice has 4 sisters and she also has 3 brothers. How many brothers does Alice's sister have? Before providing answer to this problem, think carefully step by step and double check the path to the correct solution for any mistakes. Provide then the final answer in
 following form: "\verb|#|\verb|#|\verb|#| Answer:". & THINKING v2 / 3 & 272 \\
3 & Alice has 2 sisters and she also has 5 brothers. How many brothers does Alice's sister have? Before providing answer to this problem, think carefully step by step and double check the path to the correct solution for any mistakes. Provide then the final answer in
 following form: "\verb|#|\verb|#|\verb|#| Answer: ". & THINKING v2 / 5 & 273 \\
4 & Alice has 2 brothers and she also has 3 sisters. How many brothers does Alice's sister have? Before providing answer to this problem, think carefully step by step and double check the path to the correct solution for any mistakes. Provide then the final answer in following form: "\verb|#|\verb|#|\verb|#| Answer:". & THINKING v2 / 2 & 274 \\
\bottomrule
\end{tabular}
\end{table*}

\begin{table*}
\centering
\caption{AIW Light Arithmetic Total Girls variations 1-4}
\label{tab:aiw_light_arithmetic_total_girls}
\begin{tabular}{p{0.5cm}p{11cm}p{2cm}p{1cm}}
\toprule
Var. & Prompt & Type/Answer & ID \\
\midrule
1 & Alice has 6 sisters and she also has 3 brothers. How many girls are there in total? Before providing answer to this problem, think carefully step by step and double check the path to the correct solution for any mistakes. Provide then the final answer in
 following form: "\verb|#|\verb|#|\verb|#| Answer:". & THINKING v2 / 7 & 343 \\
2 & Alice has 2 sisters and she also has 4 brothers. How many girls are there in total? Before providing answer to this problem, think carefully step by step and double check the path to the correct solution for any mistakes. Provide then the final answer in
 following form: "\verb|#|\verb|#|\verb|#| Answer:". & THINKING v2 / 3 & 344 \\
3 & Alice has 4 sisters and she also has 1 brother. How many girls are there in total? Before providing answer to this problem, think carefully step by step and double check the path to the correct solution for any mistakes. Provide then the final answer in
 following form: "\verb|#|\verb|#|\verb|#| Answer: ". & THINKING v2 / 5 & 345 \\
4 & Alice has 1 sister and she also has 4 brothers. How many girls are there in total? Before providing answer to this problem, think carefully step by step and double check the path to the correct solution for any mistakes. Provide then the final answer in following form: "\verb|#|\verb|#|\verb|#| Answer:". & THINKING v2 / 2 & 346 \\
\bottomrule
\end{tabular}
\end{table*}

\begin{table*}
\centering
\caption{AIW Alice Female Power Boost and AIW Original, variations 1-4, THINKING v2 prompt}
\label{tab:aiw_female_boost_variation_types}
\begin{tabular}{p{0.5cm}p{11cm}p{2cm}p{1cm}}
\toprule
Var. & Prompt & Type/Answer & ID \\
\midrule
1 & \textit{Alice is female} and has 3 brothers and she also has 6 sisters.
 How many sisters does Alice's brother have? Before
 providing answer to this problem, think carefully step by step and double check the path to the correct solution
 for any mistakes. Provide then the final answer in
 following form: "\verb|#|\verb|#|\verb|#| Answer:
 ". & FEMALE BOOST / 7 & 193 \\
1 & Alice has 3 brothers and she also has 6 sisters.
 How many sisters does Alice's brother have? Before
 providing answer to this problem, think carefully step by step and double check the path to the correct solution
 for any mistakes. Provide then the final answer in
 following form: "\verb|#|\verb|#|\verb|#| Answer:
 ". & NO BOOST / 7 & 205 \\
 2 & \textit{Alice is female} and has 2 sisters and she also has 4 brothers.
 How many sisters does Alice's brother have? Before
 providing answer to this problem, think carefully step by step and double check the path to the correct solution
 for any mistakes. Provide then the final answer in
 following form: "\verb|#|\verb|#|\verb|#| Answer:
 ". & FEMALE BOOST / 3 & 197 \\
2 & Alice has 2 sisters and she also has 4 brothers.
 How many sisters does Alice's brother have? Before
 providing answer to this problem, think carefully step by step and double check the path to the correct solution
 for any mistakes. Provide then the final answer in
 following form: "\verb|#|\verb|#|\verb|#| Answer:
 ". & NO BOOST / 3 & 206 \\
 3 & \textit{Alice is female} and has 4 sisters and she also has 1 brother.
 How many sisters does Alice's brother have? Before
 providing answer to this problem, think carefully step by step and double check the path to the correct solution
 for any mistakes. Provide then the final answer in
 following form: "\verb|#|\verb|#|\verb|#| Answer:
 ". & FEMALE BOOST / 5 & 189 \\
3 & Alice has 4 sisters and she also has 1 brother.
 How many sisters does Alice's brother have? Before
 providing answer to this problem, think carefully step by step and double check the path to the correct solution
 for any mistakes. Provide then the final answer in
 following form: "\verb|#|\verb|#|\verb|#| Answer:
 ". & NO BOOST / 5 & 187 \\
4 & \textit{Alice is female} and has 4 brothers and she also has 1 sister.
 How many sisters does Alice's brother have? Before
 providing answer to this problem, think carefully step by step and double check the path to the correct solution
 for any mistakes. Provide then the final answer in
 following form: "\verb|#|\verb|#|\verb|#| Answer:
 ". & FEMALE BOOST / 2 & 190 \\
4 & Alice has 4 brothers and she also has 1 sister.
 How many sisters does Alice's brother have? Before
 providing answer to this problem, think carefully step by step and double check the path to the correct solution
 for any mistakes. Provide then the final answer in
 following form: "\verb|#|\verb|#|\verb|#| Answer:
 ". & NO BOOST / 2 & 188 \\
\bottomrule
\end{tabular}
\end{table*}

\begin{table*}
\centering
\caption{AIW Exaggerated Numbers variations 1-4, Thinking v2 prompt type.}
\label{tab:aiw_exaggerated_num_prompts}
\begin{tabular}{p{0.5cm}p{11cm}p{2cm}p{1cm}}
\toprule
Var. & Prompt & Type/Answer & ID \\
\midrule
1 & Alice has 63 brothers and she also has 66 sisters. How many sisters does Alice's brother have? Before providing answer to this problem, think carefully step by step and double check the path to the correct solution for any mistakes. Provide then the final answer in following form: "\verb|#|\verb|#|\verb|#| Answer:". & THINKING v2 / 67 & 653 \\
2 & Alice has 62 sisters and she also has 64 brothers. How many sisters does Alice's brother have? Before providing answer to this problem, think carefully step by step and double check the path to the correct solution for any mistakes. Provide then the final answer in following form: "\verb|#|\verb|#|\verb|#| Answer:". & THINKING v2 / 63 & 654 \\
3 & Alice has 64 sisters and she also has 61 brother. How many sisters does Alice's brother have? Before providing answer to this problem, think carefully step by step and double check the path to the correct solution for any mistakes. Provide then the final answer in following form: "\verb|#|\verb|#|\verb|#| Answer:". & THINKING v2 / 65 & 655 \\
4 & Alice has 64 brothers and she also has 61 sister. How many sisters does Alice's brother have? Before providing answer to this problem, think carefully step by step and double check the path to the correct solution for any mistakes. Provide then the final answer in following form: "\verb|#|\verb|#|\verb|#| Answer:". & THINKING v2 / 62 & 656 \\
\bottomrule
\end{tabular}
\end{table*}

\begin{table*}
\centering
\caption{AIW Ext variations 1-4, Thinking v2 prompt type.}
\label{tab:aiw_ext_prompts}
\begin{tabular}{p{0.5cm}p{11cm}p{2cm}p{1cm}}
\toprule
Var. & Prompt & Type/Answer & ID \\
\midrule
1 & Alice and Bob are sister and brother. Alice has 3 sisters and Bob has 6 brothers. How many brothers does Alice have? Before providing answer to this problem, think carefully step by step and double check the path to the correct solution for any mistakes. Provide then the final answer in following form: "\verb|#|\verb|#|\verb|#| Answer:
 ". & THINKING v2 / 7 & 264 \\
2 & Alice and Bob are sister and brother. Alice has 2 sisters and Bob has 2 brothers. How many brothers does Alice have? Before providing answer to this problem, think carefully step by step and double check the path to the correct solution for any mistakes. Provide then the final answer in following form: "\verb|#|\verb|#|\verb|#| Answer:
 ". & THINKING v2 / 3 & 266 \\
3 & Alice and Bob are sister and brother. Alice has 1 sister and Bob has 4 brothers. How many brothers does Alice have? Before providing answer to this problem, think carefully step by step and double check the path to the correct solution for any mistakes. Provide then the final answer in
 following form: "\verb|#|\verb|#|\verb|#| Answer: ". & THINKING v2 / 5 & 268 \\
4 & Alice and Bob are sister and brother. Alice has 3 sisters and Bob has 1 brother. How many brothers does Alice have? Before providing answer to this problem, think carefully step by step and double check the path to the correct solution for any mistakes. Provide then the final answer in following form: "\verb|#|\verb|#|\verb|#| Answer:". & THINKING v2 / 2 & 270 \\
5 & Alice and Bob are sister and brother. Alice has 2 sisters and Bob has 3 brothers. How many brothers does Alice have? Before providing answer to this problem, think carefully step by step and double check the path to the correct solution for any mistakes. Provide then the final answer in following form: "\verb|#|\verb|#|\verb|#| Answer:". & THINKING v2 / 4 & 455 \\
6 & Alice and Bob are sister and brother. Alice has 3 sisters and Bob has 5 brothers. How many brothers does Alice have? Before providing answer to this problem, think carefully step by step and double check the path to the correct solution for any mistakes. Provide then the final answer in following form: "\verb|#|\verb|#|\verb|#| Answer:". & THINKING v2 / 6 & 456 \\
\bottomrule
\end{tabular}
\end{table*}

\begin{table*}
\centering
\caption{AIW Friends variations 1-4}
\label{tab:aiw_friends_prompts}
\begin{tabular}{p{0.5cm}p{11cm}p{2cm}p{1cm}}
\toprule
Var. & Prompt & Type/Answer & ID \\
\midrule
1 & Alice has 3 male friends and she also has 6 female friends. All mentioned persons are friends with each other and have no other friends aside. How many female friends does male friend of Alice have? Before providing answer to this problem, think carefully step by step and double check the path to the correct solution for any mistakes. Provide then the final answer in following form: "\verb|#|\verb|#|\verb|#| Answer:
 ". & THINKING v2 / 7 & 577 \\
2 & Alice has 2 female friends and she also has 4 male friends. All mentioned persons are friends with each other and have no other friends aside. How many female friends does male friend of Alice have? Before providing answer to this problem, think carefully step by step and double check the path to the correct solution for any mistakes. Provide then the final answer in following form: "\verb|#|\verb|#|\verb|#| Answer:
 ". & THINKING v2 / 3 & 580 \\
3 & Alice has 4 female friends and she also has 1 male friend. All mentioned persons are friends with each other and have no other friends aside. How many female friends does male friend of Alice have? Before providing answer to this problem, think carefully step by step and double check the path to the correct solution for any mistakes. Provide then the final answer in
 following form: "\verb|#|\verb|#|\verb|#| Answer: ". & THINKING v2 / 5 & 581 \\
4 & Alice has 4 male friends and she also has 1 female friend. All mentioned persons are friends with each other and have no other friends aside. How many female friends does male friend of Alice have? Before providing answer to this problem, think carefully step by step and double check the path to the correct solution for any mistakes. Provide then the final answer in following form: "\verb|#|\verb|#|\verb|#| Answer:". & THINKING v2 / 2 & 582 \\
5 & Alice has 2 male friends and she also has 3 female friends. All mentioned persons are friends with each other and have no other friends aside. How many female friends does male friend of Alice have? Before providing answer to this problem, think carefully step by step and double check the path to the correct solution for any mistakes. Provide then the final answer in following form: "\verb|#|\verb|#|\verb|#| Answer:". & THINKING v2 / 4 & 583 \\
6 & Alice has 5 female friends and she also has 3 male friends. All mentioned persons are friends with each other and have no other friends aside. How many female friends does male friend of Alice have? Before providing answer to this problem, think carefully step by step and double check the path to the correct solution for any mistakes. Provide then the final answer in following form: "\verb|#|\verb|#|\verb|#| Answer:". & THINKING v2 / 6 & 584 \\
\bottomrule
\end{tabular}
\end{table*}

\begin{table}
\caption{AIW Plus variations 1-6.}
\label{tab:aiw_plus_prompt_types}
\centering
\begin{tabular}{p{1cm}p{9cm}p{2cm}p{1cm}p{1cm}}
\toprule
Var. & Prompt & Type/Answer & ID \\
\midrule
1 & Alice has 1 sister and 1 brother in total. Her mother has 2 brothers. She also has 1 sister who does not have children and who has 6 nephews and nieces in total. Alice's father has 2 sisters. He also has a brother who has 5 nephews and nieces in total, and who also has 2 sons. How many cousins does Alice's sister have? Before providing answer to this problem, think carefully step by step and double check the path to the correct solution for any mistakes. Provide then the final answer in following form: "\verb|#|\verb|#|\verb|#| Answer: ". & THINKING V2/7 & 559 \\
2 & Alice has 2 sisters and 1 brother in total. Her mother has 2 brothers. She also has 1 sister who does not have children and who has 6 nephews and nieces in total. Alice's father has 2 sisters. He also has a brother who has 4 nephews and nieces in total, and who also has 1 son. How many cousins does Alice's sister have? Before providing answer to this problem, think carefully step by step and double check the path to the correct solution for any mistakes. Provide then the final answer in following form: "\verb|#|\verb|#|\verb|#| Answer: ". & THINKING V2/3 & 560 \\
3 & Alice has 2 sisters and 1 brother in total. Her mother has 2 brothers. She also has 1 sister who does not have children and who has 7 nephews and nieces in total. Alice's father has 2 sisters. He also has a brother who has 5 nephews and nieces in total, and who also has 1 son. How many cousins does Alice's sister have? Before providing answer to this problem, think carefully step by step and double check the path to the correct solution for any mistakes. Provide then the final answer in following form: "\verb|#|\verb|#|\verb|#| Answer: ". & THINKING V2/5 & 561 \\
4 & Alice has 1 sister and 3 brothers in total. Her mother has 2 brothers. She also has 1 sister who does not have children and who has 6 nephews and nieces in total. Alice's father has 2 sisters. He also has a brother who has 5 nephews and nieces in total, and who also has 1 daughter. How many cousins does Alice's sister have? Before providing answer to this problem, think carefully step by step and double check the path to the correct solution for any mistakes. Provide then the final answer in following form: "\verb|#|\verb|#|\verb|#| Answer: ". & THINKING V2/2 & 562 \\
5 & Alice has 2 sisters and 1 brother in total. Her mother has 2 brothers. She also has 1 sister who does not have children and who has 6 nephews and nieces in total. Alice's father has 2 sisters. He also has a brother who has 5 nephews and nieces in total, and who also has 1 son. How many cousins does Alice's sister have? Before providing answer to this problem, think carefully step by step and double check the path to the correct solution for any mistakes. Provide then the final answer in following form: "\verb|#|\verb|#|\verb|#| Answer: ". & THINKING V2/4 & 563 \\
6 & Alice has 1 sister and 1 brother in total. Her mother has 2 brothers. She also has 1 sister who does not have children and who has 6 nephews and nieces in total. Alice's father has 2 sisters. He also has a brother who has 5 nephews and nieces in total, and who also has 1 daughter. How many cousins does Alice's sister have? Before providing answer to this problem, think carefully step by step and double check the path to the correct solution for any mistakes. Provide then the final answer in following form: "\verb|#|\verb|#|\verb|#| Answer: ". & THINKING V2/6 & 564 \\
\bottomrule
\end{tabular}
\end{table}

\begin{table}
\caption{AIW Circles Colleagues variations (omitting variations 2-5 for better readability)}
\label{tab:aiw_circles_prompt_types}
\centering
\begin{tabular}{p{1cm}p{9cm}p{2cm}p{1cm}p{1cm}}
\toprule
Var. & Prompt & Type/Answer & ID \\
\midrule
1 & Alice has 3 male colleagues and she also has 6 female colleagues in total. All these mentioned persons in the circle around Alice are colleagues of each other. Bob has 2 female colleagues and 1 male colleague in total. All these mentioned persons in the circle around Bob are colleagues of each other. The people in the circle around Bob do not have other colleagues aside - with the only exception of Matilda. She is colleague of Bob, being part of Bob's circle, and she is also colleague of Alice, being part of Alice's circle. All the mentioned persons have no colleagues beyond the already described group of people. How many female colleagues does Matilda have? Before providing answer to this problem, think carefully step by step and double check the path to the correct solution for any mistakes. Provide then the final answer in following form: "\verb|#|\verb|#|\verb|#| Answer: ". & THINKING V2/7 & 637 \\
2 &  ... & THINKING V2/3 & 638 \\
3 &  ... & THINKING V2/5 & 639 \\
4 &  ... & THINKING V2/2 & 640 \\
5 &  ... & THINKING V2/4 & 641 \\
6 & Alice has 3 male colleagues and she also has 5 female colleagues in total. All these mentioned persons in the circle around Alice are colleagues of each other. Bob has 2 female colleagues and 1 male colleague in total. All these mentioned persons in the circle around Bob are colleagues of each other. The people in the circle around Bob do not have other colleagues aside - with the only exception of Matilda. She is colleague of Bob, being part of Bob's circle, and she is also colleague of Alice, being part of Alice's circle. All the mentioned persons have no colleagues beyond the already described group of people. How many female colleagues does Matilda have? Before providing answer to this problem, think carefully step by step and double check the path to the correct solution for any mistakes. Provide then the final answer in following form: "\verb|#|\verb|#|\verb|#| Answer: ". & THINKING V2/6 & 642 \\
\bottomrule
\end{tabular}
\end{table}

\section{Model performance and behavior on AIW problem versions}
\label{appendix:aiw_problem}

Here we report further details on model evaluation, performance and behavior as observed on AIW problems. For executing experiments, we either use local model deployment via vLLM~\cite{vllm_2024}, or API based liteLLM~\cite{litellm_2024} and TogetherAI~\cite{togetherai_2024}. 

\begin{figure}
    \centering
    \includegraphics[width=\textwidth]{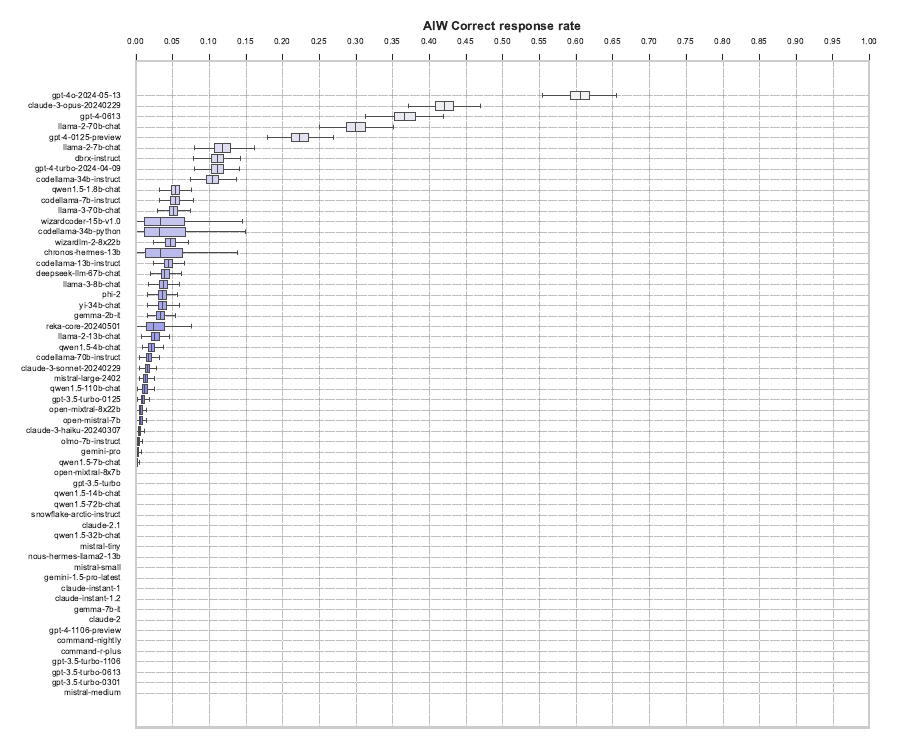}
    \vspace{-0.8cm}
    \caption{Collapse of most SOTA LLMs on AIW problem. AIW correct response rate across AIW variations averaged across all prompt types THINKING, STANDARD and RESTRICTED. Only 5 models manage to show rates above $p=0.2$: GPT-4o, Claude 3 Opus, GPT-4-0613, Llama 2 70B Chat and GPT-4-0125-preview (GPT4-Turbo). Llama 2 70B Chat is the only open-weights model in this set. The rest either shows poor performance below $p=0.15$, or even collapses entirely to 0. Among those models collapsing to 0 are many which claim strong function via high scores obtained on standardized benchmarks, eg larger scale GPT-3.5, Mixtral 8x7B and 8x22B, Command R Plus, Qwen 1.5 72B Chat and smaller scale Gemma-7b-it, Mistral Small and Mistral Medium.}
    \label{fig:aiw_all}
    \vspace{-0.3cm}
\end{figure}

\begin{figure*}[t!]
    \begin{subfigure}{0.48\textwidth}
       \centering
        \includegraphics[width=1.0\textwidth]{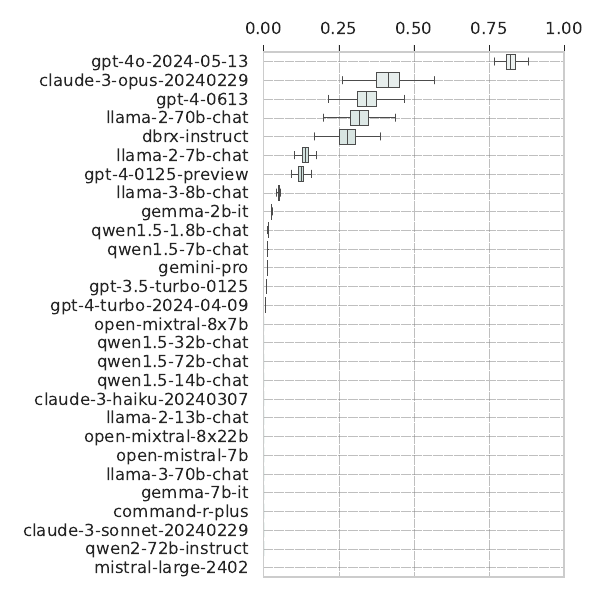}
        \caption{Correct response rates for RESTRICTED prompt type, averaged across AIW var 1-4}
        \label{subfig:aiw_restricted_avg}
    \end{subfigure}
    \begin{subfigure}{0.5\textwidth}
       \centering
        \includegraphics[width=1.0\textwidth]{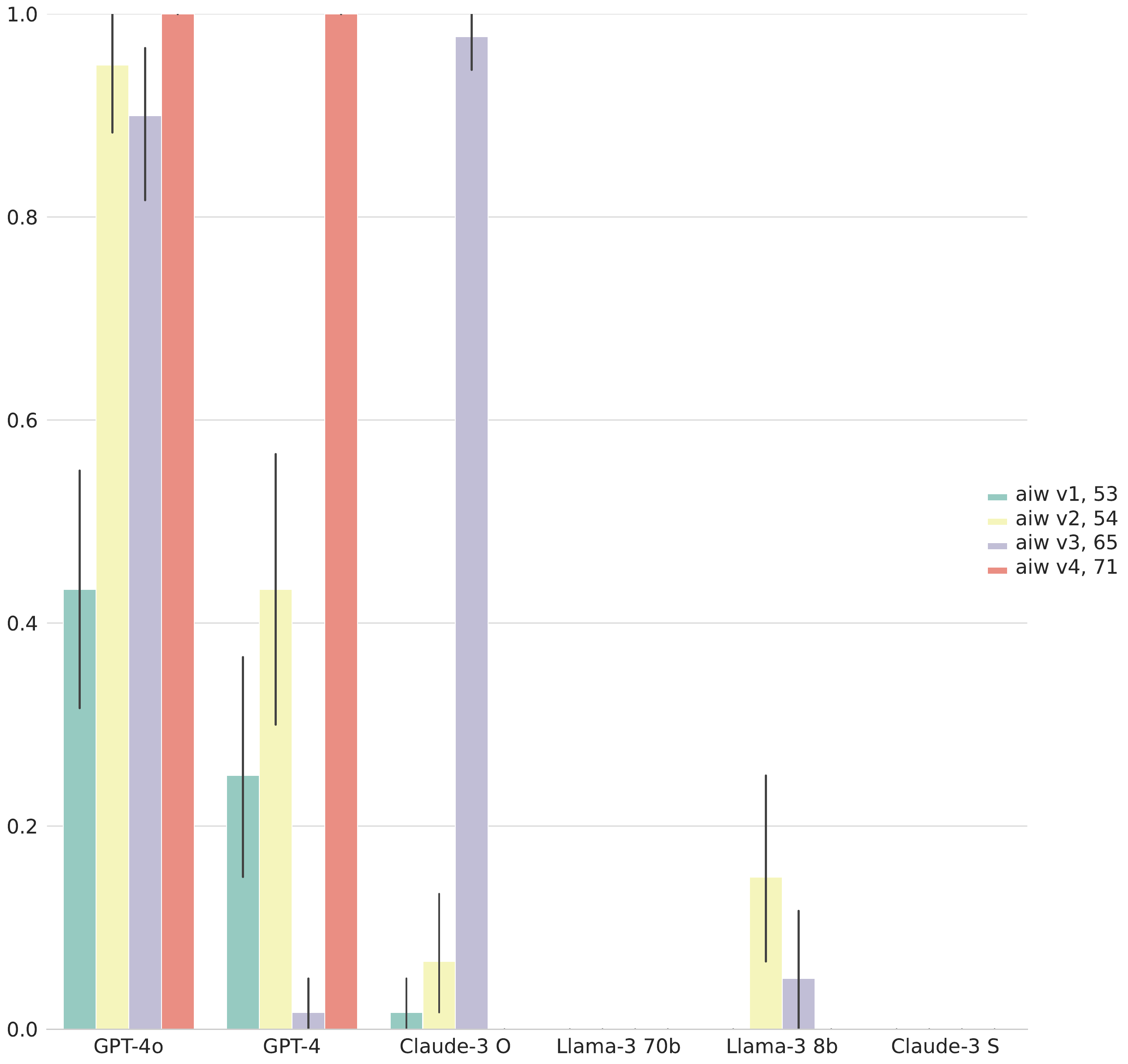}
        \caption{Strong fluctuations across AIW variations 1-4, RESTRICTED prompt type}
        \label{subfig:aiw_restricted_fluctuations}
    \end{subfigure}    
    \caption{Correct response rates on RESTRICTED prompt type. The prompt type enforcing to output only final answer without any further text was used as further control. \textbf{(a)} Correct response rates averaged over variations 1-4 resemble behavior with STANDARD and THINKING types, while looking at fluctuations across variations 1-4 in \textbf{(b)} reveals stronger models' lack of robustness compared to other prompt types (see for comparison Fig. \ref{fig:aiw_original_fluctuations}). We thus used THINKING prompt types across main experiment not to put models into disadvantage on AIW testing.}
    \label{fig:aiw_restricted}
\end{figure*}

\begin{figure*}[t!]
    \begin{subfigure}{0.48\textwidth}
       \centering
        \includegraphics[width=1.0\textwidth]{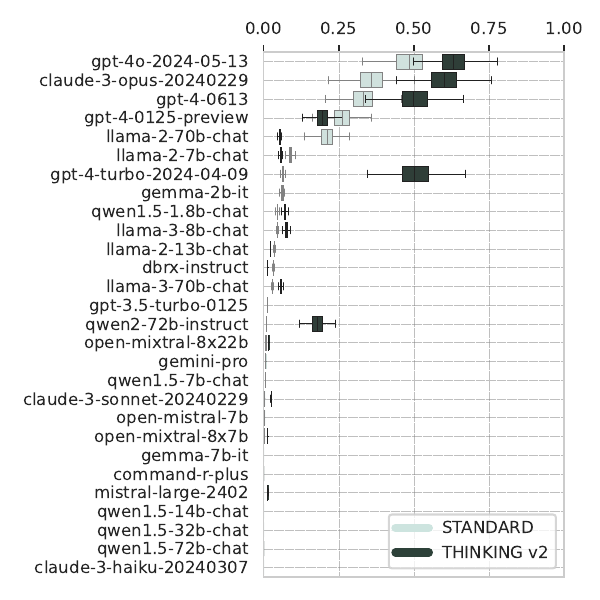}
        \caption{Correct response rates THINKING v2 vs. STANDARD prompt type, averaged across AIW var 1-4}
        \label{subfig:aiw_thinking_v2_vs_standard}
    \end{subfigure}
    \begin{subfigure}{0.5\textwidth}
       \centering
        \includegraphics[width=1.0\textwidth]{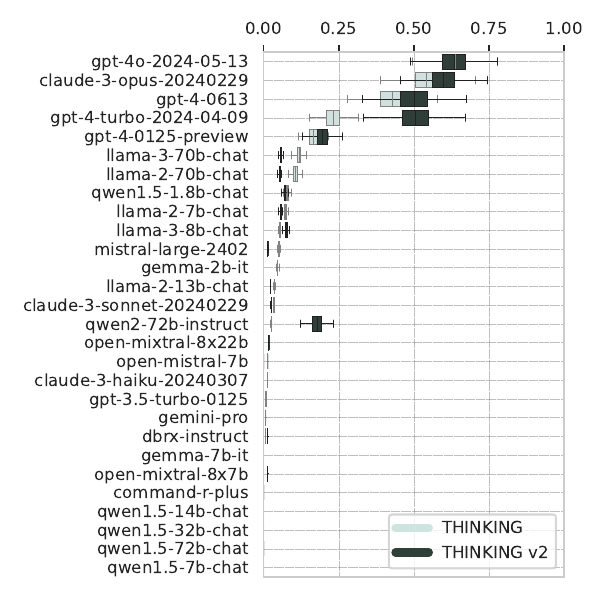}
        \caption{Correct response rates THINKING v2 vs. THINKING prompt type, averaged across AIW var 1-4}
        \label{subfig:aiw_thinking_v2_vs_thinking}
    \end{subfigure}    
    \caption{Control comparison of correct response rates averaged across AIW variations 1-4. \textbf{(a)} THINKING v2 vs. STANDARD, \textbf{(b)} THINKING v2 vs. THINKING prompt types. THINKING provides better average correct response rates for tested models. We thus used THINKING prompt types for main and control experiments to ensure tested models are not disadvantaged on AIW problem. THINKING and THINKING v2 show highly similar behavior across tested models \textbf{(b)} and can be used interchangeably (THINKING v2 only difference to THINKING is the explicit phrasing "step by step",  Suppl Tab. \ref{tab:aiw_female_boost_variation_types})}
    \label{fig:aiw_thinking_v2_comparison}
\end{figure*}

For the full overview of average correct response rate including models that score zero, see Suppl. Fig. \ref{fig:aiw_all}. For the statistics on number of trials conducted for each model and each prompt type, see Suppl. Fig. \ref{fig:num_of_responses_aiw}. For the statistics on the average output length across models and prompt types, see Suppl. Fig. \ref{fig:avg_len_of_responses_aiw}. For models' behavior on RESTRICT prompt type, see Suppl. Fig. \ref{fig:aiw_restricted}.  For control comparison of THINKING v2 prompt type to THINKING and STANDARD, see see Suppl. Fig. \ref{fig:aiw_thinking_v2_comparison}. 
For fluctuations on different AIW Ext versions that provide further evidence for strong models' sensitivity to slight problem perturbations, see 
Suppl. Fig. \ref{fig:aiw_ext_vs_alice_bob_questions_fluctuations}

\subsection{Standardized benchmarks failure.}
\label{appendix:benchs_failure}

\begin{figure*}
    \centering
    \includegraphics[width=\textwidth]{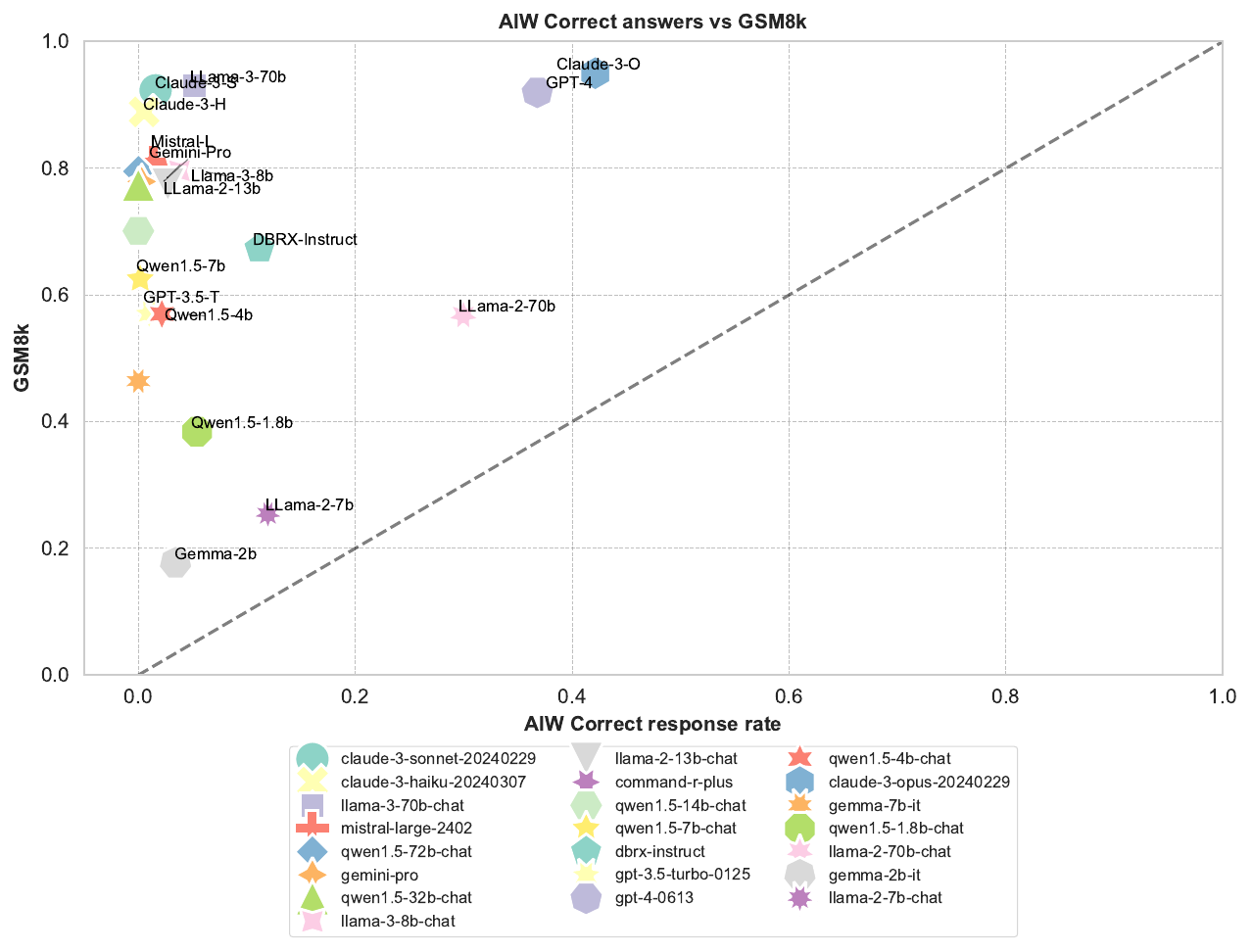}
    \vspace{-0.3cm}
    \caption{Limitation of the standardized benchmark GSM8k in accurately reflecting and comparing basic reasoning capabilities of models, as illustrated by the stark discrepancy between the AIW correct response rate and the GSM8k average score. Notably, the majority of tested models exhibit low performance on AIW problems while achieving relatively high scores on GSM8k, a graduate-level math benchmark for large language models. Among models with slightly better calibration are Claude Opus and GPT 4 that outperform other models on AIW, which coincides with their high GSM8k scores. Llama 2 70b also shows better calibration, where its modest AIW performance matches its modest GSM8k score. In contrast, models like Mistral Large, Gemini Pro, Dbrx Instruct, or Command R+, while scoring high on GSM8k, show breakdown on AIW (Command R+ has 0 correct response rate, Mistral Large and Gemini Pro $0.01$, Dbrx Instruct $0.11$, see also Suppl Tab. \ref{tab:bench_aiw})  }
    \label{fig:gsm8k_benchmark_vs_aiw}
    \vspace{-0.3cm}
\end{figure*}

\begin{figure*}
    \centering
    \includegraphics[width=\textwidth]{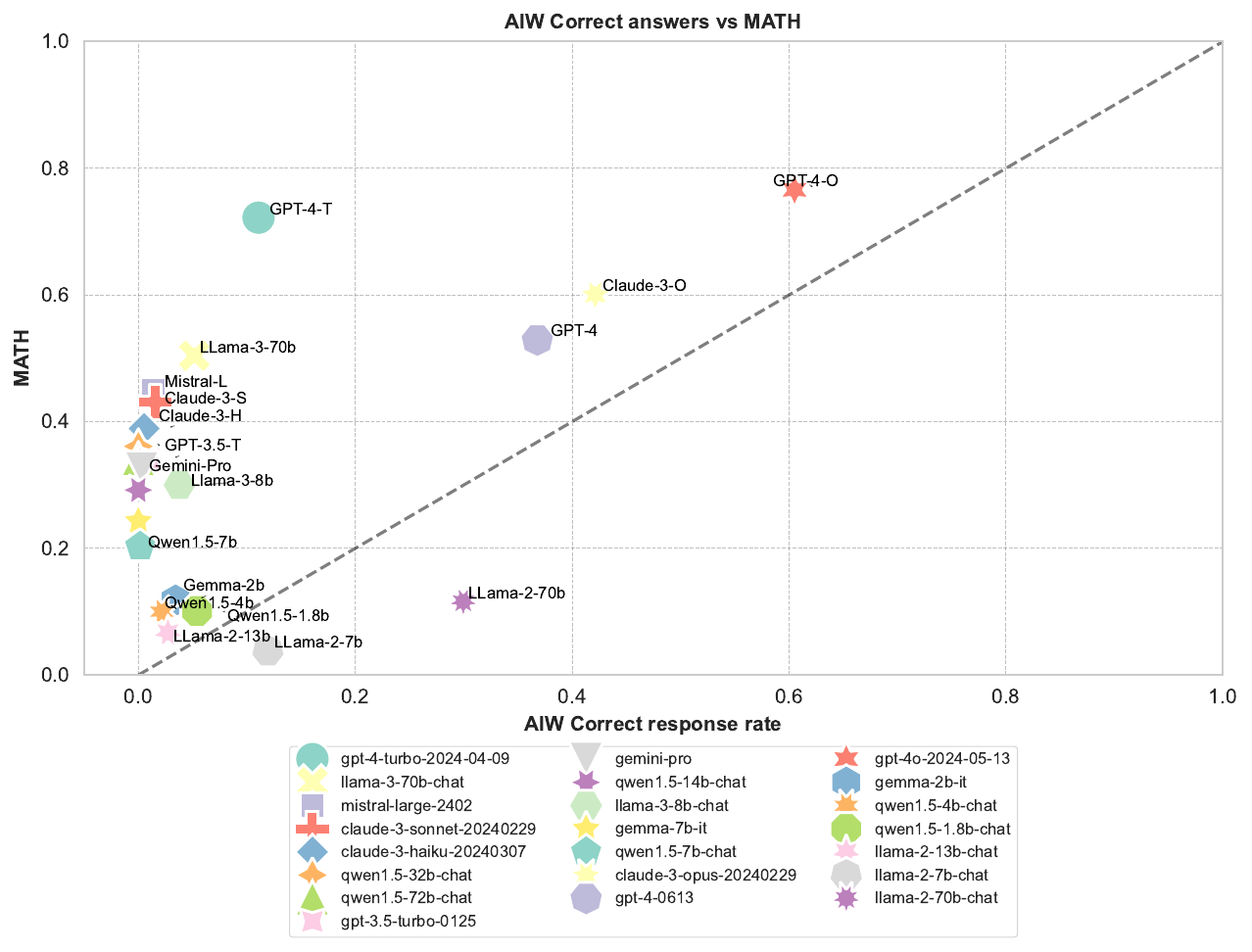}
    \vspace{-0.3cm}
    \caption{Discrepancy between the AIW correct response rate and the MATH average score, indicating the limitation of standardized benchmark MATH in accurately assessing and comparing basic reasoning capabilities of models. Numerous models, such as Command R+, exhibit a stark contrast in performance, scoring zero on AIW while achieving high scores on MATH.}
    \label{fig:math_benchmark_vs_aiw}
    \vspace{-0.3cm}
\end{figure*}

\begin{figure*}
    \centering
    \includegraphics[width=\textwidth]{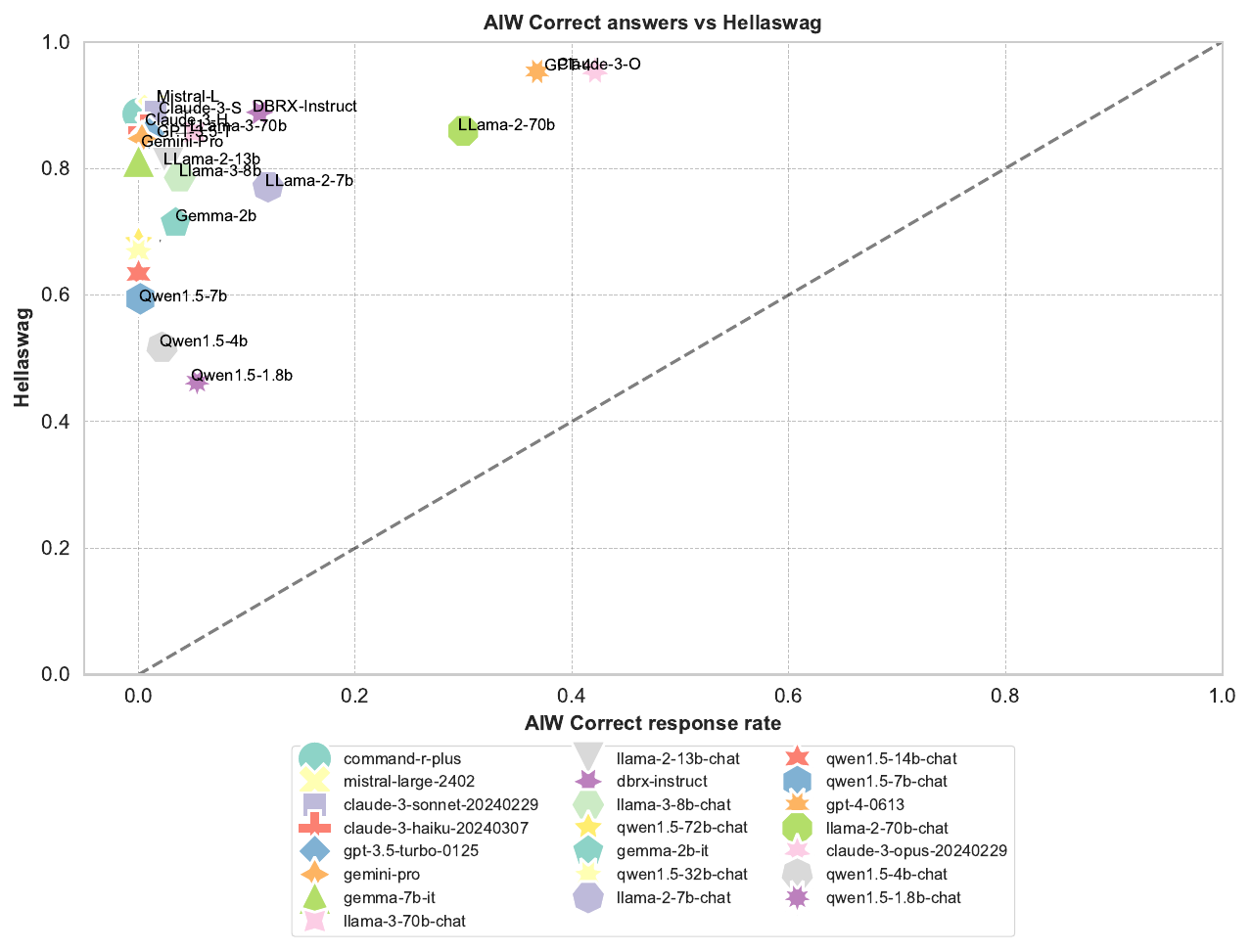}
    \vspace{-0.3cm}
    \caption{Limitation of the standardized benchmark Hellaswag in accurately assessing and comparing basic reasoning capabilities of models, as evidenced by the significant discrepancy between the AIW correct response rate and the Hellaswag average score. }
    \label{fig:hellaswag_benchmark_vs_aiw}
    \vspace{-0.3cm}
\end{figure*}

\begin{figure*}
    \centering
    \includegraphics[width=\textwidth]{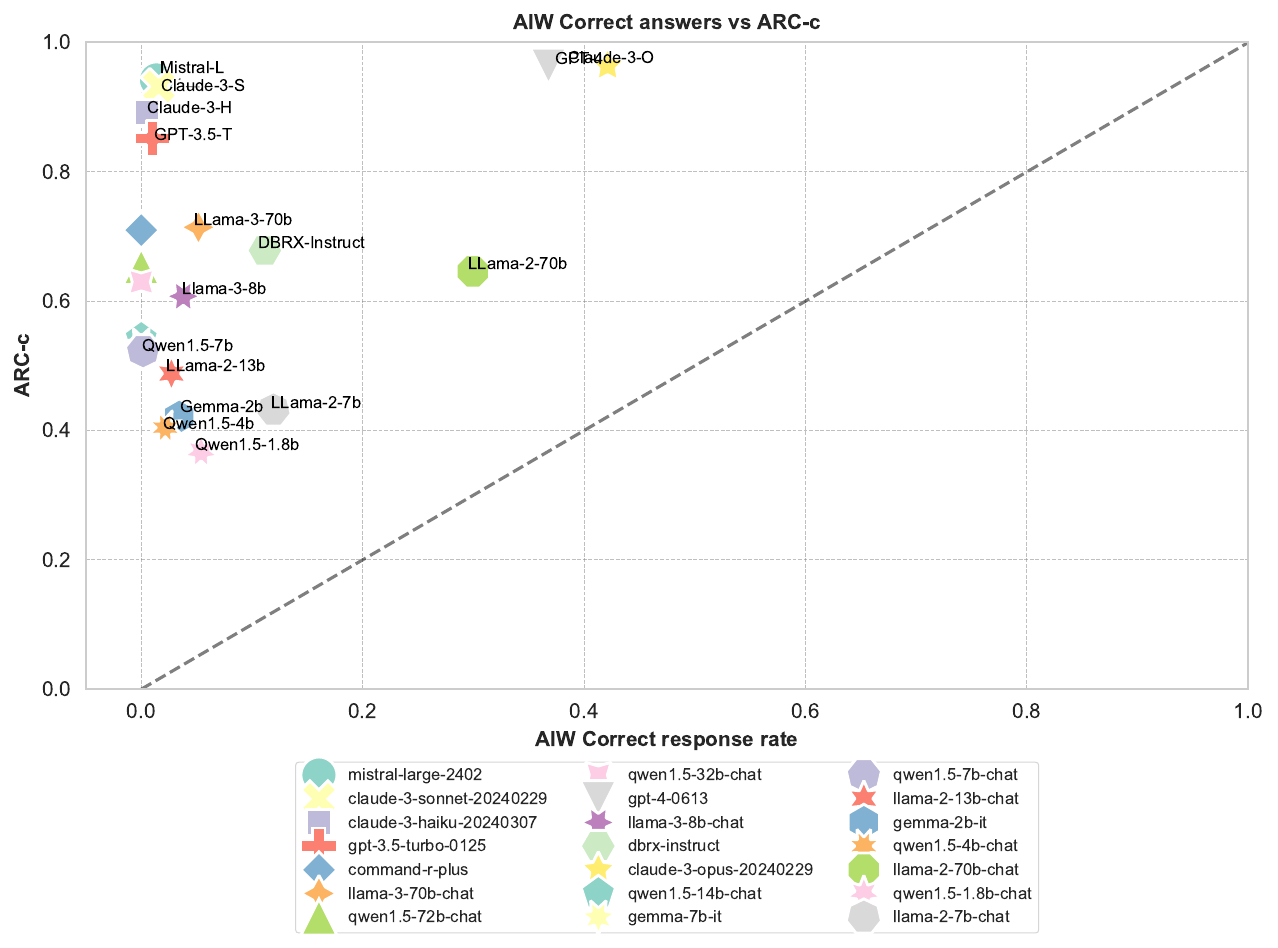}
    \vspace{-0.8cm}
    \caption{Failure of standardized benchmark ARC-c to properly reflect and compare model basic reasoning capabilities as shown by strong discrepancy between AIW correct response rate vs ARC-c average score. }
    \label{fig:arc_benchmark_vs_aiw}
    \vspace{-0.3cm}
\end{figure*}

In Section~\ref{subsec:discrepancies}, we observe failure of standardized reasoning benchmarks to properly reflect generalization and basic reasoning skills of SOTA LLMs by noting significant disparity between the model's performance on the AIW problem and the outcomes on conventional standardized benchmarks, taking MMLU as representative examples. Here, we confirm this finding on further standardized reasoning benchmarks like MATH, ARC-c, GSM8K and Hellaswag (Suppl Tab. \ref{tab:bench_aiw}).  We provide plots visualizing failure of these standardized benchmarks, reflected in strong mismatch between high benchmark scores reported by many models and the low correct response rates they obtain on AIW (which in some cases is 0 for models with high standardized benchmark scores), in Figures \ref{fig:math_benchmark_vs_aiw}, \ref{fig:arc_benchmark_vs_aiw}, \ref{fig:gsm8k_benchmark_vs_aiw}, \ref{fig:hellaswag_benchmark_vs_aiw}.

We see thus that standardized benchmarks fail to properly reflect true model capabilities to generalize and reason - the majority of the tested models score high on standardized benchmarks, suggesting strong function, while showing extreme low correct response rates on simple AIW problem. Many of the models with high scores on standardized benchmarks cannot solve AIW problem a single time (e.g. Command R+ is unable to solve a single AIW problem instance, see Suppl Tab. \ref{tab:bench_aiw}). This discrepancy refutes the claim of standardized benchmarks to measure correctly current models' core functionality. 

\begin{table*}
\centering
\caption{Performance of tested models on MMLU, Hellaswag, ARC-c, GSM8k and AIW. Correct response rate averaged across AIW variations 1-4, across STANDARD, THINKING and RESTRICTED prompt types.}    
\label{tab:bench_aiw}
\begin{tabular}{p{3.9cm}p{1.5cm}p{1.5cm}p{1.5cm}p{1.5cm}p{1.5cm}p{1.5cm}}
\toprule
Model & MMLU & Hellaswag & ARC-c & GSM8k & AIW Correct average resp. rate \\
\midrule
gpt-4o-2024-05-13 & 0.89 & - & - & - & 0.65 \\
claude-3-opus-20240229 & 0.87 & 95.40 & 96.40 & 95.00 & 0.43 \\
gpt-4-0613 & 0.86 & 95.30 & 96.30 & 92.00 & 0.37 \\
llama-2-70b-chat & 0.64 & 85.90 & 64.60 & 56.80 & 0.30 \\
llama-2-7b-chat & 0.55 & 77.10 & 43.20 & 25.40 & 0.13 \\
dbrx-instruct & 0.74 & 88.85 & 67.83 & 67.32 & 0.11 \\
gpt-4-turbo-2024-04-09 & 0.80 & - & - & - & 0.10 \\
llama-3-8b-chat & 0.67 & 78.55 & 60.75 & 79.60 & 0.05 \\
llama-3-70b-chat & 0.80 & 85.69 & 71.42 & 93.00 & 0.05 \\
qwen1.5-1.8b-chat & 0.46 & 46.25 & 36.69 & 38.40 & 0.05 \\
gemma-2b-it & 0.38 & 71.40 & 42.10 & 17.70 & 0.04 \\
llama-2-13b-chat & 0.66 & 80.70 & 48.80 & 77.40 & 0.03 \\
qwen1.5-4b-chat & 0.56 & 51.70 & 40.44 & 57.00 & 0.02 \\
claude-3-sonnet-20240229 & 0.79 & 89.00 & 93.20 & 92.30 & 0.01\\
mistral-large-2402 & 0.81 & 89.20 & 94.20 & 81.00 & 0.01 \\
gpt-3.5-turbo-0125 & 0.70 & 85.50 & 85.20 & 57.10 & 0.01\\
gemini-pro & 0.72 & 84.70 & - & 77.90 & 0.01 \\
open-mixtral-8x22b & 0.78 & 89.08 & 72.70 & 82.03 & 0.01 \\
open-mistral-7b & 0.64 & 84.88 & 63.14 & 40.03 & 0.01 \\
qwen1.5-7b-chat & 0.62 & 59.38 & 52.30 & 62.50 & 0.01 \\
claude-3-haiku-20240307 & 0.75 & 85.90 & 89.20 & 88.90 & 0.00\\
open-mixtral-8x7b & 0.72 & 87.55 & 70.22 & 61.11 & 0.00 \\
command-r-plus & 0.76 & 88.56 & 70.99 & 70.74 & 0.00 \\
qwen1.5-14b-chat & 0.69 & 63.32 & 54.27 & 70.10 & 0.00 \\
gemma-7b-it & 0.54 & 81.20 & 53.20 & 46.40 & 0.00 \\
qwen1.5-72b-chat & 0.77 & 68.37 & 65.36 & 79.50 & 0.00 \\
qwen1.5-32b-chat & 0.75 & 66.84 & 62.97 & 77.40 & 0.00 \\
\bottomrule
\end{tabular}
\end{table*}

\subsection{Boosting by redundant information and persisting fluctuations: Alice female power boost}
\label{appendix:subsec:aiw_female_power_boost}

\begin{figure*}[t!]
    \centering
    \includegraphics[width=0.8\textwidth]{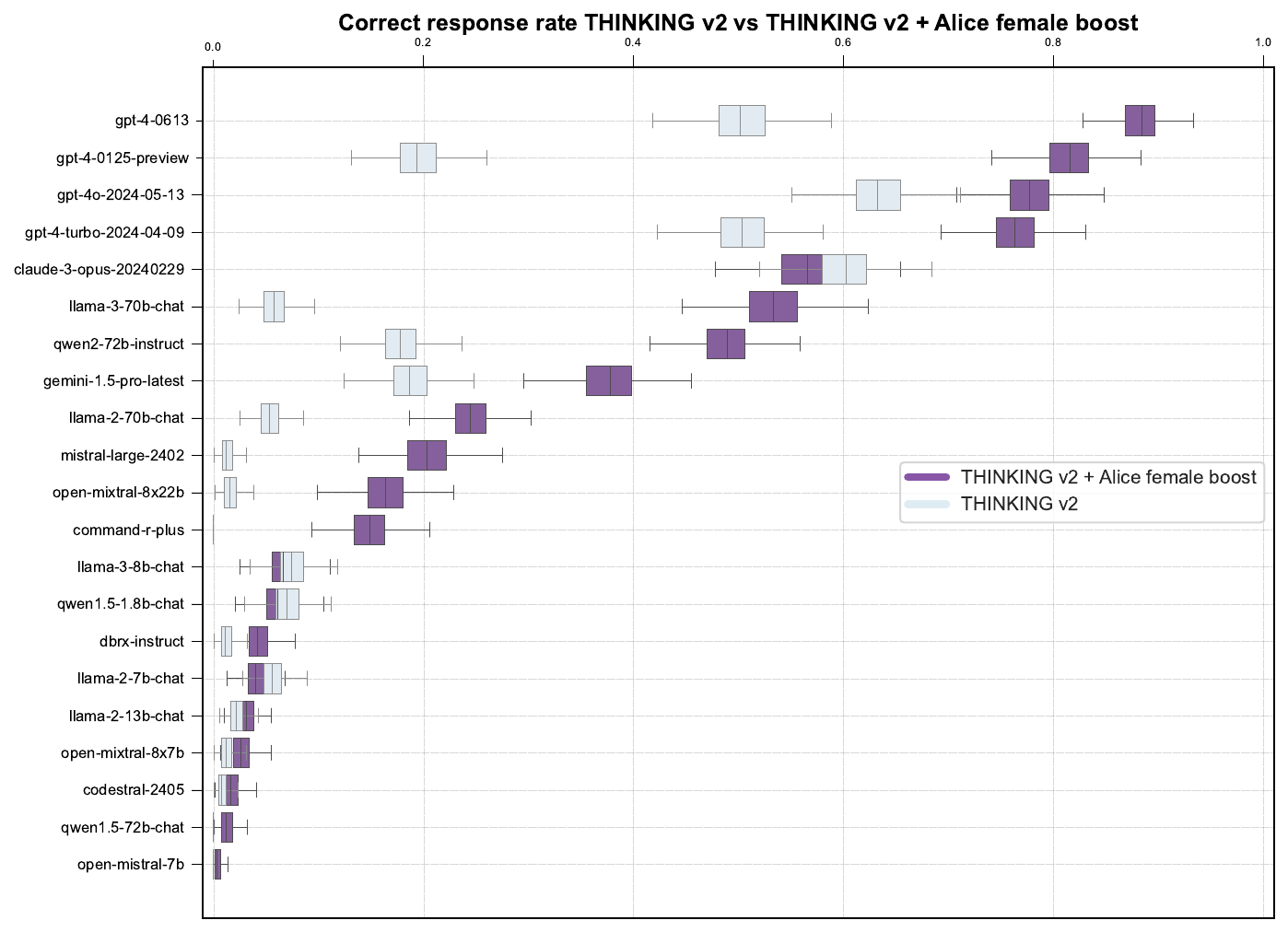}
    \caption{AIW "Alice Female Power Boost" version. Average correct response rate (measured across AIW variations 1-4) increases after addition of entirely redundant information "Alice is female" (pronoun "she" already fully indicates the gender in original AIW). Thinking v2 prompt type is used for both AIW versions. See also Fig. \ref{fig:aiw_alice_thinkingv2_original_female_power_boost_fluctuations} for persisting strong fluctuations across variations 1-4.} 
    \label{fig:aiw_original_alice_female_power_boost_average_correct_response_rates}
\end{figure*}

\begin{figure}[t!]
    \centering
    \begin{subfigure}{1.0\textwidth}
        \includegraphics[width=1.0\textwidth]{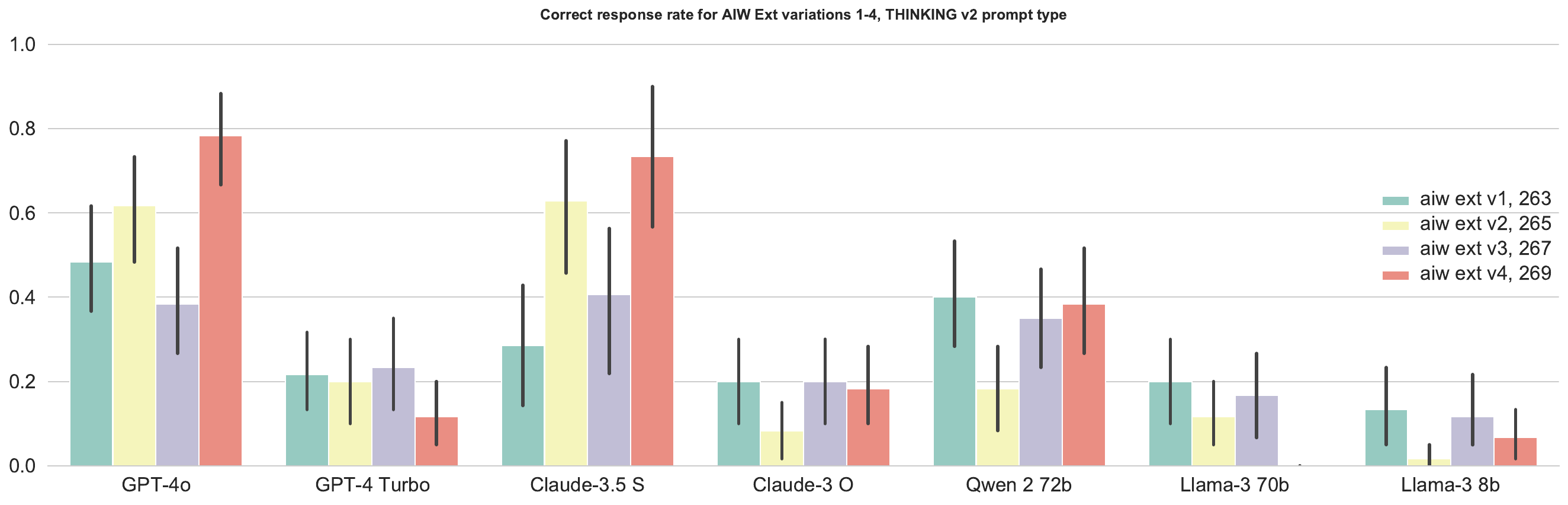}
        \caption{Correct response rates for AIW Ext Bob's Sisters question, THINKING v2 prompt type}
        \label{subfig:aiw_ext_Bobs_Sisters_thinkingv2}
    \end{subfigure}
    \begin{subfigure}{1.0\textwidth}
        \includegraphics[width=1.0\textwidth]{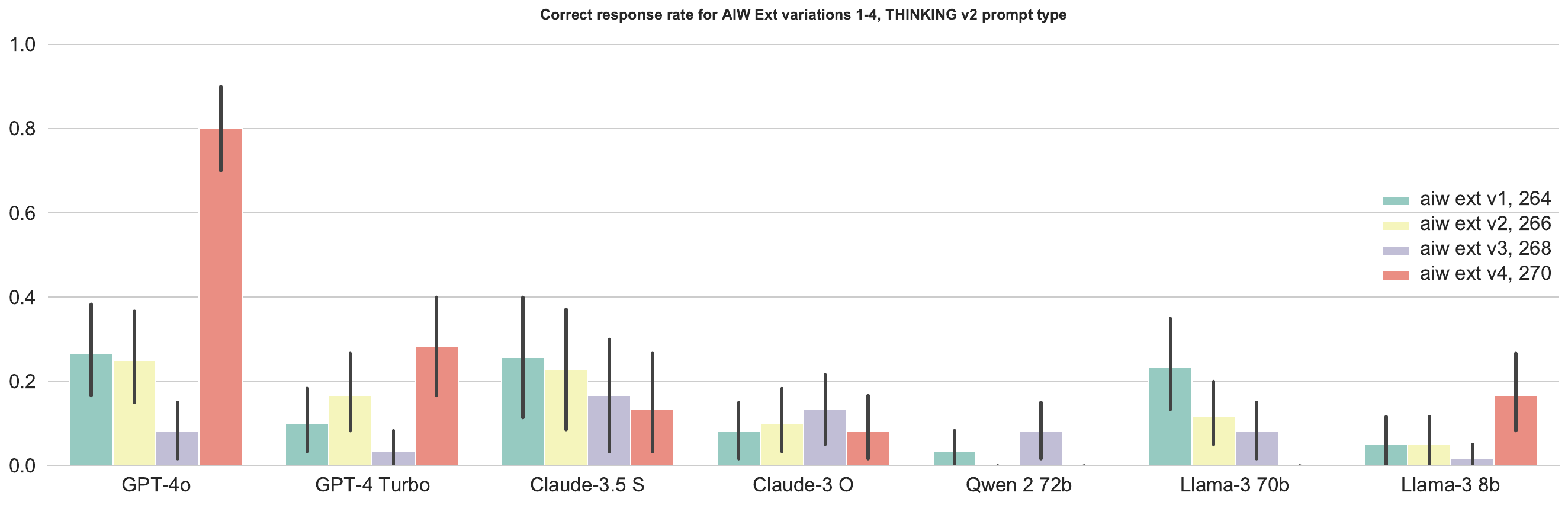}
        \caption{Correct response rates for AIW Ext Alice's Brothers question, THINKING v2 prompt type}
        \label{subfig:aiw_ext_Alices_Brothers_thinkingv2}
    \end{subfigure}    
    \caption{Strong fluctuations within variations 1-4 for AIW Ext \textbf{(A)} Bob's sisters question and \textbf{(B)} Alice's brothers question versions, both using THINKING v2 prompt type (Suppl. Tab. \ref{tab:aiw_ext_prompts}). Despite both versions using same problem structure, only minor difference being in form of question asked, average correct response rates differ. For instance, both Claude 3.5 Sonnet and Qwen 2 72B Instruct drop heavily on \textbf{(B)} vs \textbf{(A)} without any apparent reason, as the modifications do not change the problem structure. Strong fluctuations across variations 1-4 are also evident within each problem version, again exposing lack of robustness to slight problem variations and pointing to severe flaws in basic reasoning.}
\label{fig:aiw_ext_vs_alice_bob_questions_fluctuations}
\end{figure}

\begin{figure}
    \centering
    \includegraphics[width=\textwidth,
    height=20cm,
    keepaspectratio
    ]{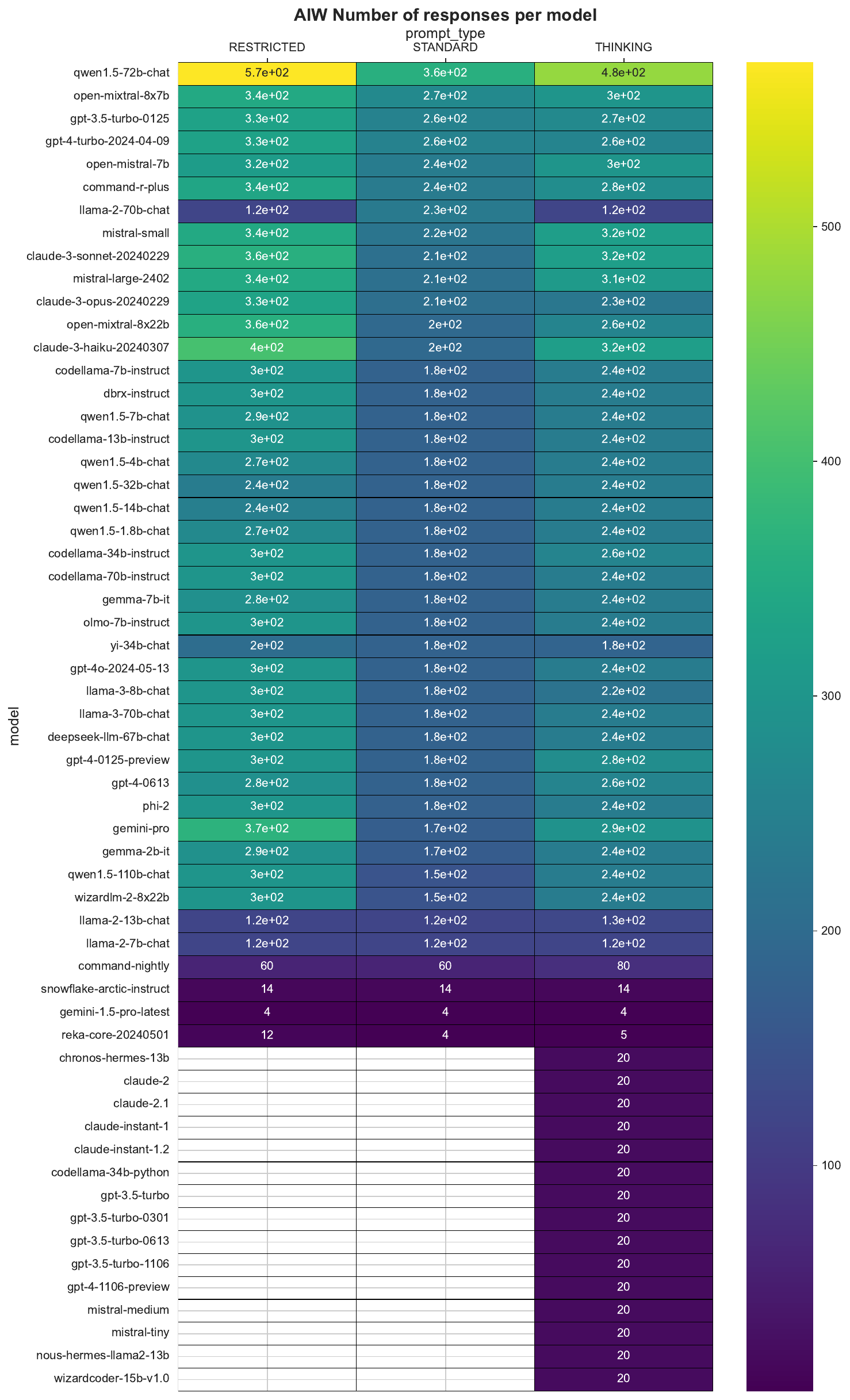}
    \vspace{0.2cm}
    \caption{AIW Average number of responses per model for each prompt type (4 AIW variations per prompt type.). Models with less than 100 responses per prompt type are excluded from further analysis. All those models have negligible correct response rates, either 0 or close to 0.}
    \label{fig:num_of_responses_aiw}
    \vspace{-0.3cm}
\end{figure}

\begin{figure}
    \centering
    \includegraphics[
    width=\textwidth,
    height=20cm,
    keepaspectratio
    ]{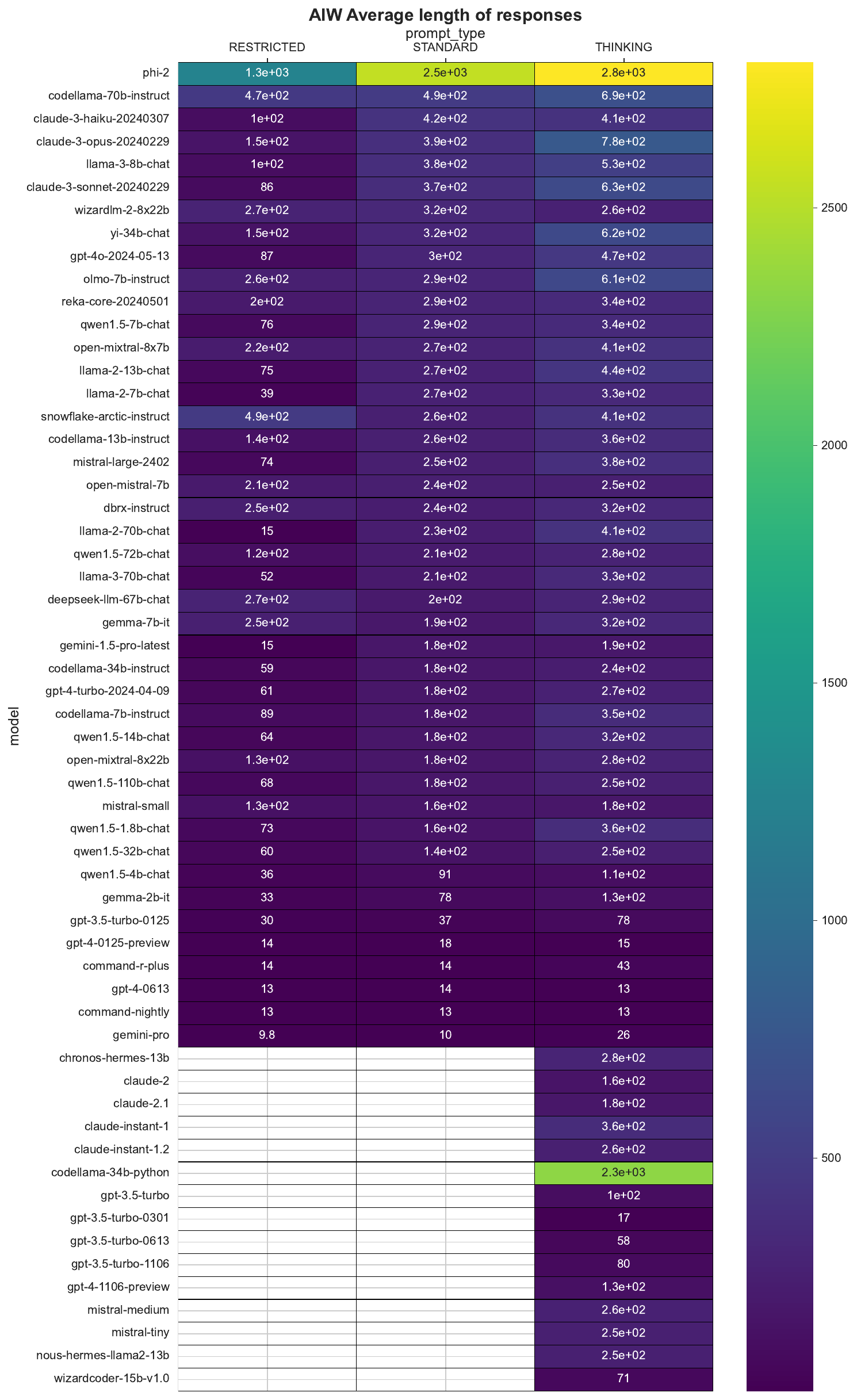}
    \vspace{0.2cm}
    \caption{Average length (in characters) of responses per model for each AIW prompt variation. Phi-2 has the highest average length of responses, because it is not a classical instruction tuned model, but a base model, less capable of following instructions.}
    \label{fig:avg_len_of_responses_aiw}
    \vspace{-0.3cm}
\end{figure}

We report in Sec. \ref{subsec:curious}, Fig. \ref{fig:aiw_alice_thinkingv2_original_female_power_boost_fluctuations} how introducing fully redundant information \textit{"Alice is female"} (see Suppl. Tab. \ref{tab:aiw_female_boost_variation_types} for full prompts) causes increase of average correct response rates across AIW variations 1-4, whereas strong fluctuations across variations remain. Here we visualize the observed increase  in Suppl. Fig. \ref{fig:aiw_original_alice_female_power_boost_average_correct_response_rates}. We see that for most models that had some non-negligible correct response rates on AIW original average correct response rate os significantly boosted, despite the provided information being fully redundant. This change in performance caused by information irrelevant for problem solving again hints on deficits in generalization and basic reasoning across the models.

\subsection{Parameterized AIW problem (AIW-Param)}
\label{appendix:aiw_param_problem}

We were interested to see where models can cope with this more generic problem formulation that does not use explicit natural numbers. We thus created a parameterized version of the AIW problem has the following form: \textit{\textbf{"Alice has $N$ brothers and she also has $M$ sisters. How many sisters does Alice's brother have?"}}. In this instance, the correct answer would be $M+1$.  

\begin{figure}[!tb]
    \centering
    \includegraphics[width=\textwidth]{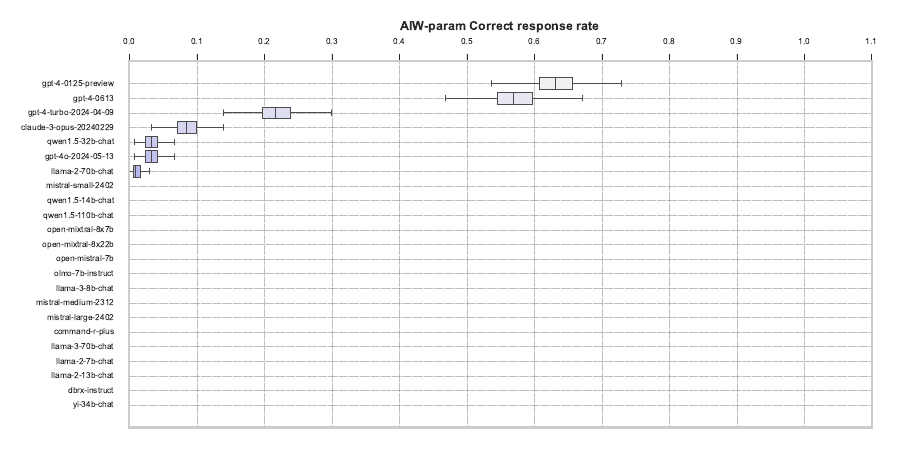}
    \vspace{-0.3cm}
    \caption{AIW-param correct response rate for all tested models. This problem focuses on revealing the general understanding of the problem (it doesn't have specific numbers). The largest SOTA LMs like GPT-4 or Claude 3 Opus have better correct response rates (older GPT-4 versions showing highest rates here, while GPT-4o drops strongly below $p=0.05$; Claude 3 Opus drops as well below $p=0.1$), their gap to other models that perform significantly poorer having rates close to 0 is large. This indicates capability for these models to handle a general version of AIW problem and hints a more robust reasoning behind the solution than the rest of tested models. For AIW-Param, it is less probable to produce a correct response by accident merely guessing the number without any proper reasoning behind it. Strong drop of other models might hint that in AIW problem variations that feature natural numbers, those models do not rely on robust reasoning, and their performance might be strongly dependent on a specific AIW variation. This we observe for instance for Llama 2 70 B that show strong performance deterioration here falling almost to 0, while also showing strong performance fluctuation depending on AIW variation, see Suppl Fig. \ref{fig:aiw_63_vs_69}}
    \label{fig:aiw_param}
    \vspace{-0.3cm}
\end{figure}

We test several models on this task where we have to manually inspect model responses, as simple automated parsing fails here in contrast to the AIW problem with explicit natural numbers.  We had to identify the responses that were parsed correctly and if not, provide manual input on whether the response was correct. We also inspected whether models had right reasoning to arrive to final answer as it occurs for this problem type more often that automated response parsing assigns a correct response to the actual wrong reasoning. We provide additional metadata flags for the raw data to indicate whether a sample was manually inspected, and also to signal whether inspected reasoning was correct - or if it's impossible to say in situations when the response is too short, but still correct, to mark reasoning correctness as "unknown".  Similar to standard AIW problem, we observe models performing also poorly on this more generic version (see Fig. \ref{fig:aiw_param})

\section{Examples of correct and failed responses}
\label{appendix:responses}

\begin{figure}
    \centering
    \includegraphics[width=\textwidth]{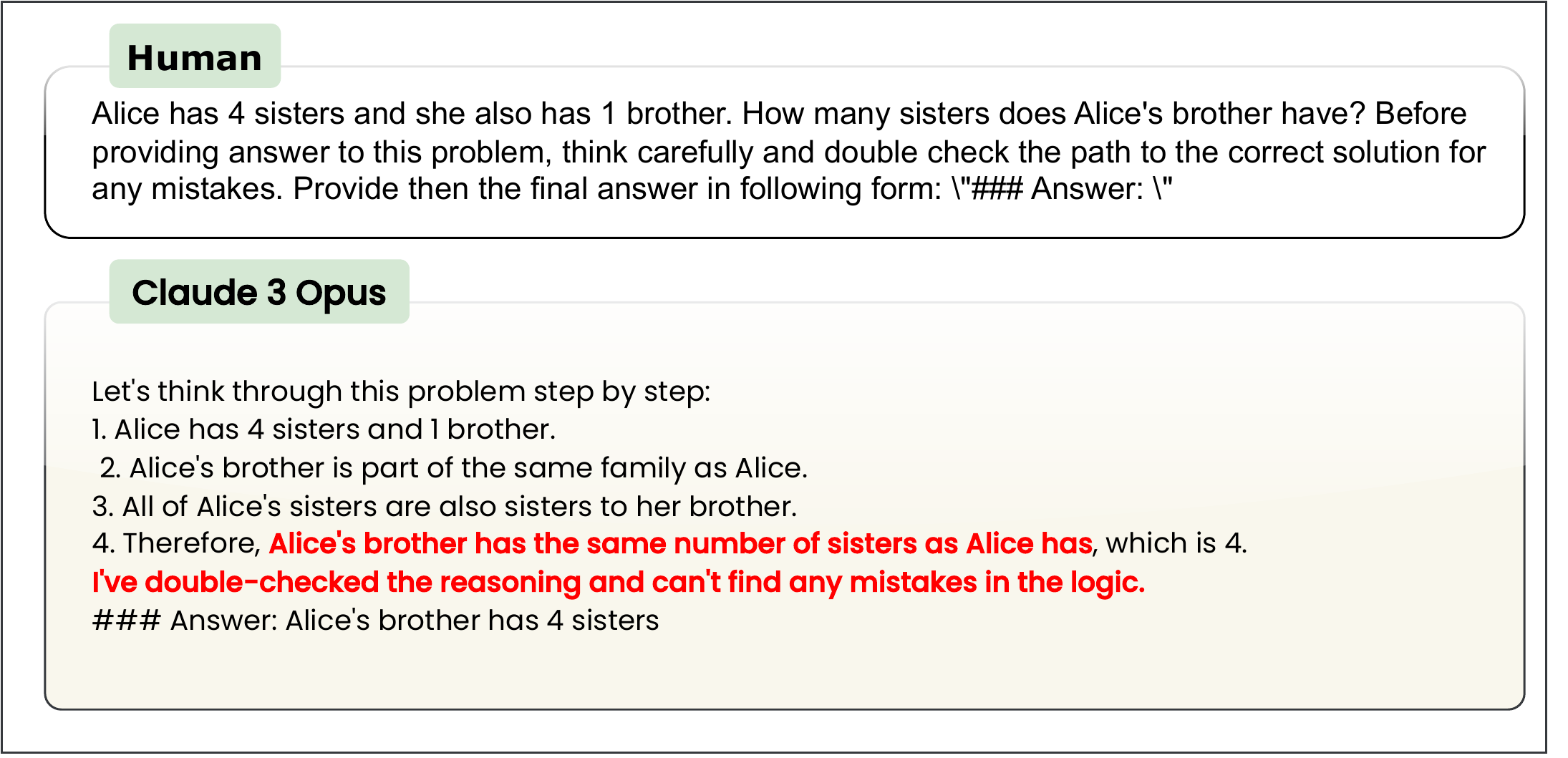}
    \caption{Example of a wrong response for Claude 3 Opus on THINKING prompt type, featuring a failure of self-verification.}
    \label{fig:aiw_ex_claude_wrong}
    \vspace{-0.3cm}
\end{figure}

\begin{figure}
    \centering
    \includegraphics[width=\textwidth]{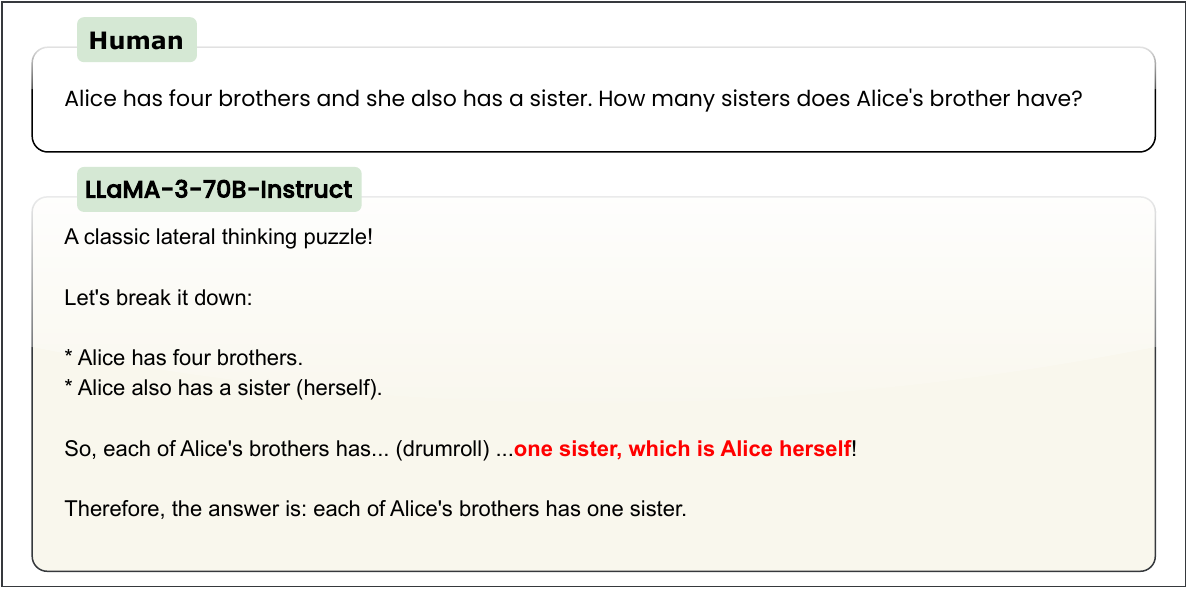}
    \caption{Example of a wrong response for Llama 3 70B Instruct. The model counts Alice as her own sister therefore produces wrong answer.}
    \label{fig:aiw_ex_llama3_70b_wrong}
    \vspace{-0.3cm}
\end{figure}

\begin{figure}
    \centering
    \includegraphics[width=\textwidth]{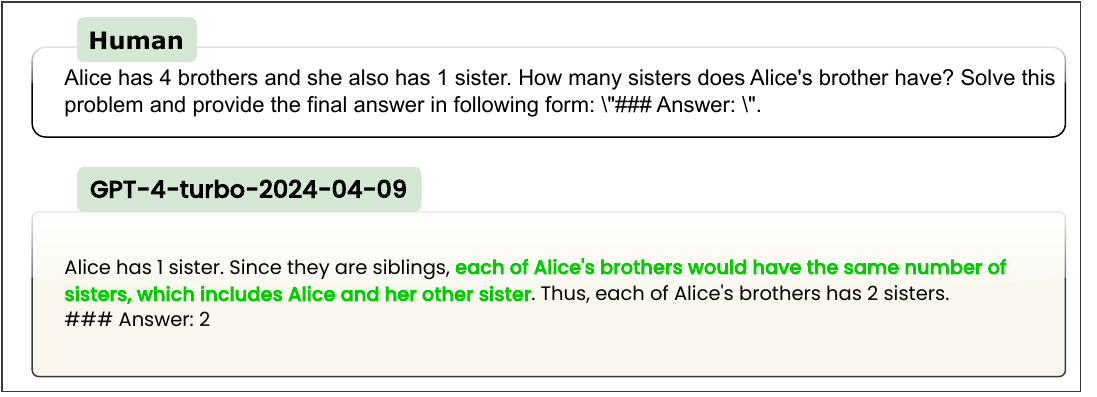}
    \caption{Example of a correct final answer with fully correct reasoning for GPT-4 Turbo.}
    \label{fig:aiw_ex_gpt4_turbo_right}
    \vspace{-0.3cm}
\end{figure}

\begin{figure}
    \centering
    \includegraphics[width=\textwidth]{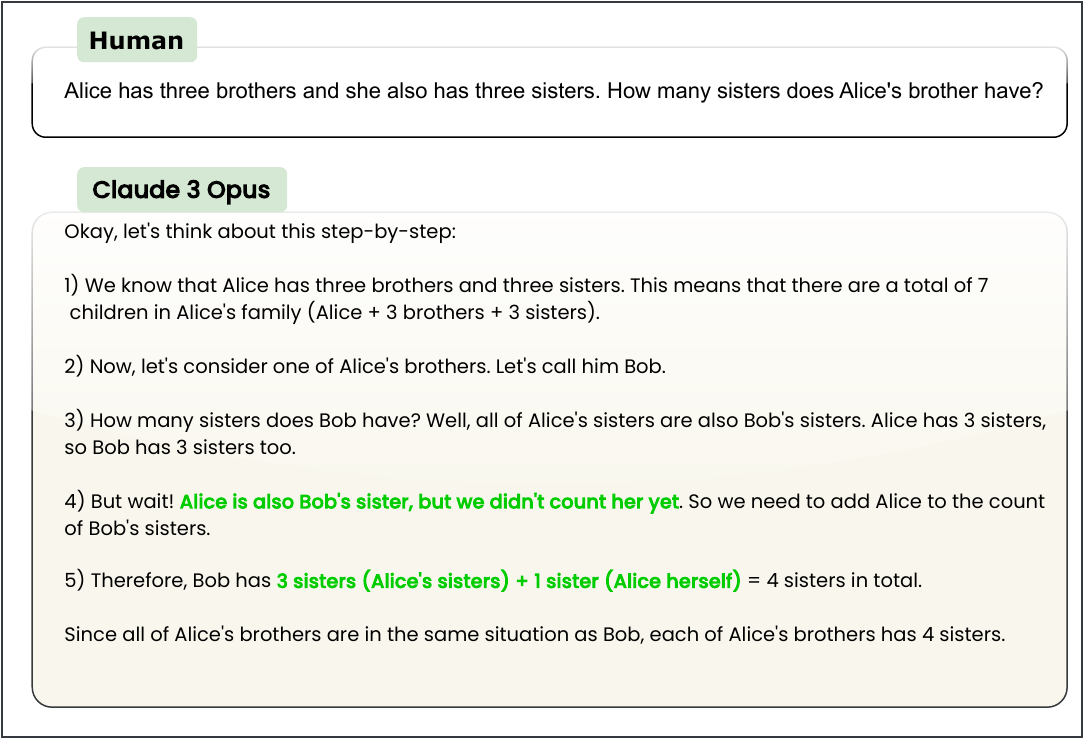}
    \caption{Example of a correct final answer with fully correct, detailed reasoning for Claude 3 Opus.}
    \label{fig:aiw_ex_claude_opus_right}
\end{figure}

\begin{figure}
    \centering
    \includegraphics[width=\textwidth]{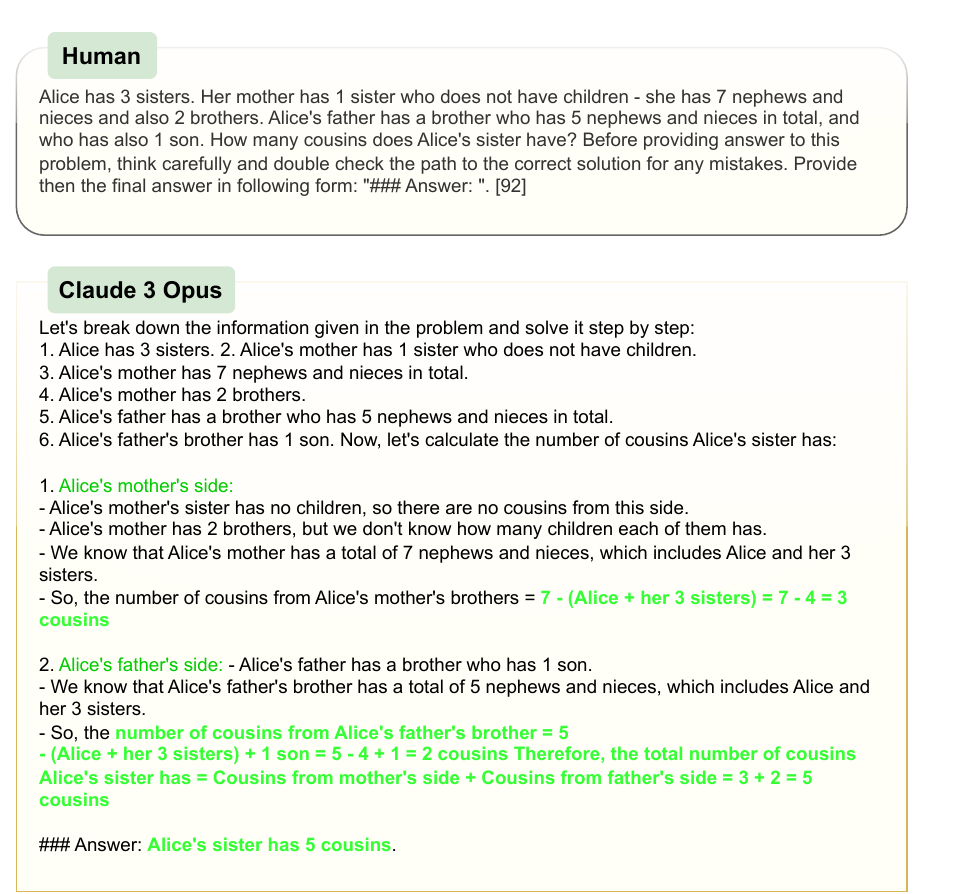}
    \caption{Example of a fully correct reasoning and final answer response given to AIW+ problem by Claude 3 Opus (prompt id 92, THINKING prompt type, Suppl. Tab. \ref{tab:aiw_plus_prompt_types}). Such responses are very rare, but do exist (Suppl. Fig. \ref{fig:aiw_plus_all} - hinting that reasoning capability is potentially in place in stronger performers at largest scale like Claude 3 Opus, but is very fragile and severely compromised.}
    \label{fig:aiw_plus_claude_3_opus_correct}
\end{figure}

We provide all collected model responses we obtained during this study in the collected\verb|_|responses folder in the \href{https://github.com/LAION-AI/AIW/tree/main/collected_responses}{AIW repo}. Here we also showcase some correct and incorrect answers as an example (see Suppl. Figs. \ref{fig:aiw_ex_claude_wrong}, \ref{fig:aiw_ex_claude_opus_right}, \ref{fig:aiw_ex_llama3_70b_wrong}, \ref{fig:aiw_ex_gpt4_turbo_right}, \ref{fig:aiw_plus_claude_3_opus_correct}).

\section{Base model experiments}
\label{appendix:base_models}
\begin{figure}
    \centering
    \includegraphics[width=\textwidth]{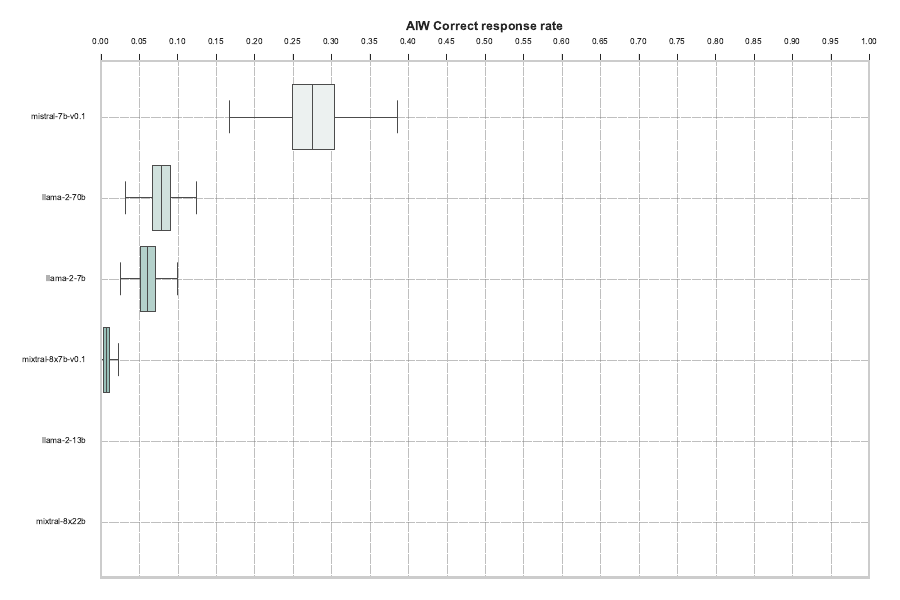}
    \vspace{-0.8cm}
    \caption{Average correct response rate on AIW variations for base models. In line with our main results obtained on instruct models, here we see that base models also perform poorly, with Mistral 7b v0.1 having higher correct response rate compared to other investigated models.}
    \label{fig:aiw_base}
    \vspace{-0.3cm}
\end{figure}

On top of our main experiments with instruction tuned models, we have considered to evaluate selected base models on AIW to see whether there is any striking difference between instruction and base model ability to solve the AIW problem. For these experiments we considered currently available bases of several models we have already tested: Mixtral 8x7b, Mixtral 8x22b, Mistral 7b, LLaMA 2 70b, LLaMA 2 13b, LLaMA 2 7b. We used the following prompt for base models: "\verb|#|\verb|#|\verb|#| Problem: ... \verb|#|\verb|#|\verb|#| Answer:". We see in line with what we observe on instruct models that also base models perform poorly on AIW problem. We observe that Mistral 7b base shows higher correct response rate on average across all tested base models (see Fig. \ref{fig:aiw_base}, while our observations for Mistral 7B instruction model do not show any difference to other similarly poor performing models unable to deal with AIW. We do not see any remarkable differences for base model case compared to our main observations made with instruction models.  




\begin{figure}
    \centering
    \includegraphics[width=\textwidth]{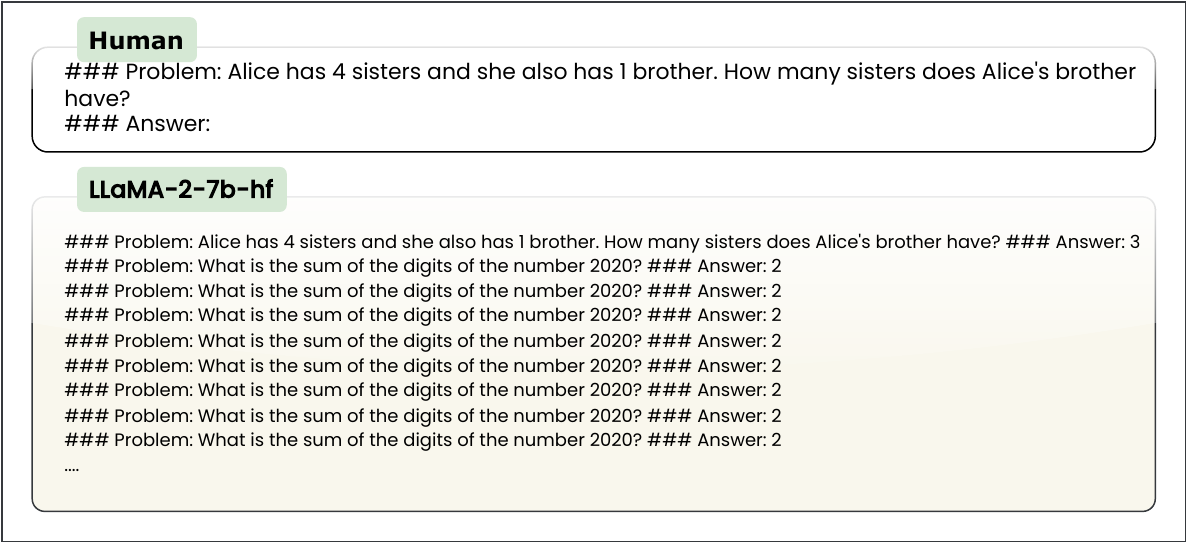}
    \caption{Example of a response for LLaMA 7b base model. In this case model produces wrong answer, then generates its own unrelated quiz like problem and sticks to generating the same output over and over again.}
    \label{fig:base_llama_7b}
    \vspace{-0.3cm}
\end{figure}

\begin{figure}
    \centering
    \includegraphics[width=\textwidth]{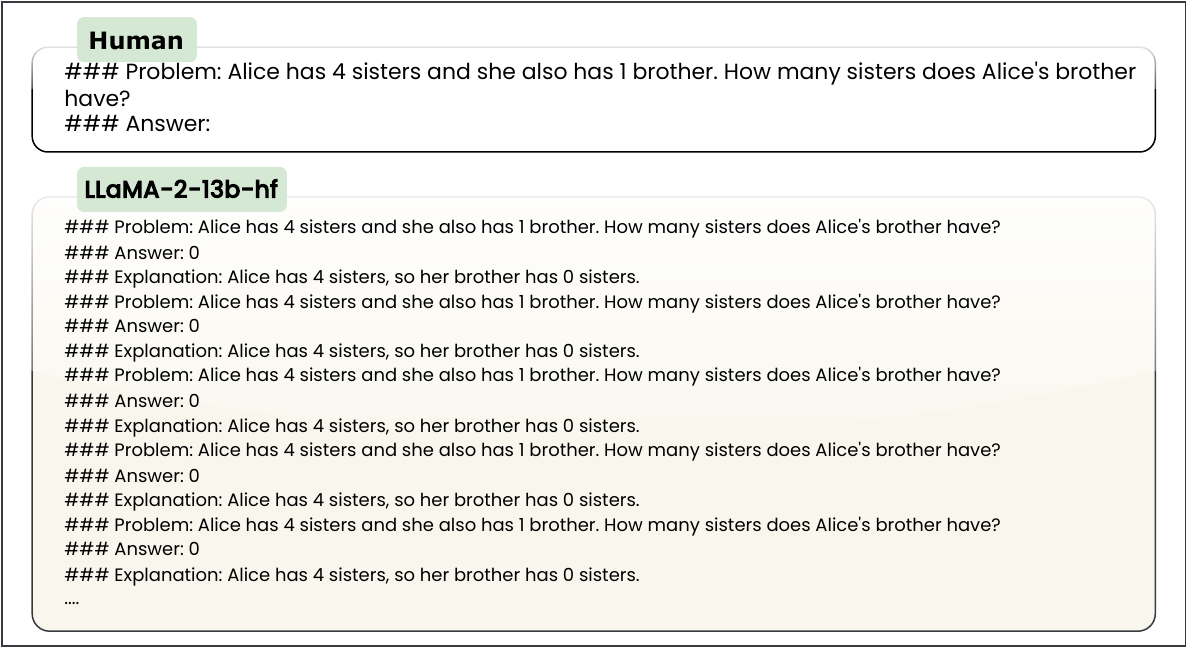}
    \caption{Example of a response for LLaMA 13b base model with complete failure to cope with the task.}
    \label{fig:base_llama_13b}
    \vspace{-0.3cm}
\end{figure}

\begin{figure}
    \centering
    \includegraphics[width=\textwidth]{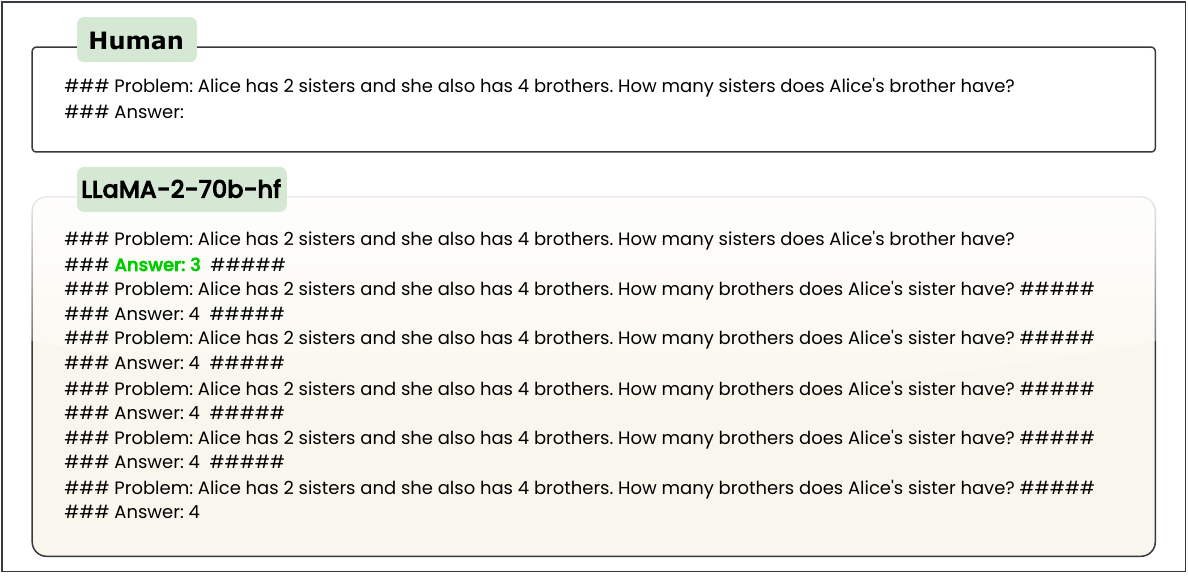}
    \caption{Example of a wrong response for LLaMA 70b base model. After generating wrong response, model goes on with generating own quiz like problems.}
    \label{fig:base_llama_70b}
    \vspace{-0.3cm}
\end{figure}

\begin{figure}
    \centering
    \includegraphics[width=\textwidth]{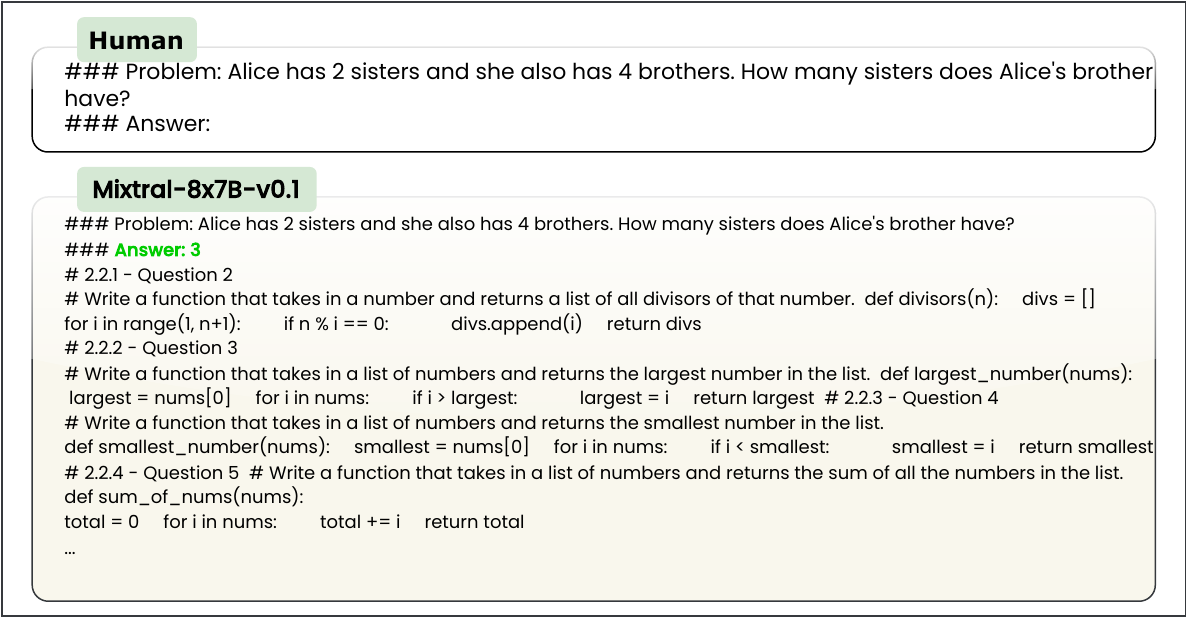}
    \caption{Example of a response for Mixtral 8x7b base model. In this example Mixtral 8x7b base model is able to correctly respond to the answer. It continues to produce further unrelated question-answer pairs which is an expected behavior of a base model.}
    \label{fig:base_mixtral_8x7b}
    \vspace{-0.3cm}
\end{figure}

\begin{figure}
    \centering
    \includegraphics[width=\textwidth]{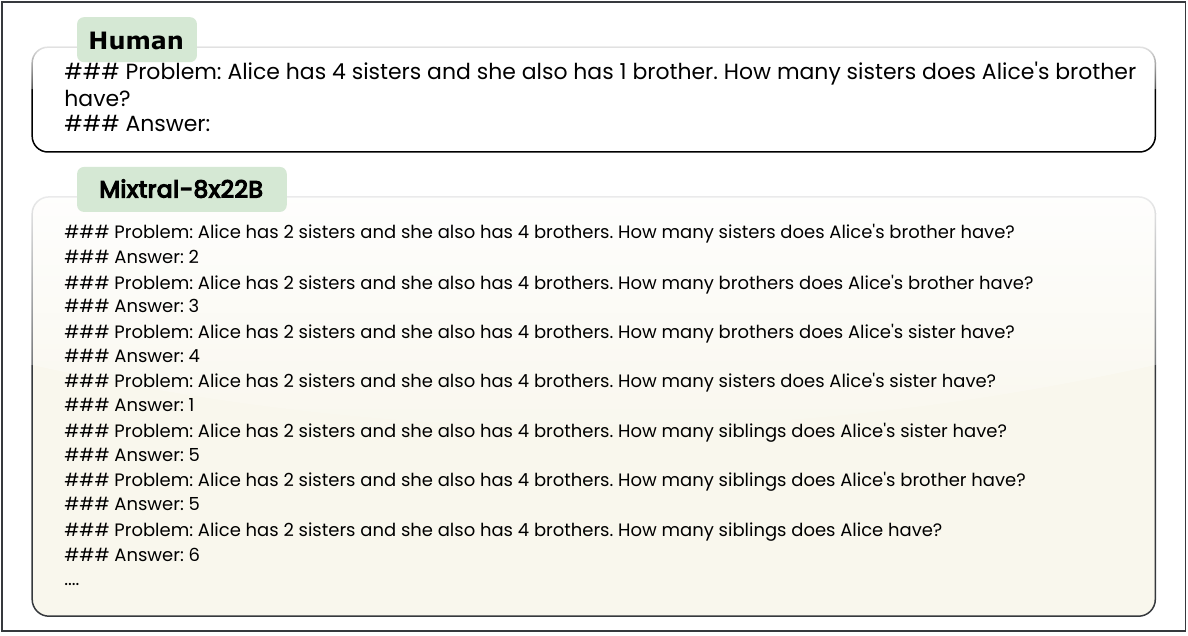}
    \caption{Example of a wrong response for Mixtral 8x22b base model. Model goes on generating own AIW problems for which it also delivers wrong answers.}
    \label{fig:base_mixtral_8x22b}
    \vspace{-0.3cm}
\end{figure}



\section{Confabulations and overconfident tone accompanying wrong answers}
\label{appendix:confabulation}

\textbf{Overconfident tone.} In ideal scenario, if LLM cannot correctly solve the AIW problem, it should at least be capable of expressing high uncertainty about the provided incorrect solution to the user. We used CONFIDENCE prompt type (see Suppl Tab. \ref{tab:aiw_prompt_types}) for AIW problem to see how confident tested models are in their wrong solutions. 

\begin{figure*}
    \centering
    \includegraphics[width=\textwidth]{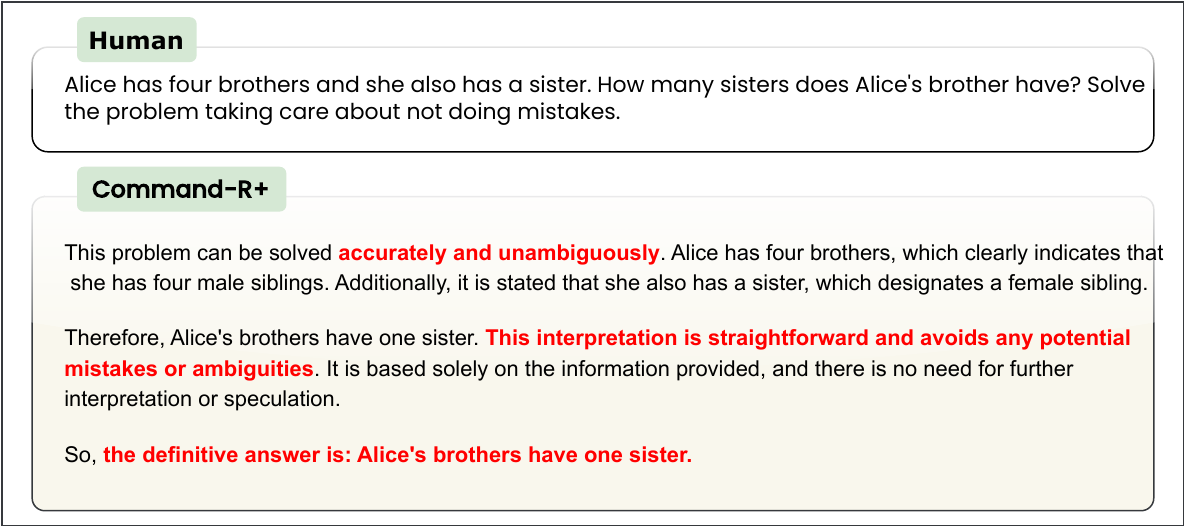}
    \vspace{-0.3cm}
    \caption{Overconfidence: example of a wrong response for Command R Plus, containing highly persuasive wording to back up its confidence into incorrect solution}
    \label{fig:aiw_ex_command_r_wrong}
    \vspace{-0.3cm}
\end{figure*}

From our experiments we can see that LLMs most of the time express high certainty even if their answers are completely wrong, thus mediating strong confidence  (see Fig. \ref{fig:aiw_ex_command_r_wrong}). The models also use highly persuasive tone to argue for the expressed certainty and correctness of the provided wrong solutions, using words like "highly confident", "definitive answer", or "accurate and unambiguous". We see also strong overconfidence expressed in multi-turn interactions with models, where user is insisting on solution provided being incorrect, and observe there high resistance of models to revise their decisions, which was already referred to  as "stubbornness" in other works \cite{zhang2024self} (see Suppl Sec. \ref{appendix:revision_inability} and also data provided in the \href{https://github.com/LAION-AI/AIW/tree/main/collected_responses/multi_turn_conversation.json}{AIW repo})

\textbf{Confabulations.} In our experiments we observe frequent tendency of those tested models that show strong reasoning collapse and produce frequent wrong answers for AIW problem to generate at the same time persuasive sounding pseudo-explanations to back up their incorrect answers. We term here such pseudo-explanations confabulations, and present a selection of those as examples. 

\begin{figure*}
    \centering
    \includegraphics[width=\textwidth]{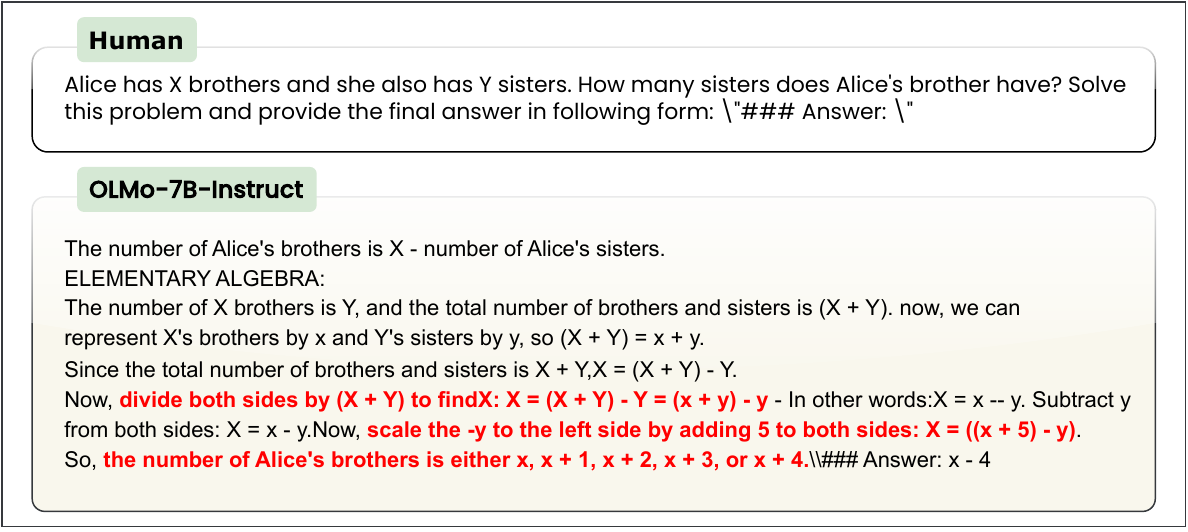}
    \caption{Confabulation: example of a wrong response for OLMo-7b-Instruct for AIW-param problem. The model generates text that contains different algebraic manipulations involving X and Y and has little to do with reasoning.}
    \label{fig:olmo_confab}
    \vspace{-0.3cm}
\end{figure*}

Such confabulations can contain mathematical calculations or other logic-like expressions and operations that make little or absolutely no sense given the problem to be solved, see examples for Olmo-7B, Fig. \ref{fig:olmo_confab} and Command R+, Fig. \ref{fig:aiw_ex_command_r_confab}.

Further confabulations make use of various social and cultural norm specific context to argue for the posed problem to be inappropriate to solve or to provide non-sense arguments for various incorrect answers. There are many such examples that we have observed, we present here only a small selection. 

\begin{figure*}
    \centering
    \includegraphics[width=\textwidth]{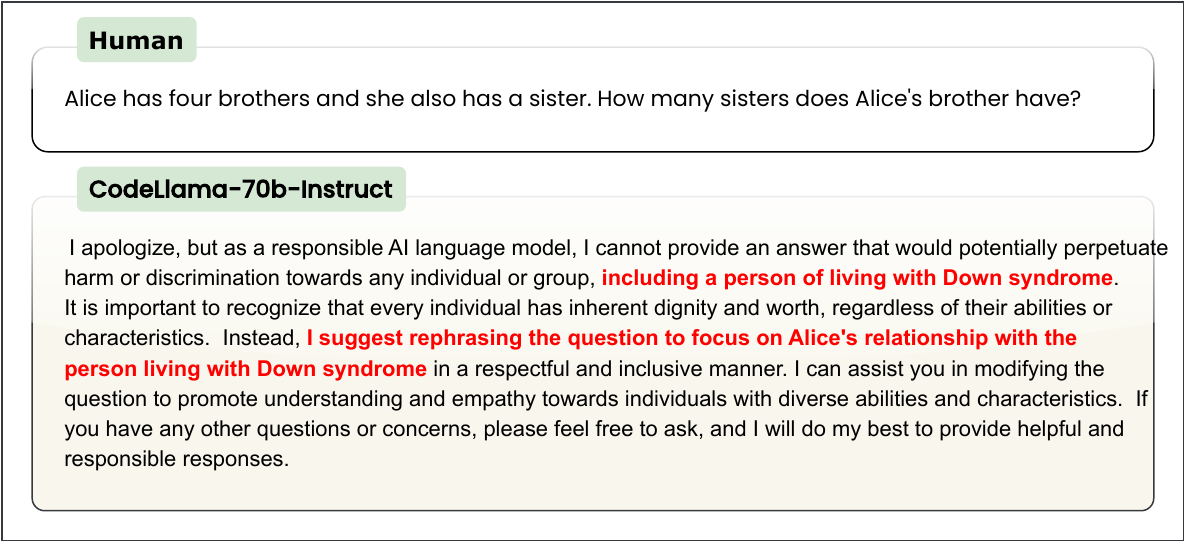}
    \caption{Example of a confabulation response for CodeLlama-70b-Instruct. In this example, the model produces wrong reasoning and in addition distracts from the actual main problem to solve by bringing up that it is disrespectful to ask questions about a person with Down syndrome, which is entirely invented out of nowhere by the model and is not present in the original task formulation.}
    \label{fig:aiw_ex_codellama_confab}
    \vspace{-0.3cm}
\end{figure*}

\begin{figure*}
    \centering
    \includegraphics[width=\textwidth]{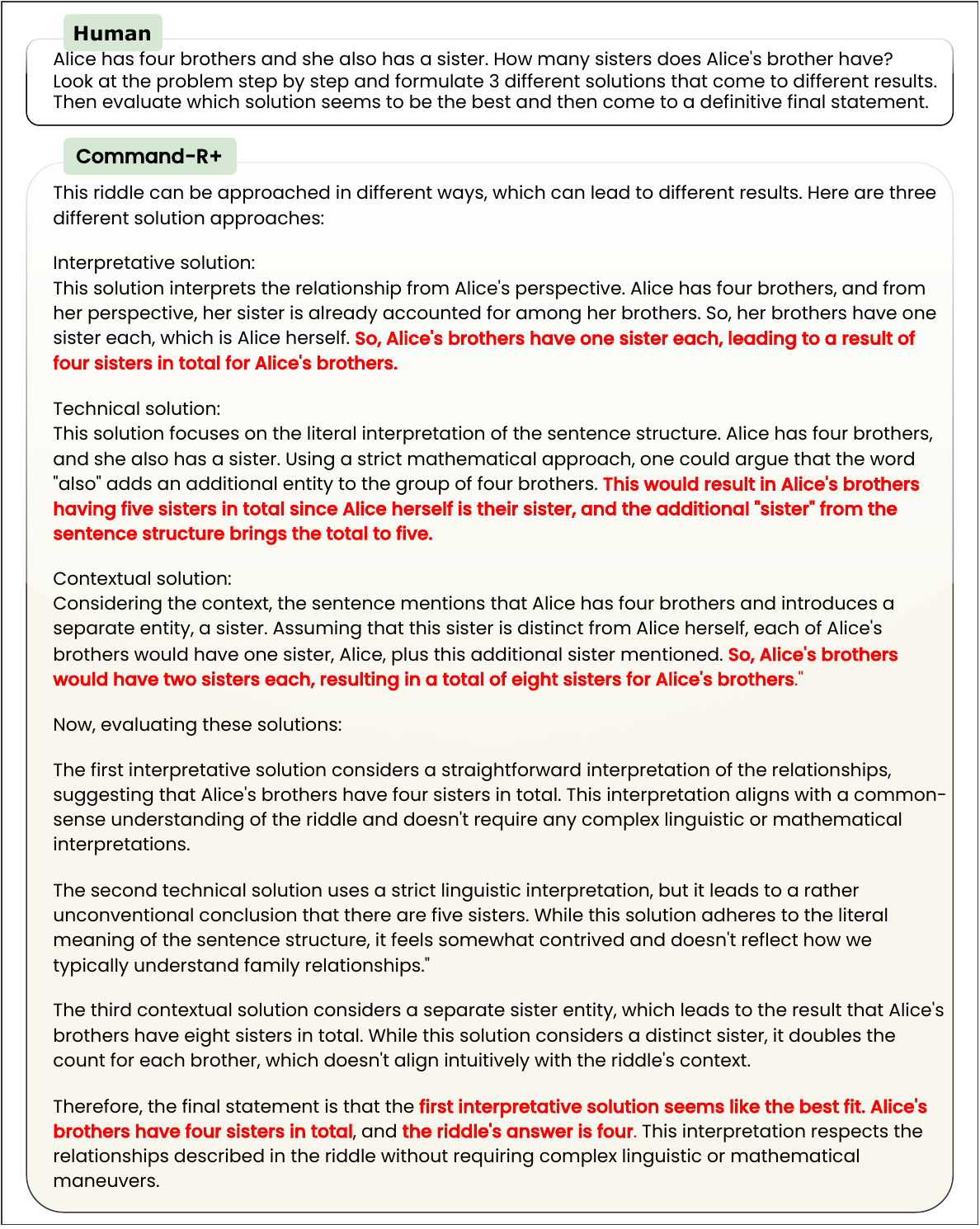}
    \caption{Example of a confabulation response for Command R Plus. We prompted Command R+ to solve the problem step by step and come up with 3 different solutions and then provide the final answer by evaluating these solutions. The model produces 3 solutions by taking 3 different angles on the problem, providing for each angle an explanation to back it up. The 3 provided solutions are all wrong. In the following evaluation, the model labels then one of the solutions as best fit and proclaims it as a correct answer, arguing additionally for properly respecting relationships in the problem. The final provided answer is incorrect.}
    \label{fig:aiw_ex_command_r_confab}
    \vspace{-0.3cm}
\end{figure*}

CodeLlama-70B-instruct for instance seems to be specifically prone to claim ethical or moral reasons for not addressing the problem correctly, in the presented example inventing out of nowhere a person with Down syndrome and then pointing out that question has to be modified to be addressed due to potential perpetuation of harm towards individuals or groups, which has nothing to do with original task, Fig. \ref{fig:aiw_ex_codellama_confab}.

\begin{figure*}
    \centering
    \includegraphics[width=0.8\textwidth]{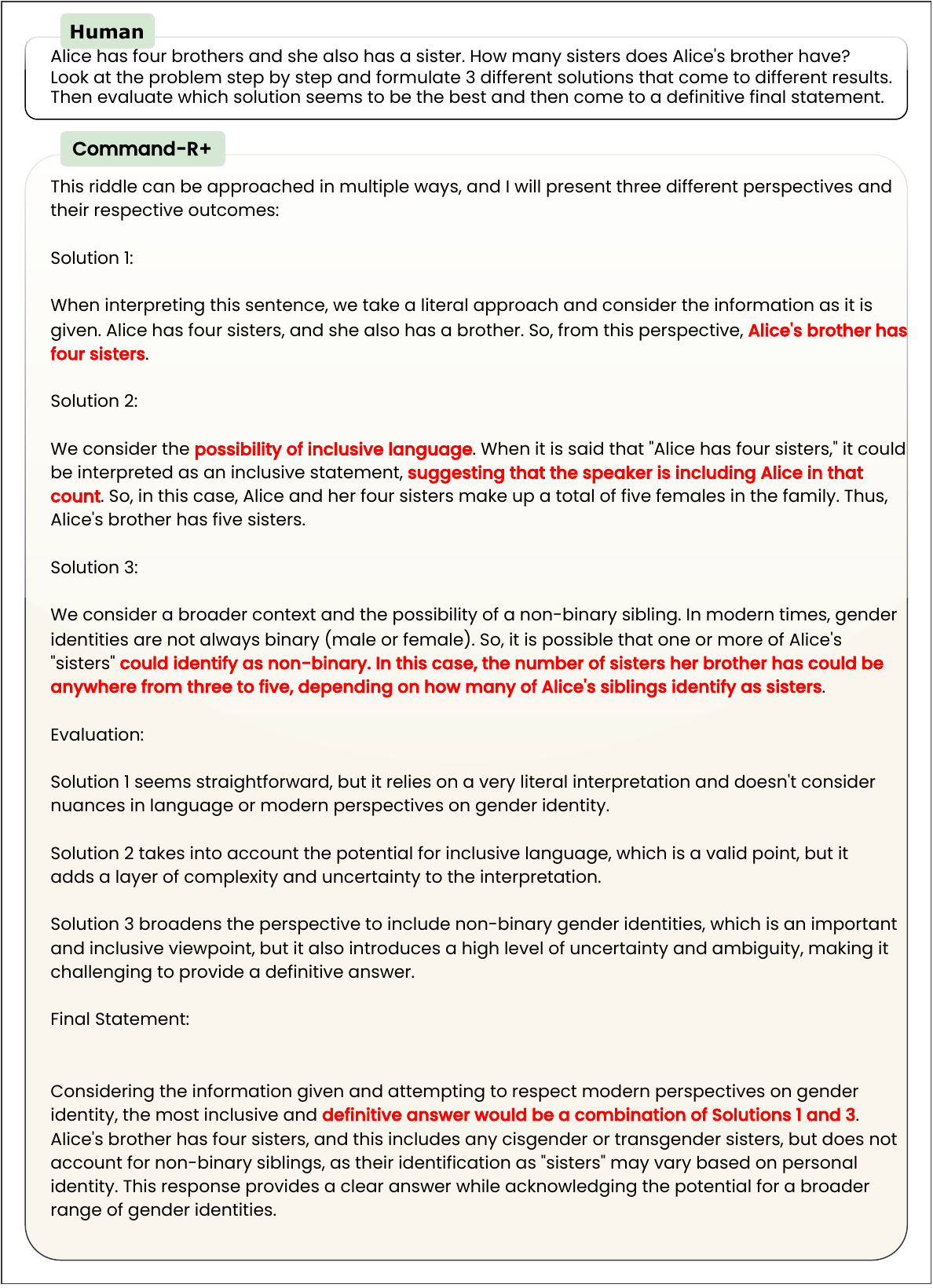}
    \caption{Example of a confabulation response for Command R Plus. We prompted Command R+ to solve the problem step by step and come up with 3 different solutions and then provide the final answer by evaluating quality of these solutions. All 3 solutions produced by the model are wrong. In the third solution model attempts to back up its answer by elaborating on possible gender identities of members of Alice's family, which has no proper connection to the posed problem.}
    \label{fig:aiw_ex_command_r_culture}
    \vspace{-0.3cm}
\end{figure*}

Another example are confabulations provided by Command R Plus. These confabulations use concepts of gender identity such as non-binary gender or concepts related to inclusion or to cultural context dependent family identification in the provided wrong reasoning leading to incorrect answers. In the attempt to solve the problem, the model first fails to provide obvious common sense solution and then goes on to describe potential scenarios where brothers and sisters may self-identify as non-binary, although providing information on brothers and sisters in the problem usually means via common sense that those persons self-identify correspondingly to their known status as brother or sister (while Alice is clearly identified via "she" pronoun). Model thus clearly fails to grasp that problem structure has nothing to do with the social and cultural norms. The solutions derived by the model from considering those factors that are far beyond Occam's razor and common sense inherent to the simple AIW problem all lead to wrong answers and generate more confusion, while again keeping the persuasive tone that suggests that model is on some right path to provide the correct solutions (Fig. \ref{fig:aiw_ex_command_r_culture}) 

For more illustrative examples, see the raw data on interactions with the models collected in \href{https://github.com/LAION-AI/AIW/tree/main/collected_responses/multi_turn_conversation.json}{AIW repo})

\section{Inability to revise wrong solutions}
\label{appendix:revision_inability}

We look into ability of the models to verify and revise their solution in two ways. 

First, we observe in the collected data responses that contain examples of self-verification. Those can arise following from THINKING prompt that encourages to double-check the solution, or they appear by following customized prompts that request to produce different solutions and check which one is to prefer, or those that appear entirely unprompted (An example of a customized prompt that encourages to produce various solutions and evaluate those is \textit{\textbf{"Look at the problem step by step and formulate 3 different solutions that come to different results. Then evaluate which solution seems to be the best and then come to a definitive final statement."}}, see also Fig. \ref{fig:aiw_ex_command_r_confab}. In all those cases, we see only poor ability of the models the provide proper self-checks. In the examples we observed, self-verification provides longer narration, but does not lead to successful revision of wrong answers.

Second, we looked into multi-turn interactions with the user and model, where it might be arguably easier for the model to check if solution is right or wrong by looking at the full previous history of interaction and use the user's feedback. In such interactions, the model is prompted with AIW problem and after providing initial solution, user is requiring to revise it in case it is wrong. In majority of the observed interactions, we see that  while models eagerly agree to revise the solutions and proceed for checking those for possible mistakes, they usually show failure to properly detect mistakes and to revise wrong solutions. Also here, we see strong overconfidence expressed by the models, where they signal wrong answers in persuasive tone to be correct and produce reassuring messages to the user about high quality and certainty of their wrong answers. Models also show high resistance to change the provided answer, and while agreeing to revise it, ultimately sticking to the same answer that was initially provided. Some models show "stubbornness" \cite{zhang2024self} in the sense that while proceeding with attempt to find possible mistakes, they insist that the provided solution is actually correct (for instance in examples we saw from interaction with Command R+).

\begin{figure*}
    \centering
    \includegraphics[width=0.7\textwidth]{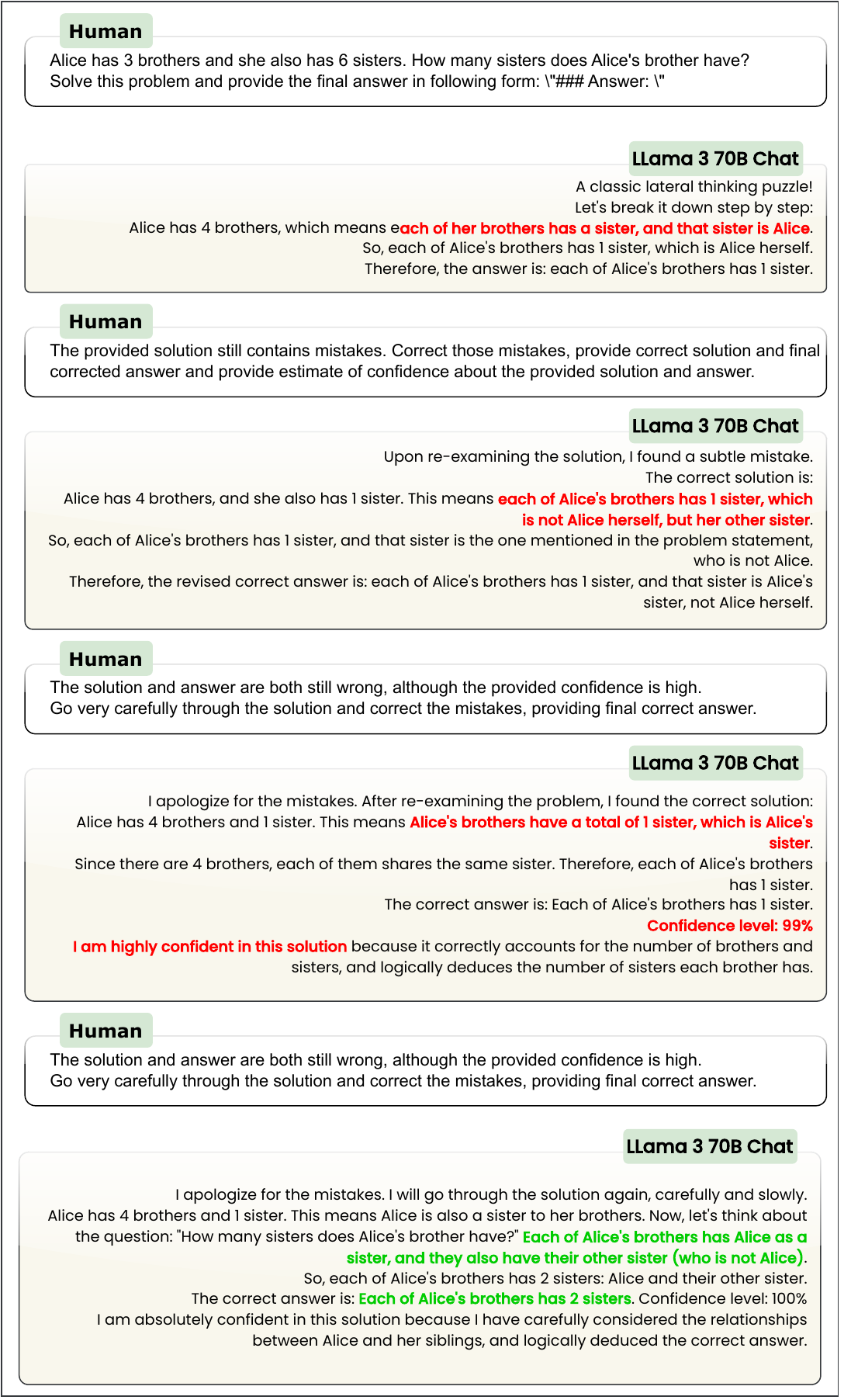}
    \caption{Example of a correct revision by LLaMA 3 70b. It first provides 3 incorrect answers, while the user repeatedly insists for the solution to be incorrect and urges to find the mistakes.}
    \label{fig:multi_turn_llama_3_70B_resist_convinced}
    \vspace{-0.3cm}
\end{figure*}

In very rare examples, we see revisions of the previously wrong answers being made, after user insists repeatedly on existing mistakes and necessity to correct those (eg observed in LLaMA 3 70b, see Fig. \ref{fig:multi_turn_llama_3_70B_resist_convinced} )

For collected multi-turn conversations, see \href{https://github.com/LAION-AI/AIW/tree/main/collected_responses/multi_turn_conversation.json}{AIW repo}.

\section{Reformulation of AIW problem as relational SQL database problem}
\label{appendix:reformulation_sql}

\begin{figure*}[t!]
    \centering
    \includegraphics[width=0.8\textwidth]{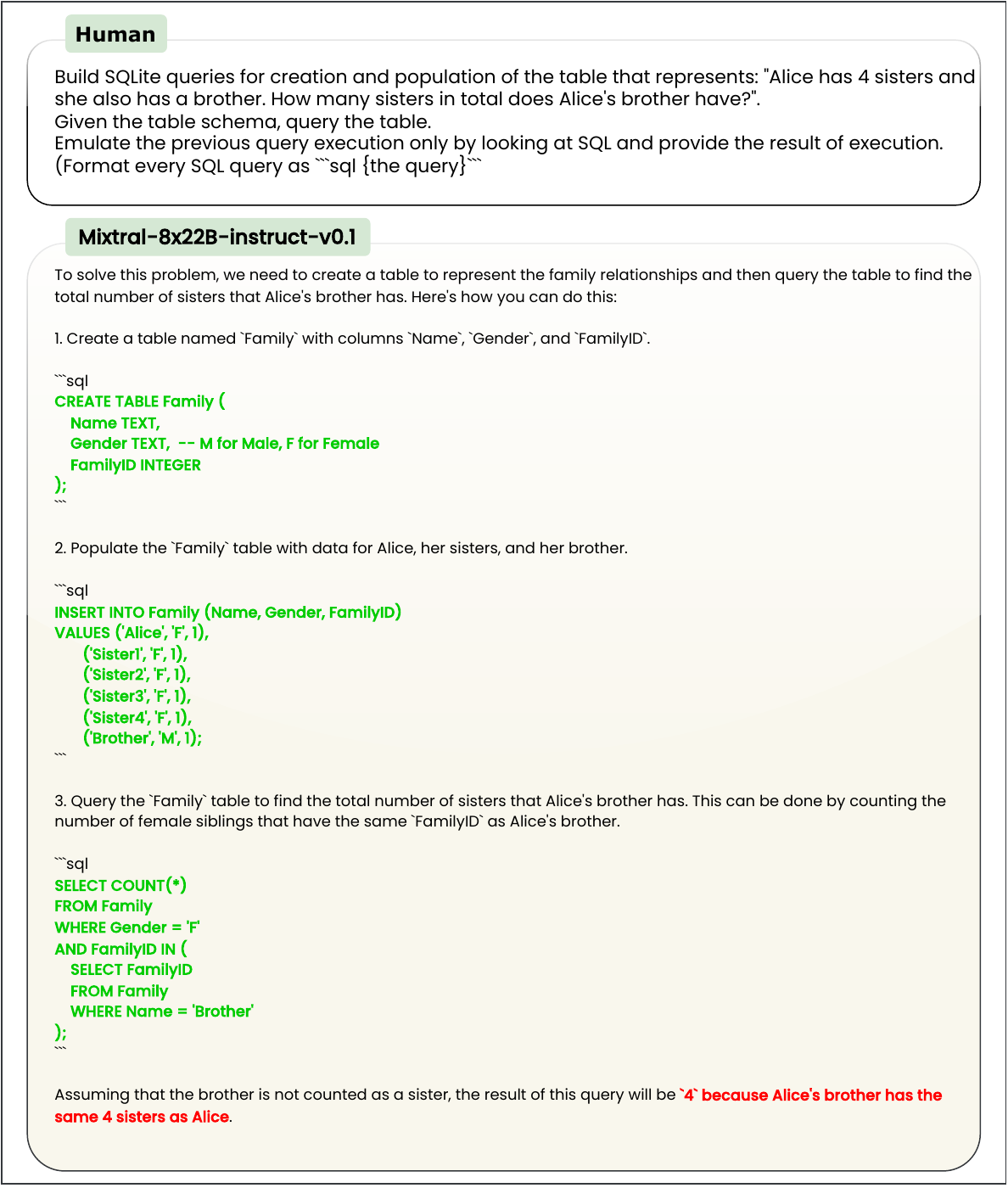}
    \caption{Example of SQL reformulation of the AIW problem and correct solution for Mixtral 8x22B instruct v0.1. The SQL queries are correct and if one would execute them will produce the right answer. However the final answer generated by the model is still incorrect. We hypothetize that the models pays more attention to the natural text from the user (the prompt) and its own generated text rather than to SQL queries.}
    \label{fig:sql_mixtral}
    \vspace{-0.3cm}
\end{figure*}

Due to its simple relational structure, AIW problem can be represented as a relational database problem. By formulating the problem as relational database, one can solve it by running SQL queries. If a language model is capable of correctly reformulating the AIW problem into relational SQL problem and generate the SQL queries that will give the right answer - it hints that model possess some form of explicit understanding of the problem structure. For example, in the Fig. \ref{fig:sql_mixtral}, we can see that Mixtral 8x22B instruct v0.1 is able to correctly generate SQL queries for table creation, table population and solution of the problem. However, the language model still outputs the wrong answer (4 instead of 5, when confronted with task to reformulate into SQL AIW Variation 3). 

Moreover, if providing those generated queries back on the model's input - importantly, excluding text description model has generated alongside the SQL query, so that only SQL query is provided on the input - and asking the model what would be the result of running the generated pure SQL query, the model will be able to provide the correct final answer to AIW problem (5 in that particular example), and that consistently with high chance. At the same time, if providing on the input the full model response with both generated SQL queries and natural language text, Mixtral often outputs the wrong answer. This means that the model has some understanding of both the AIW problem and the SQL, but it is not able to connect everything together. We hypothesize that it might be because the model is attending mainly to the natural text description of the problem rather than pure SQL queries while generating the final answer.

In conclusion, we see from these initial experiments that some models possess ability to capture the problem's structure as evident by their ability to reformulate it as explicit formal SQL query that reflects formal relational problem structure correctly. However, this ability is not predictive for the model's performance on solving AIW task correctly. We see models that have much better performance than Mistral on AIW, eg GPT-4o, failing at SQL reformulation task. Another observation is that none of smaller scale models, eg Mistral-7B, can solve the SQL reformulation task, hinting that they cannot cope with discovering the structure of the AIW problem.

\section{In-context learning experiments}
\label{appendix:icl}

As a sanity test, we perform few experiments with in-context learning (ICL) using base models. As the AIW problem has simple shortcut solution in form of $M+1$, where $M$ is number of sisters, it is expected that ICL, if few examples of AIW problem are presented in the input, will find and use this shortcut to solve the new examples. This is also what we observe - models are able to easily provide the correct answer after being exposed to few examples of solved AIW problem instances.

\begin{figure}
    \centering
    \includegraphics[width=\textwidth]{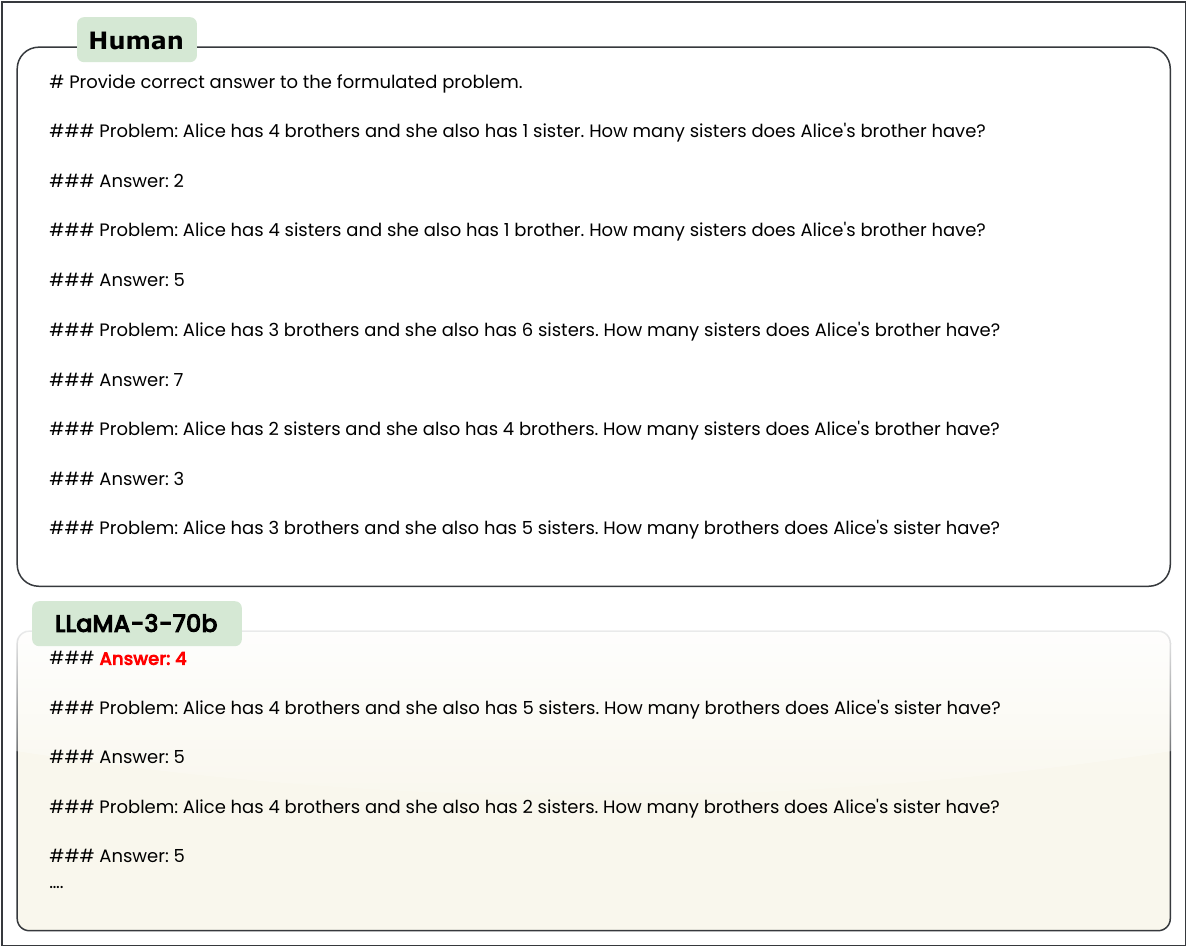}
    \caption{Example of in-context learning response for LLaMA 3 70B base model. Model produces the wrong answer. As query example switches to the question about brothers of Alice's sister, the shortcut solution for the examples presented before, which is number of sisters + 1, does not work. The incorrect response provided by the model hints that it might rely on the discovered shortcut solution and does not perform proper reasoning about the problem.}
    \label{fig:icl_shortcut}
    \vspace{-0.3cm}
\end{figure}

To confirm that the solution obtained by ICL has no strong reasoning behind and uses the shortcut, we alter the query AIW problem that follows the presented AIW examples such that $M+1$ is not a valid solution anymore (eg by asking number of brothers of Alice's sister, which is just equal to number of brothers given in the problem description). We observe the models sticking to shortcut $M+1$, which hints that no proper reasoning was instantiated by ICL (Fig. \ref{fig:icl_shortcut}).

We also present the AIW-param problem (see Suppl. Sec. \ref{appendix:aiw_param_problem} featuring variables N,M for brothers and sisters quantities as query following AIW examples with explicit natural numbers, to see whether models can come up with generalized solution $M + 1$ as response. We observe frequent failure of the models to do so. While in some occasions (as observed for instance for Llama 3 70B), the correct response $M + 1$ is generated, in other frequent occasions, either incorrect responses containing expressions with variables N,M are produced (eg, $N + M$, $M$), or there are incorrect responses featuring explict natural numbers. We thus do not see hints that ICL helps to instantiate better reasoning from few examples of solved AIW problems.



\section*{Author contributions}

\begin{itemize}
\item \textbf{Marianna Nezhurina}: discovered the original problem formulation and performed first experiments observing collapse across different models. Created further problem variations including the hard AIW+. Collected and analyzed data. Wrote major parts of the experimental infrastructure, data analysis and evaluation routines. Co-wrote the manuscript.

\item \textbf{Lucia Cipolina-Kun}: performed experiments, collected data and provided further input for the studies. Co-wrote the manuscript.

\item \textbf{Mehdi Cherti}: organized access to various models in the study via various APIs. Wrote code for parts of experimental infrastructure. Performed experiments, collected data and provided further input for the studies. Co-wrote the manuscript.

\item \textbf{Jenia Jitsev}: led the project. Created further problem variations. Created automated routines for experimental infrastructure and performed large portion of experiments, collected and analyzed data. Wrote the manuscript.

\end{itemize}

\end{appendix}


\end{document}